\documentclass{article} %
\usepackage[final]{colm2026_conference}
\usepackage[T1]{fontenc}

\usepackage{microtype}
\usepackage{hyperref}
\usepackage{url}
\usepackage{booktabs}
\usepackage{graphicx}
\usepackage{xcolor}
\usepackage[table]{xcolor}
\usepackage{tcolorbox}
\usepackage{pgfplots}
\usepackage{pgfplotstable}
\usepackage{subcaption}
\usepackage{multirow}
\usepackage{enumitem}
\usepackage{amsmath}
\usepackage{wrapfig}
\usepackage{fancyvrb, fvextra}

\pgfplotsset{compat=1.18}
\usepgfplotslibrary{groupplots}

\usepackage{lineno}

\definecolor{darkblue}{rgb}{0, 0, 0.5}
\hypersetup{colorlinks=true, citecolor=darkblue, linkcolor=darkblue, urlcolor=darkblue}

\definecolor{clrDCLM}{HTML}{8b8b8b}
\definecolor{clrFinePhrase}{HTML}{EBA937}
\definecolor{clrNemotron}{HTML}{76b900}
\definecolor{clrREWIRE}{HTML}{1877F2}
\definecolor{clrCosmopedia}{HTML}{e15759}
\definecolor{clrSmolLM}{HTML}{4e79a7}
\definecolor{clrGemma}{HTML}{59a14f}
\definecolor{clrQwen}{HTML}{f28e2b}
\definecolor{clrFalcon}{HTML}{b07aa1}
\definecolor{clrGranite}{HTML}{76b7b2}
\definecolor{clrLlama}{HTML}{ff9da7}
\definecolor{clrFWEduHQ}{HTML}{59a14f}
\definecolor{clrFWEduLQ}{HTML}{e15759}
\definecolor{clrSYNTH}{HTML}{b07aa1}

\usepackage{listings}

\definecolor{codegreen}{rgb}{0,0.6,0}
\definecolor{codegray}{rgb}{0.5,0.5,0.5}
\definecolor{codepurple}{rgb}{0.58,0,0.82}
\definecolor{backcolour}{rgb}{0.95,0.95,0.92}

\newcommand{\huggingfacedown}{\includegraphics[height=0.75em]{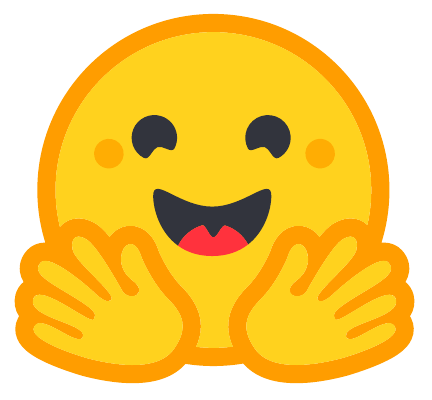}}
\newcommand{\githubdown}{\includegraphics[height=0.75em]{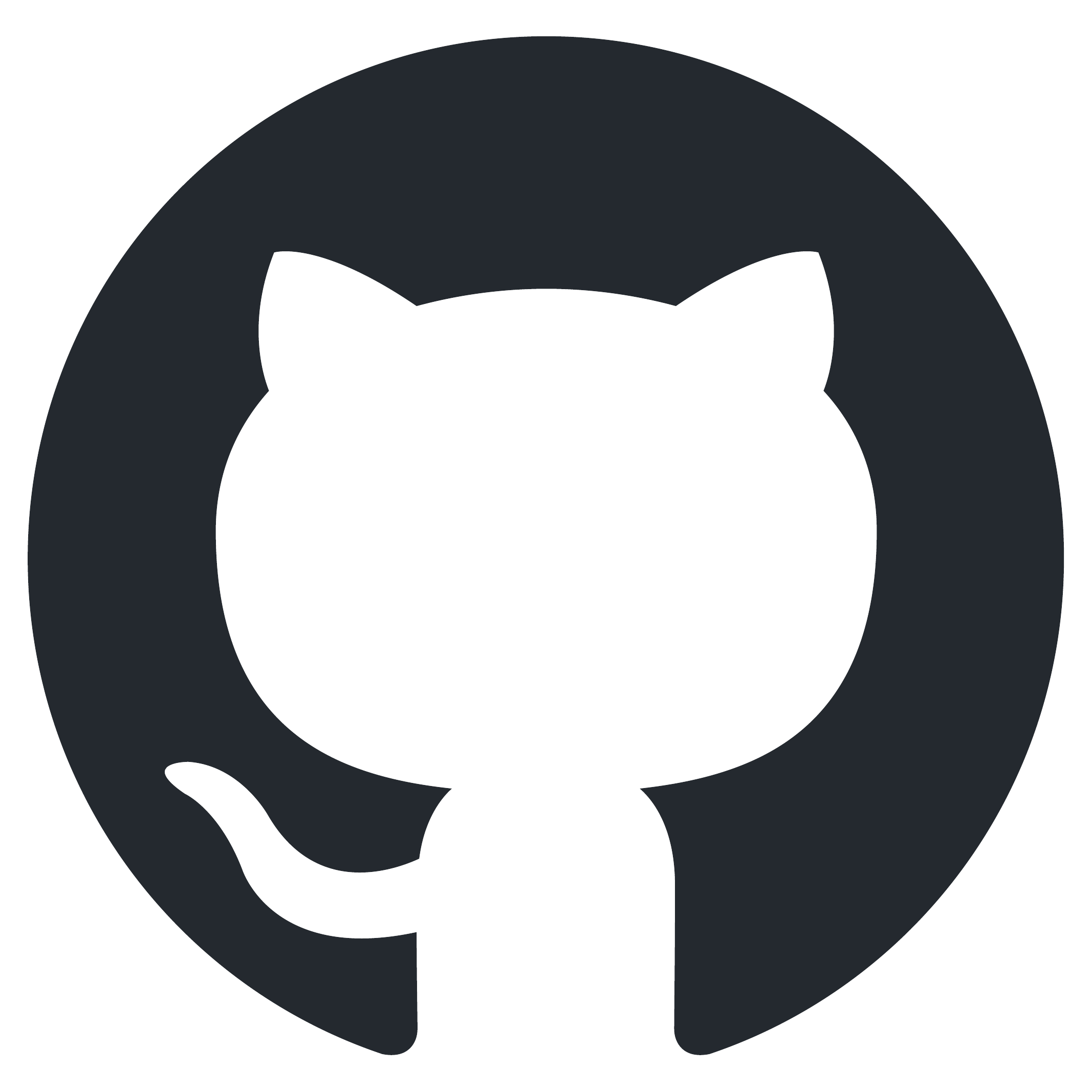}}

\title{How Can We Synthesize High-Quality Pretraining Data?\\A Systematic Study of Prompt Design, Generator Model, and Source Data}

\author{
Joel Niklaus$^{1}$\thanks{Equal contribution} \quad Atsuki Yamaguchi$^{2}$\footnotemark[1] \quad Michal Štefánik$^{3}$ \quad Guilherme Penedo$^{1}$\\
\textbf{Hynek Kydlíček}$^{1}$ \quad \textbf{Elie Bakouch}$^{1}$ \quad \textbf{Lewis Tunstall}$^{1}$ \quad \textbf{Edward Emanuel Beeching}$^{1}$\\
\textbf{Thibaud Frere}$^{1}$ \quad \textbf{Colin Raffel}$^{1}$ \quad \textbf{Leandro von Werra}$^{1}$ \quad \textbf{Thomas Wolf}$^{1}$\\
$^{1}$Hugging Face
\quad$^{2}$University of Sheffield, United Kingdom\\
$^{3}$National Institute of Informatics, Japan\\
\\
\begin{minipage}{0.8\textwidth}
    \hangindent=3em
    \hangafter=1
    \quad \quad \quad \githubdown~Pipeline code: \url{https://github.com/huggingface/finephrase}
\end{minipage}
\\[5pt]
\quad \quad \quad\huggingfacedown~\textsc{FinePhrase} dataset: \url{https://hf.co/datasets/HuggingFaceFW/finephrase}
\vspace{-2em}
}

\begin{document}

\ifcolmsubmission
\linenumbers
\fi

\maketitle

\begin{abstract}
Synthetic data is a standard component in training large language models, yet systematic comparisons across design dimensions, including rephrasing strategy, generator model, and source data, remain absent.
We conduct extensive controlled experiments, generating over one trillion tokens, to identify critical factors in rephrasing web text into synthetic pretraining data.
Our results reveal that structured output formats, such as tables, math problems, FAQs, and tutorials, consistently outperform both curated web baselines and prior synthetic methods.
Notably, increasing the size of the generator model beyond 1B parameters provides no additional benefit.
Our analysis also demonstrates that the selection of the original data used for mixing substantially influences performance.
By applying our findings, we develop \textbf{\textsc{FinePhrase}}, a 486-billion-token open dataset of rephrased web text.
We show that \textsc{FinePhrase} outperforms all existing synthetic data baselines while reducing generation costs by up to 30 times.
We provide the dataset, all prompts, and the generation framework to the research community.\looseness=-1
\end{abstract}

\section{Introduction}
\label{sec:intro}
The pretraining data landscape of large language models (LLMs) has undergone three paradigm shifts.
First, models like GPT~\citep{radford2018improving} and BERT~\citep{devlin-etal-2019-bert} relied on curated corpora including BookCorpus~\citep{7410368} and Wikipedia.
Scaling initiatives like C4~\citep{c4} and The Pile~\citep{thepile} later expanded available datasets into hundreds of gigabytes of web text.
This expansion followed scaling laws, which indicate that increased pretraining duration and data volume consistently improve downstream performance~\citep{kaplan2020scalinglawsneurallanguage,NEURIPS2022_c1e2faff}.
Efforts such as FineWeb~\citep{fineweb} and DCLM~\citep{datacomp} further brought pretraining data to the trillion-token scale, covering most of the crawlable web.
As the volume of available web text plateaus, the focus of the field has moved from quantity to quality: modern development pipelines now employ neural classifiers to filter for educational content~\citep{gunasekar2023textbooksneed,NEURIPS2023_fa3ed726} and analyze data repetition extensively~\citep{NEURIPS2023_9d89448b}.

Synthetic data constitutes the latest step in this progression, appearing as a solution to the exhaustion of usable real-world data.
Rather than merely filtering web text, recent work uses LLMs to rephrase text into variants that preserve information while optimizing the presentation~\citep[\textit{inter alia.}]{wrap,kiyomaru-etal-2026-scaling}.
The community has adopted this approach at scale: \citet{nemotroncc} rephrased two trillion tokens for Nemotron-CC, and several recent LLMs utilized hundreds of billions of synthetic tokens during pretraining~\citep{phi4,nemotron3,qwen3,singh2026arceetrinitylargetechnical}.

However, the design space for synthetic data generation is expansive and currently lacks comprehensive assessment.
Prior work has explored individual approaches in isolation.
For instance, WRAP~\citep{wrap} focuses on stylistic rewriting; Nemotron-CC~\citep{nemotroncc} extracts QA pairs and knowledge lists; BeyondWeb~\citep{beyondweb} proposes continuation and summarization rephrasing; and REWIRE~\citep{rewire} targets the guided transformation of low-quality documents.
While each study reports gains relative to a specific baseline, no systematic comparison exists across these methods.
Furthermore, the interaction between key design choices, such as the choice of generator models and types of source data, remains unexplored, leaving researchers without a clear strategy to maximize the utility of synthetic data for pretraining.

To address these limitations, we conduct a controlled analysis across three axes: \textbf{the rephrasing strategy}, \textbf{the choice of generator models}, and \textbf{the selection of source and mix-in data}. First, we investigate rephrasing strategy by comparing established rephrasing prompts from prior work against four new pedagogical structured formats: tutorials, FAQs, tables, and math reformulations.
These formats leverage the benefits of structured signals and educational content for the learning process of the language model~\citep{gunasekar2023textbooksneed,GalkeLukas2024Dnna,cheng-etal-2024-instruction,yamaguchi-etal-2026-enhancing}.
Second, to determine the influence of generator models, we test six model families with sizes ranging from 135M to 27B parameters.
Finally, we evaluate how the quality of the source text and the integration of original web data (i.e., mix-in data) affect outcomes. This clarifies whether the source quality or the mix-in dataset selection serves as the primary driver of performance.

Our main contributions are as follows:
\begin{itemize}[nosep,leftmargin=*]
    \item We present a systematic comparison of synthetic pretraining data across the strategy of rephrasing, the generator model, and the source data (\S\ref{sec:experiments}).
    \item We show that the choice of rephrasing prompt is the dominant factor for downstream performance; large models (10B+) are not always necessary to perform effective rephrasing; the inherent quality of the source data establishes a strong basis, though source quality has minimal impact when paired with robust mix-in data; and output diversity is more important than formatting consistency (\S\ref{sec:experiments} \& \S\ref{sec:analyses}).
    \item We create and release \textbf{\textsc{FinePhrase}}, a 486B token dataset and associated generation framework. Using SmolLM2 1.7B~\citep{smollm2}, it outperforms all existing synthetic data baselines at up to a 30-fold reduction in generation cost relative to previous methods (\S\ref{sec:finephrase}).\looseness=-1
\end{itemize}

\section{A Systematic Framework for Pretraining Data Synthesis}
\label{sec:method}

\begin{wrapfigure}{r}{0.39\textwidth}
  \vspace{-3em}
  \centering
  \includegraphics[width=0.4\textwidth]{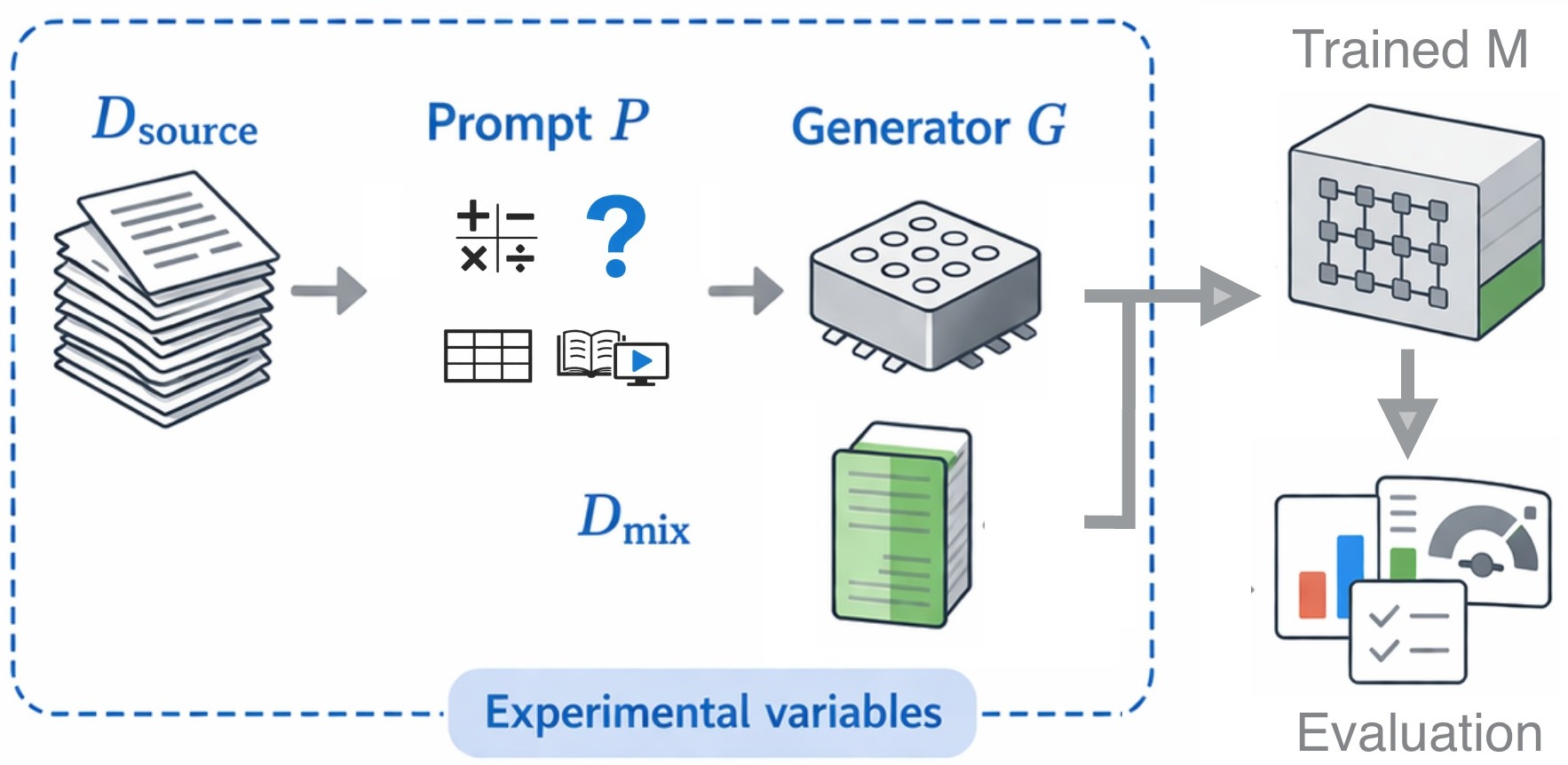}
  \caption{Overview of our experimental methodology.}
  \label{fig:overview}
  \vspace{-1em}
\end{wrapfigure}

\subsection{Problem Setting}
To address the scaling limits of crawlable web data, we systematically evaluate methods for generating synthetic pretraining data by rephrasing web corpora.
Unlike prior work evaluating isolated rephrasing methodologies~\citep[\textit{inter alia.}]{wrap,nemotroncc,rewire}, we conduct a controlled ablation of the key components within the generation pipeline (Figure~\ref{fig:overview}).
The goal is a reproducible experimental framework for synthetic data generation that reduces reliance on intuition and replaces ad hoc practice with a rigorously designed pipeline.
By isolating prompt design, model scale, and data composition, this research identifies configurations that maximize performance while minimizing compute allocation.\looseness=-1

We construct the tested pretraining configurations from a rephrasing prompt $\mathcal{P}$ (\S\ref{subsec:rephrasing}), an instruction-tuned generator model $\mathcal{G}$ (\S\ref{subsec:generator}), and both a mix-in  $\mathcal{D}_{\text{mix}}$ and a source dataset $\mathcal{D}_{\text{source}}$ (\S\ref{subsec:src_mixin_data}).
We validate the effectiveness of each configuration by training an LM $\mathcal{M}$ from scratch on a mixture of rephrased samples and $\mathcal{D}_{\text{mix}}$ for 21B tokens.
Success is defined by the aggregate performance of the trained model across evaluation benchmarks spanning factual knowledge, reading comprehension, and reasoning (\S\ref{subsec:evaluation}).
This framework enables the identification of the optimal generation configuration, which we employ to create \textsc{FinePhrase}, a synthetic dataset consisting of 486B tokens.

\subsection{Rephrasing Strategy}
\label{subsec:rephrasing}
The first axis of investigation focuses on the strategy of rephrasing $\mathcal{P}$, which defines the structural transformation of raw web documents into synthetic pretraining samples.
We investigate rephrasing strategies by comparing established prompts from prior work against four new structured pedagogical formats.

\paragraph{Established Prompts.}
We evaluate eight existing rephrasing strategies that vary by design and objective.
The analysis incorporates five prompts from the Nemotron-HQ-Synth framework~\citep{nemotroncc}:
(i) \textit{Diverse QA Pairs} generates questions across various cognitive levels with accompanying answers;
(ii) \textit{Extract Knowledge} restructures factual content into passages resembling textbook style;
(iii) \textit{Distill} produces a condensed version that retains essential information and technical terminology;
(iv) \textit{Wikipedia} paraphrases the text using formal English characteristic of Wikipedia; and
(v) \textit{Knowledge List} organizes key information into a list.
Furthermore, we include the \textit{Guided Rewrite} prompt from REWIRE~\citep{rewire}. This approach employs an expert persona to perform meta-reasoning and planning before rewriting content from the voice of the original author.
Finally, we examine two baseline strategies from BeyondWeb~\citep{beyondweb}: \textit{Continue}, which generates a text extension matching the existing style, and \textit{Summarize}, which creates a standalone summary of the document.
We include all baseline prompt templates in Appendix \ref{appendix:prompt}.

\paragraph{Pedagogical Structured Prompts.}

\begin{table}[t]
\centering
\footnotesize
\renewcommand{\arraystretch}{0.7}
\setlength{\aboverulesep}{1pt}
\setlength{\belowrulesep}{1pt}
\caption{Excerpts of prompt templates for structured pedagogical rephrasing.
Full templates are available in Appendix \ref{appendix:structured_prompts}.}
\label{tab:prompt_templates}
\resizebox{0.98\linewidth}{!}{
\begin{tabular}{lp{15cm}}
\toprule
\textbf{Format} & \textbf{Instructional Content Example} \\ \midrule
\texttt{math} & ``Rewrite the document to create a mathematical word problem... Provide a step-by-step solution that shows the calculation process clearly.''\\
\texttt{faq}  & ``Rewrite the document as a comprehensive FAQ... Order questions logically from foundational to advanced... Ensure the FAQ works as a standalone document.''\\
\texttt{table} & ``Rewrite the document as a structured table that organizes the key information, then generate one question-answer pair based on the table.''\\
\texttt{tutorial} & ``Rewrite the document as a clear, step-by-step tutorial... Use numbered steps or bullet points where appropriate to enhance clarity.''
 \\\bottomrule
\end{tabular}
}
\end{table}

We evaluate four formats in Table \ref{tab:prompt_templates} to determine if transforming content into pedagogical frameworks enhances the utility of the resulting tokens.
We define a pedagogical format as a data structure that organizes information into discrete, logical, and queryable units.
Specifically, we consider four formats: \texttt{math}, \texttt{faq}, \texttt{table}, and \texttt{tutorial}. This approach follows the observation of \citet{gunasekar2023textbooksneed} that educational clarity drives model performance.
By utilizing these formats, the synthesis process converts flat web text into structured signals.
Prior work suggests these signals often benefit language model pretraining~\citep{GalkeLukas2024Dnna,cheng-etal-2024-instruction,yamaguchi-etal-2026-enhancing}.
We keep the generator model $\mathcal{G}$ and the source dataset $\mathcal{D}_\text{source}$ constant when studying prompt choices to ensure that we have isolated the influence of the prompt.\looseness=-1

\subsection{Generator Model}
\label{subsec:generator}
Our second axis investigates how the generator model $\mathcal{G}$ influences synthetic data characteristics.
We define this axis through two dimensions: parameter scale and architectural family.\looseness=-1

\paragraph{Parameter Scale.}
We evaluate the influence of model size on the quality of the rephrased data.
To isolate the effect of scale, we utilize Gemma 3~\citep{gemma3} in five sizes (270M, 1B, 4B, 12B, 27B) due to its uniquely granular, publicly available open-weights size availability, and SmolLM2 in three sizes (135M, 360M, 1.7B), which presents a complete scaling suite for this family.
This range identifies the minimum capability threshold for synthetic data generation as well as the saturation point where additional model capacity fails to improve the utility of synthetic tokens.\looseness=-1

\paragraph{Architectural Families.}
The underlying architecture provides a second dimension for evaluating generator influence.
We examine six distinct open-source model families at a fixed scale of approximately 1B parameters: Gemma 3, Llama 3.2~\citep{llama3}, Qwen 3~\citep{qwen3}, Granite 3.1~\citep{granite3}, Falcon 3~\citep{falcon3}, and SmolLM2.
By evaluating models across these diverse families, we can assess the sensitivity of synthetic output to specific architectural variations.

\subsection{Mix-in and Source Data}
\label{subsec:src_mixin_data}

Our third axis explores the influence of data composition on pretraining success.
This analysis focuses on the interaction between the source dataset $\mathcal{D}_{\text{source}}$, which provides the content for rephrasing, and the mix-in dataset $\mathcal{D}_{\text{mix}}$, which consists of the original web tokens used during pretraining.

\paragraph{Necessity of Mixing.}
We first evaluate whether training exclusively on synthetic data suffices to achieve optimal performance or whether original web tokens remain indispensable.

\paragraph{Selection of Mix-in Data.}
We compare several candidates for the mix-in component, including DCLM, Cosmopedia, and two tiers (high/low quality) of FineWeb~\citep{fineweb}.
The objective is to identify which original corpus provides the most effective complementary signal to synthetic data.

\paragraph{Source Data Quality and Up-cycling.}
We apply identical rephrasing strategies to the same four source datasets: DCLM, Cosmopedia~\citep{cosmopedia}, and the two FineWeb subsets.
By fixing the mix-in dataset while varying the quality of $\mathcal{D}_{\text{source}}$, we assess whether the synthesis process can up-cycle noisy web text into high-utility training tokens.

\section{Experimental Setup}
\label{sec:setup}

\paragraph{Model Architecture and Tokenizer.}
Across all experiments, we train models with a 1.2B-parameter Qwen 2 architecture~\citep{qwen2} featuring 28 layers, a hidden dimension of 2048, an intermediate size of 6144, and attention mechanisms with 16 query heads and 8 key-value heads through grouped-query attention~\citep{gqa}.
For tokenization, we employ the Llama 3.2 tokenizer with a vocabulary size of 128,256.

\paragraph{Data.}
Each experiment involves rephrasing approximately 10.5 billion tokens.
If the completion count falls below the target because of the specific prompt, the process incorporates repeated samples to ensure a consistent volume across configurations.
The default setup employs a 50/50 mixture of synthetic and original tokens, utilizing Gemma 3 1B as the generator $\mathcal{G}$ and a high-quality subset of FineWeb (FWHQ) for both the source $\mathcal{D}_{\text{source}}$ and the mix-in $\mathcal{D}_{\text{mix}}$.
The FineWeb-Edu classifier scores samples from 0 to 5 via Llama-3-70B-Instruct based on their educational quality.
We regard scores of 4 or 5 as HQ and scores of 0 or 1 as low-quality (LQ) to evaluate the influence of the quality of the seed data.

We select FWHQ as both the default source and mix-in dataset to maintain strict distributional control across our prompt and scale ablations, isolating potential synthesis effects from cross-corpus distribution shifts.
Furthermore, all external synthetic baselines (i.e., rephrasing prompts) evaluated in our pipeline are subjected to these exact same token budgets, source distributions, and 50/50 FWHQ mix-in constraints.

\paragraph{Hyperparameters.}
We pretrain all models using 64 NVIDIA H100 GPUs.
The global batch size is 512 samples with a sequence length of 4096 tokens.
Each model is trained for 10,000 steps, totaling approximately 21 billion tokens.
We use the AdamW optimizer~\citep{adamw} with a learning rate of $5\times10^{-4}$, $\beta_1=0.9$, $\beta_2=0.95$, a weight decay of 0.1, and gradient clipping at 1.0.
All runs leverage bfloat16 precision, Flash Attention 2~\citep{flashattention2}, and rotary position embeddings~\citep{rope}.

\paragraph{Evaluation.}
\label{subsec:evaluation}
Following \citet{kydlicek2025finepdfs}, we assess performance across 12 benchmarks organized into 6 categories. We employ 3-shot cloze-format prompting and report macro-averaged scores:
\textbf{General Knowledge:} ARC (Easy) \citep{arc}, MMLU Redux \citep{mmluredux};
\textbf{Reading Comprehension:} SQuAD v2 \citep{squad2}, DROP \citep{drop};
\textbf{Reasoning:} OpenBookQA \citep{openbookqa}, XCSQA \citep{xcsqa};
\textbf{Natural Language Understanding:} WinoGrande \citep{winogrande}, PIQA \citep{piqa}, HellaSwag \citep{hellaswag};
\textbf{Math:} GSM8K \citep{gsm8k}; and 
\textbf{Table Understanding:} WikiTableQuestions \citep{wikitablequestions}, TriviaQA \citep{triviaqa}.
We include the full breakdown of results for every individual task in Appendix \ref{appendix:results}.

\section{Results}
\label{sec:experiments}

\subsection{Rephrasing Strategy}
\label{subsec:results_rephrase}

\begin{wraptable}{r}{0.35\textwidth}
    \vspace{-6em}
    \tiny
    \centering
    \caption{Macro-averaged scores ($\times 100$) for different existing datasets at 10K steps. Best and second best in \textbf{bold} and \underline{underline}. $\star$ indicates a synthetic dataset. Full results are in Table \ref{tab:existing_data_appendix}.
    }
    \label{tab:existing_data}
    \renewcommand{\arraystretch}{0.8}
    \setlength{\aboverulesep}{1pt}
    \setlength{\belowrulesep}{1pt}
    \resizebox{\linewidth}{!}{
    \begin{tabular}{@{}c@{\hspace{3pt}}l@{\hspace{3pt}}c@{}}
        \toprule
        & \textbf{Dataset} & \textbf{Macro Avg} \\ [-1pt]
        \midrule
        \multirow{4}{*}{\rotatebox{90}{\tiny \resizebox{0.15\linewidth}{!}{\shortstack{Non-\\synthetic}}}} & DCLM & \textbf{13.77} \\
        & Ultra-FineWeb & 13.00 \\
        & FineWeb-HQ & 11.82 \\
        & FineWeb-LQ & 8.83 \\ [-1pt]
        \midrule
        \multirow{4}{*}{\rotatebox{90}{\resizebox{0.15\linewidth}{!}{Synthetic}}} & Nemotron-HQ-Synth$\star$ & \underline{13.54} \\
        & REWIRE$\star$ & 13.49 \\
        & Cosmopedia$\star$ & 10.33 \\
        & SYNTH$\star$ & 10.03 \\ [-1pt]
        \bottomrule
    \end{tabular}
    }
    \vspace{-2em}
\end{wraptable}

\paragraph{Preliminary Investigation: Existing Datasets.}
We first evaluate existing pretraining datasets to establish a performance baseline.
This evaluation considers four popular curated web corpora: DCLM, Ultra-FineWeb~\citep{ultrafineweb}, and the high/low-quality FineWeb subsets; and four synthetic datasets: Nemotron-HQ-Synth, REWIRE, Cosmopedia, and SYNTH~\citep{synthpleias}.
As indicated in Table \ref{tab:existing_data}, DCLM achieves the highest macro-average performance of 13.77 and serves as the primary baseline for subsequent comparisons.

The synthetic datasets Nemotron-HQ-Synth and REWIRE yield results comparable to the performance of DCLM, with a negligible delta of 0.28 points.
However, because Nemotron-HQ-Synth consists of five distinct strategies (\S\ref{subsec:rephrasing}), the aggregate score may mask the high performance of specific transformations while incorporating the deficiencies of others.
This lack of granularity necessitates the systematic ablation of these components, which we present in the following analysis (Table \ref{tab:baselines}).

\begin{wraptable}{r}{0.35\textwidth}
    \vspace{-1.3em}
    \centering
    \tiny
    \caption{Macro-averaged scores for different existing rephrasing approaches. Full results are in Table \ref{tab:baselines_appendix}.}
    \label{tab:baselines}
    \renewcommand{\arraystretch}{0.8}
    \setlength{\aboverulesep}{1pt}
    \setlength{\belowrulesep}{1pt}
    \resizebox{\linewidth}{!}{
    \begin{tabular}{@{}c@{\hspace{3pt}}lc}
        \toprule
        & \textbf{Prompt} & \textbf{Macro Avg} \\[-1pt]
        \midrule
        & \cellcolor{gray!20}DCLM & \cellcolor{gray!20}13.77 \\[-1pt]
        \midrule
        \multirow{5}{*}{\rotatebox{90}{\resizebox{0.2\linewidth}{!}{Nemotron-CC}}} & Diverse QA Pairs & \textbf{14.58}\\
        & Distill & 13.17\\
        & Wikipedia & 13.14\\
        & Knowledge List & 13.11\\
        & Extract Knowledge & 11.81\\ [-1pt]
        \midrule
        \resizebox{0.08\linewidth}{!}{REWIRE} & Guided Rewrite & 13.72\\ [-1pt]
        \midrule
        \multirow{2}{*}{\resizebox{0.08\linewidth}{!}{\shortstack{Beyond\\Web}}} & Summarize & 13.01\\
        & Continue & \underline{13.73}\\ [-1pt]
        \bottomrule
    \end{tabular}
    }
    \vspace{-2.5em}
\end{wraptable}

\paragraph{Established Prompts.}
We now examine the existing rephrasing prompts described in \S\ref{subsec:rephrasing} to identify the most effective transformation.
Table \ref{tab:baselines} illustrates that among established strategies, only \textit{Diverse QA Pairs}, \textit{Continue}, and \textit{Guided Rewrite} rival the DCLM baseline.

The success of \textit{Diverse QA Pairs} is noteworthy.
It achieves a score of 14.58, representing a substantial improvement over 13.77 for DCLM.
This result suggests that the conversion of raw text into a structured query-response format creates a more effective training signal than simple restatement (e.g., \textit{Distill} and \textit{Wikipedia}). 
This indeed supports the hypothesis that structured pedagogical prompts enhance the utility of the resulting tokens (\S\ref{subsec:rephrasing}).
In contrast, strategies such as \textit{Extract Knowledge} or \textit{Knowledge List} show a substantial performance deficit, with drops ranging from 0.6 to 1.96.
This divergence indicates that most rephrasing methodologies provide negligible utility beyond curated web text.

\begin{wraptable}{r}{0.34\textwidth}
    \centering
    \tiny
    \vspace{-3em}
    \caption{New pedagogical structured prompts vs. DCLM. Full results are in Table \ref{tab:new-prompts_appenendix}.}
    \label{tab:new-prompts}
    \renewcommand{\arraystretch}{0.8}
    \setlength{\aboverulesep}{1pt}
    \setlength{\belowrulesep}{1pt}
    \resizebox{0.8\linewidth}{!}{
    \begin{tabular}{lc}
        \toprule
        \textbf{Prompt} & \textbf{Macro Avg}\\[-1pt]
        \midrule
        \rowcolor{gray!20} DCLM & 13.77\\[-1pt]
        \midrule
        \texttt{math} & 15.31\\
        \texttt{faq} & 14.45\\
        \texttt{table} & 14.83\\
        \texttt{tutorial} & 14.30\\[-1pt]
        \bottomrule
    \end{tabular}
    }
    \vspace{-1em}
\end{wraptable}

\paragraph{Pedagogical Structured Prompts.}
We evaluate four pedagogically structured prompts: \texttt{math}, \texttt{faq}, \texttt{table}, and \texttt{tutorial}.
As shown in Table \ref{tab:new-prompts}, all four formats consistently surpass the DCLM baseline: \texttt{math} (+1.54), \texttt{table} (+1.06), \texttt{faq} (+0.68), and \texttt{tutorial} (+0.53).
Notably, the \texttt{math} and \texttt{table} formats also exceed the performance of the most effective existing prompt, \textit{Diverse QA Pairs}, by margins of 0.73 and 0.25 points, respectively.
These results confirm that the restructuring of source data into pedagogically rich formats provides greater utility than simple paraphrasing.
We examine the specific task domains where these formats improve performance in \S\ref{subsec:tradeoff}.

\subsection{Generator Model}
\label{subsec:results_model}

\begin{wrapfigure}{r}{0.31\textwidth}
\vspace{-3.8em}
\centering
\includegraphics[width=\linewidth]{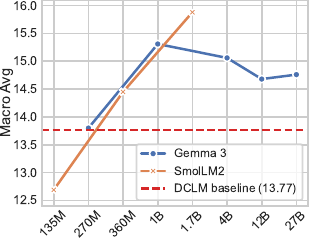}
\caption{Macro-averaged scores for different Gemma 3 (using \texttt{math}) and SmolLM2 (using \texttt{tutorial}) model scales.
Full results are in Table \ref{tab:model_scale_appendix}.
}
\vspace{-1.5em}
\label{tab:model_scale}
\end{wrapfigure}

\paragraph{Parameter Scale.}
To evaluate each generator family under its optimal configuration, we pair them with the best-performing prompt format identified during preliminary sweeps: the \texttt{math} prompt for Gemma 3 and the \texttt{tutorial} prompt for SmolLM2.
Figure \ref{tab:model_scale} compares the impact of model size on the quality of rephrased data.
We observe a clear performance saturation point at the 1B parameter scale for the Gemma 3 family.
The 1B model (15.31) outperforms the 27B variant (14.76), suggesting diminishing returns from increased model capacity.
This trend persists within the SmolLM2 family using the \texttt{tutorial} prompt, which yields macro-average scores of 12.69 for 135M, 14.45 for 360M, and 15.88 for 1.7B.

\begin{wraptable}{r}{0.5\textwidth}
    \centering
    \tiny
    \caption{Macro-averaged scores across student model scales (500M to 6.2B) and Gemma 3-based generator model scales (270M to 27B) trained for 21B tokens on a 50/50 FWHQ mixture using the \texttt{math} prompt.}
    \label{tab:student_generator_scaling}
    \renewcommand{\arraystretch}{0.6}
    \setlength{\aboverulesep}{0.5pt}
    \setlength{\belowrulesep}{0.6pt}
    \resizebox{\linewidth}{!}{
\begin{tabular}{@{}l@{\hskip 5pt}c@{\hskip 5pt}c@{\hskip 5pt}c@{\hskip 5pt}c@{\hskip 5pt}c@{}}
\toprule
& \multicolumn{5}{c}{\textbf{Generator Scale}} \\
\cmidrule(lr){2-6}
\textbf{Student Scale} & \textbf{270M} & \textbf{1B} & \textbf{4B} & \textbf{12B} & \textbf{27B} \\
\midrule
\textbf{500M} & 10.96 & 12.14 & 11.50 & 11.49 & 11.50 \\
\textbf{1.2B} & 13.80 & 15.31 & 15.06 & 14.68 & 14.76 \\
\textbf{2.9B} & 13.80 & 15.97 & 14.75 & 16.40 & 15.72 \\
\textbf{6.2B} & 14.39 & 16.85 & 16.69 & 17.06 & 16.69 \\
\bottomrule
\end{tabular}
}
\end{wraptable}

To ensure these generator scaling trends are not restricted to our proxy 1.2B student model, we additionally evaluate pretraining student models scaled across 500M, 1.2B, 2.9B, and 6.2B parameters (Table \ref{tab:student_generator_scaling}).
While absolute downstream performance naturally increases with student capacity, the relative impact of generator scale remains consistent across all student model sizes.
First, a 270M generator appears insufficient regardless of student scale.
Second, performance saturates at the $\sim$1B generator threshold: for instance, when training a 6.2B student model, upgrading the generator from 270M to 1B yields a substantial +2.46 point macro score jump, whereas scaling the generator further to 12B provides only a marginal +0.21 improvement, and both the 4B and 27B generators underperform the 1B baseline.
Consequently, these multi-scale student experiments confirm that effective synthesis requires a generator of at least 1B parameters, but scaling to 12B or 27B is not justified by pretraining gains.

\begin{wraptable}{r}{0.27\textwidth}
\vspace{-1.5em}
\centering
\tiny
\caption{Comparison of model families across the four new prompts (all within the 1B to 1.7B parameter order of magnitude).
Full results are in Table \ref{tab:model-family_appendix}.
}
\label{tab:model-family}
\renewcommand{\arraystretch}{0.8}
\setlength{\aboverulesep}{1pt}
\setlength{\belowrulesep}{1pt}
\resizebox{\linewidth}{!}{
\begin{tabular}{lc}
\toprule
\textbf{Family} & \textbf{Macro Avg}\\[-1pt]
\midrule
\rowcolor{gray!20} DCLM & 13.77\\[-1pt]
\midrule
SmolLM2 & 16.55\\
Falcon 3 & 15.54\\
Qwen 3 & 14.49\\
Gemma 3 & 14.72\\
Granite 3.1 & 14.87\\
Llama 3.2 & 14.79\\[-1pt]
\bottomrule
\end{tabular}
}
\vspace{-2em}
\end{wraptable}

An exception to this saturation trend arises with the complex \textit{Guided Rewrite} prompt from REWIRE (Appendix \ref{appendix:prompt_rewire}). 
In this case, the 4B model demonstrates a moderate gain (14.58) over the 270M and 1B models (13.29 and 13.72).\footnote{The full results are in Table \ref{tab:guided_rewrite} in Appendix \ref{appendix:results}.}
This outcome likely results from the multi-step formatting requirements of the prompt, which may elevate the minimum capability threshold for the generator.
These results indicate that \textit{large models are not always necessary for effective synthesis}; this finding contrasts with the claim of REWIRE that larger models are inherently required for successful rephrasing~\citep{rewire}. 
Furthermore, comparing Gemma 3 1B against 12B across varying levels of source data quality yields heterogeneous results. For the majority of prompts, the 12B model provides no consistent advantage, irrespective of the quality of the source data (Table \ref{tab:varying_quality_scale_prompt} in Appendix \ref{appendix:results}).\looseness=-1

\paragraph{Architectural Families.}
Table \ref{tab:model-family} compares the results of six architectural families described in \S\ref{subsec:generator} averaged across the four pedagogical prompts (\S\ref{subsec:rephrasing}).
The SmolLM2 1.7B model surpasses all other architectural families by margins ranging from 1.01 to 2.06.
This performance advantage is primarily attributed to superior reading comprehension metrics.
Specifically, the performance of SmolLM2 exceeds the average of the other families by at least 5.12 points on SQuAD v2 (Table \ref{tab:model-family_appendix}).
This superior performance likely stems from the specific composition of the instruction-tuning data of SmolLM2, which includes explicit rewrite tasks (i.e., Smol-Rewrite)~\citep{smollm2}.
Consequently, the model possesses a preexisting proficiency for structural transformation prior to the application of the experimental prompts.\footnote{While model sizes in Table \ref{tab:model-family} vary, our analysis includes exact-scale matches (Falcon 3 and Qwen 3 are also 1.7B). The outperformance of SmolLM2 over exact-scale counterparts suggests that post-training data alignment, rather than raw parameter volume alone, can influence generation utility.}\looseness=-1

\subsection{Mix-in and Source Data}
\label{subsec:mix_in_src_results}

\begin{wraptable}{r}{0.39\textwidth}
\vspace{-2em}
\centering
\tiny
\caption{Comparison of synthetic-only and mixed training. Full results are in Table \ref{tab:synth-vs-mixed_appendix}.}
\label{tab:synth-vs-mixed}
\renewcommand{\arraystretch}{0.8}
\setlength{\aboverulesep}{1pt}
\setlength{\belowrulesep}{1pt}
\resizebox{\linewidth}{!}{
\begin{tabular}{lcc}
\toprule
\textbf{Prompt} & \textbf{Synthetic Only} & \textbf{Mixed} \\ [-1pt]
\midrule
\rowcolor{gray!20} DCLM & \multicolumn{2}{c}{13.77} \\ [-1pt]
\midrule
\texttt{math} & 15.20 & \textbf{15.31} \\
\texttt{faq} & 13.12 & \textbf{14.45} \\
\texttt{table} & 13.74 & \textbf{14.83} \\
\texttt{tutorial} & 12.32 & \textbf{14.30} \\ [-1pt]
\bottomrule
\end{tabular}
}
\vspace{-1.5em}
\end{wraptable}

\paragraph{Necessity of Mixing.}
Table \ref{tab:synth-vs-mixed} demonstrates that training exclusively on synthetic data yields suboptimal performance.
Across all evaluated prompts, the integration of synthetic samples with original web text consistently surpasses configurations restricted to synthetic data.
Consistent with the findings of \citet{demystifyingsynth}, these results suggest that while pure synthetic data may not surpass the efficacy of natural text, a mixture can accelerate convergence.
This improvement stems primarily from the restoration of natural language understanding (NLU) capabilities (e.g., WinoGrande, PIQA, and HellaSwag), which pure synthetic data fails to provide in isolation (Table \ref{tab:synth-vs-mixed_appendix}).
Without the inclusion of original web tokens, the model remains susceptible to the risk of model collapse~\citep{modelcollapse}, as the generator fails to replicate the rich linguistic diversity of the crawlable web.
We further discuss the performance trade-offs between synthetic pedagogical data and original web text in \S\ref{subsec:tradeoff}.

\begin{wraptable}{r}{0.4\textwidth}
    \vspace{-1.5em} %
    \centering
    \tiny
    \renewcommand{\arraystretch}{0.8}
    \setlength{\aboverulesep}{1pt}
    \setlength{\belowrulesep}{1pt}

    \begin{minipage}{\linewidth}
        \centering
        \caption{Comparison of different mix-in dataset $\mathcal{D}_\text{mix}$ choices with FWHQ (HQ) or FWLQ (LQ) as $\mathcal{D}_\text{source}$. Full results in Table \ref{tab:mix-in_appendix}.}
        \label{tab:mix-in}
        \resizebox{0.8\linewidth}{!}{
            \begin{tabular}{lcc}
                \toprule
                 & \multicolumn{2}{c}{$\mathcal{D}_\text{source}$} \\
                \textbf{Mix-in ($\mathcal{D}_\text{mix}$)} & \textbf{HQ} & \textbf{LQ} \\
                \midrule
                \rowcolor{gray!20} FineWeb & 11.82 & 8.83\\
                \midrule
                Cosmopedia & 12.99 & 10.83\\
                DCLM & 14.32 & 12.62\\
                FWHQ & 14.30 & 12.99\\
                FWLQ & 13.61 & 9.63\\
                \bottomrule
            \end{tabular}
        }
    \end{minipage}

    \vspace{2em} %

    \begin{minipage}{\linewidth}
        \centering
        \caption{Comparison of different source dataset $\mathcal{D}_\text{source}$ choices. Full results in Table \ref{tab:source_data_appendix}.}
        \label{tab:source_data}
        \resizebox{\linewidth}{!}{
            \begin{tabular}{lcc}
                \toprule
                \textbf{$\mathcal{D}_{\text{source}}$} & \textbf{$\mathcal{D}_{\text{source}} = \mathcal{D}_{\text{mix}}$} & \textbf{$\mathcal{D}_{\text{mix}} = \text{HQ}$} \\
                \midrule
                Cosmopedia & 10.36 & 13.88\\
                DCLM & 13.69 & 14.77\\
                FWHQ & 14.30 & 14.30\\
                FWLQ & 9.63 & 12.99\\
                \bottomrule
            \end{tabular}
        }
    \end{minipage}
    \vspace{-3em}
\end{wraptable}

\paragraph{Selection of Mix-in Data.}

We evaluate the impact of the selection of $\mathcal{D}_{\text{mix}}$ by fixing $\mathcal{D}_{\text{source}}$ to FWHQ or FWLQ and employing the \texttt{tutorial} prompt.
As shown in Table \ref{tab:mix-in}, all configurations incorporating a mix-in dataset surpass the source-only baseline (indicated in gray).
In high-quality source settings, DCLM and FWHQ serve as the most effective mix-in components, outperforming the baseline by 2.50 and 2.48 points, respectively.
Conversely, when rephrasing low-quality data, the inclusion of FWHQ as the mix-in dataset yields slightly superior utility (12.99) compared to DCLM (12.62).
These results establish that the selection of the mix-in dataset is a critical determinant of final performance, with FWHQ and DCLM representing the optimal choices for the mix-in component.

\paragraph{Source Data Quality and Up-cycling.}

We further examine the influence of the quality of $\mathcal{D}_{\text{source}}$ using the \texttt{tutorial} prompt.
When we set $\mathcal{D}_{\text{source}} = \mathcal{D}_{\text{mix}}$, the quality of the source material serves as a primary factor of the final performance of the model (Table~\ref{tab:source_data}).
Specifically, large variance exists between configurations, ranging from 9.63 for FWLQ to 14.3 for FWHQ.
However, a distinct trend emerges when the quality of $\mathcal{D}_{\text{mix}}$ is fixed at a high level: 
the inherent quality of $\mathcal{D}_{\text{source}}$ becomes less dominant.
In this high-quality mix-in setting, the performance delta narrows to a maximum of 1.78 points.
These results demonstrate that the rephrasing of low-quality data yields competitive results if the training process pairs the synthetic samples with a robust mix-in corpus.

Specifically, as shown in Table \ref{tab:source_data}, substituting half of the high-quality training budget with rephrased low-quality tokens ($\mathcal{D}_{\text{source}}=\text{FWLQ}$ with $\mathcal{D}_{\text{mix}}=\text{FWHQ}$, achieving 12.99) actually improves downstream performance by +1.17 points over using 100\% high-quality data alone (FWHQ = 11.82), while outperforming raw FWLQ (8.83) by +4.16 points.
This observation is encouraging: it demonstrates that the rephrasing pipeline is not simply being carried by the high-quality mix-in, but rather successfully up-cycles noisy web text into high-utility training tokens, thereby effectively expanding the reservoir of available source data for pretraining.

\section{Analysis}
\label{sec:analyses}

Having completed our systematic study to determine broad trends, we now dig into finer-grained details of our results to better understand specific factors and broader trends.

\begin{figure}[t]
    \centering
    \includegraphics[width=\linewidth,trim={0 0.5em 0 0},clip]{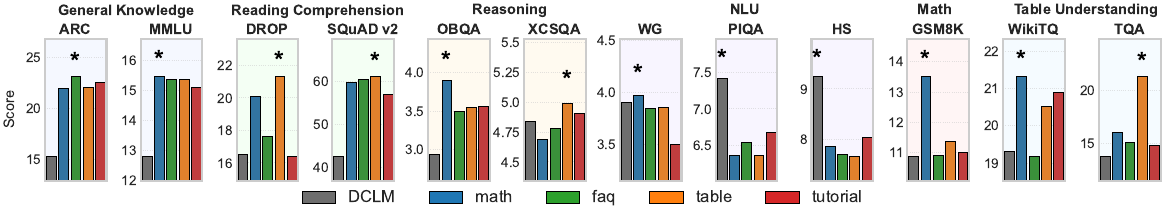}
    \caption{Fine-grained evaluation of pedagogical formats against the DCLM baseline. The best result in each task is marked with $\star$. We use the best-performing SmolLM2 1.7B as $\mathcal{G}$.}
    \label{tab:trade_off}
\end{figure}

\subsection{Performance Trade-off between Synthetic and Web Data}
\label{subsec:tradeoff}

A fine-grained evaluation reveals a performance trade-off between synthetic pedagogical data and original web text.
Figure \ref{tab:trade_off} shows a tension between specialized knowledge and commonsense reasoning, particularly within Natural Language Understanding (NLU). 
Synthetic formats consistently fall behind the DCLM baseline on HellaSwag (HS) and PIQA.
For instance, the \texttt{table} format achieves only 7.66 on HS compared to 9.25 for DCLM.

Nevertheless, these formats achieve higher aggregate scores (\S\ref{subsec:results_rephrase}) because improvements in General Knowledge and Reading Comprehension outweigh these specific deficits.
Specifically, the \texttt{table} format improves SQuAD v2 scores by 18.72 over the baseline, while the \texttt{faq} format increases ARC scores by 7.92.
Crucially, we must separate ``general pretraining data quality'' from ``benchmark-format alignment.''
Structured synthesis yields high-density signals for specialized knowledge and logic by explicitly surfacing reasoning chains and relational data that are otherwise implicit or noisy in source web text. However, it inherently reduces the unstructured linguistic diversity found in natural text.
As discussed in \S\ref{subsec:mix_in_src_results}, original web data acts as a synergistic component that reinstates the performance of the model in NLU; rephrasing cannot replicate this capability in isolation.

\subsection{Qualitative Analysis: Template Collapse and Output Diversity}
\label{subsec:template-collapse}

We evaluate 1,000 outputs from SmolLM2 1.7B and Qwen 3 1.7B using the \texttt{math} prompt to identify the qualitative factors that drive the performance gap between these architectural families.
The analysis focuses on three primary criteria: solution completeness, variance in output length, and structural repetition.

The results reveal a performance paradox: while Qwen 3 achieves 100\% instruction adherence, with all samples containing distinct problem and solution sections, only 68\% of the outputs from SmolLM2 feature complete solutions (see examples in Appendix \ref{appendix:qualitative}).
Despite this inconsistency, SmolLM2 data yields superior downstream performance as observed in \S\ref{subsec:results_model}.
We attribute this to \textit{template collapse}: a phenomenon where the extreme consistency of a generator produces repetitive and uniform training data that saturates the learning signal.
For example, 115 outputs from Qwen 3 begin with identical introductory text.
Conversely, the most frequent starting pattern in the data of SmolLM2 appears only three times.
Furthermore, SmolLM2 shows a wider length variance (4 to 4,000 tokens) compared to the rigid formatting of Qwen 3 (100 to 2,600 tokens).

To confirm this trend across architectural families, we extend our audit quantitatively across 10,000 generations per 1B-class model family in Appendix \ref{appendix:format_adherence_audit}.
We find that Qwen 3 suffers severe collapse (7,619/10k outputs share the exact prefix \texttt{Problem:}), while SmolLM2 maintains 1,897 distinct openings.
Furthermore, surface adherence does not guarantee arithmetic accuracy or topic relevance. 
Semantic Vendi scores~\citep{friedman2023the} confirm a negative correlation ($\rho = -0.29, p < 0.01$) between rigid formatting consistency and downstream performance.
These findings establish that structural and linguistic diversity during pretraining outweighs strict format adherence.

\begin{figure}[t]
    \centering
    \resizebox{0.9\textwidth}{!}{
    \begin{tikzpicture}
    \begin{axis}[
        width=\linewidth,
        height=7.5cm, 
        xlabel={GPU hours (log scale)},
        ylabel={Macro Avg},
        xmode=log,
        xmin=150, xmax=16000,
        ymin=12, ymax=18,
        grid=major,
        grid style={dashed, gray!20},
        legend style={font=\tiny, at={(1.02,1)}, anchor=north west, cells={anchor=west}},
        point meta=explicit symbolic,
        nodes near coords,
        every node near coord/.append style={
            font=\footnotesize, 
            anchor=west, 
            xshift=1pt,
            opacity=0.7
        },
        xlabel style={font=\small},
        ylabel style={font=\small},
    ]

    \addlegendimage{empty legend} \addlegendentry{\textbf{Prompts:}}
    \addlegendimage{empty legend} \addlegendentry{m: \texttt{math}, f: \texttt{faq},}
    \addlegendimage{empty legend} \addlegendentry{
    t: \texttt{table}, u: \texttt{tutorial}}
    \addlegendimage{empty legend} \addlegendentry{
    i: Distill, w: Wikipedia}
    \addlegendimage{empty legend} \addlegendentry{
    r: Guided Rewrite}
    \addlegendimage{empty legend} \addlegendentry{
    d: Diverse QA Pairs}
    \addlegendimage{empty legend} \addlegendentry{
    k: Knowledge List}
    \addlegendimage{empty legend} \addlegendentry{}
    
    \addplot[only marks, mark=diamond, mark size=1.6pt, clrSmolLM]
        coordinates {(537.1,12.68) [u]};
    \addlegendentry{SmolLM2 135M}
    
    \addplot[only marks, mark=triangle, mark size=1.6pt, clrSmolLM]
        coordinates {(657.2,14.45) [u]};
    \addlegendentry{SmolLM2 360M}

    \addplot[only marks, mark=*, mark size=1.6pt, clrSmolLM]
        coordinates {(435.4,16.18) [f] (313.8,16.97) [m] (204.6,17.17) [t] (696.2,15.87) [u]};
    \addlegendentry{SmolLM2 1.7B}

    \addplot[only marks, mark=o, mark size=1.5pt, clrGemma]
        coordinates {(387.8,13.79) [m] (485.1,13.36) [u] (249.3,13.29) [r]};
    \addlegendentry{Gemma 3 270M}

    \addplot[only marks, mark=square*, mark size=1.5pt, clrGemma]
        coordinates {
            (5784.6,14.45) [f] (931.5,15.31) [m] (790.1,14.82) [t] (531.1,14.29) [u]
            (802.9,13.17) [i] (925.5,14.58) [d] (1200.8,13.11) [k] (1276.2,13.14) [w]
            (2435.6,13.72) [r]
        };
    \addlegendentry{Gemma 3 1B}

    \addplot[only marks, mark=diamond*, mark size=2pt, clrGemma!70!black]
        coordinates {(270.5,15.06) [m] (503.0,14.44) [u] (267.1,14.58) [r]};
    \addlegendentry{Gemma 3 4B}

    \addplot[only marks, mark=triangle*, mark size=2pt, clrGemma!40!black]
        coordinates {(2369.7,13.86) [f] (1727.1,14.68) [m] (597.6,14.90) [t] (3383.2,14.22) [u]};
    \addlegendentry{Gemma 3 12B}

    \addplot[only marks, mark=pentagon*, mark size=2pt, black]
        coordinates {(1595.7,14.76) [m] (3841.1,14.70) [u] (11116.4,14.23) [r]};
    \addlegendentry{Gemma 3 27B}

    \addplot[only marks, mark=star, mark size=1.8pt, clrFalcon]
        coordinates {(946.3,14.76) [f] (397.5,15.93) [m] (357.9,16.44) [t] (1435.7,15.01) [u]};
    \addlegendentry{Falcon 3 1B}

    \addplot[only marks, mark=x, mark size=1.8pt, clrLlama]
        coordinates {(2057.1,15.33) [f] (297.5,14.94) [m] (327.9,14.84) [t] (760.5,14.04) [u]};
    \addlegendentry{Llama 3.2 1B}

    \addplot[only marks, mark=+, mark size=1.8pt, clrGranite]
        coordinates {(776.4,14.29) [f] (497.0,15.39) [m] (533.1,15.71) [t] (1083.8,14.06) [u]};
    \addlegendentry{Granite 3.1 1B}

    \addplot[only marks, mark=asterisk, mark size=1.8pt, clrQwen]
        coordinates {(975.9,13.74) [f] (224.4,14.38) [m] (343.5,15.34) [t] (1127.6,14.49) [u]};
    \addlegendentry{Qwen 3 1.7B}

    \node[pin={[align=center]270:{\footnotesize \textbf{Cheapest}\\\textbf{run}}}] at (axis cs:204.6,17.17) {};
    \node[pin={[align=center]270:{\footnotesize \textbf{Most}\\ 
    \footnotesize \textbf{expensive}\\ \footnotesize \textbf{run}}}] at (axis cs:11116.4,14.23) {};

    \end{axis}
    \end{tikzpicture}
    }
    \caption{GPU cost vs. performance. Symbols distinguish model variants; adjacent characters represent prompt formats (e.g., \textbf{t}: table, \textbf{m}: math).}
    \label{fig:cost}
\end{figure}
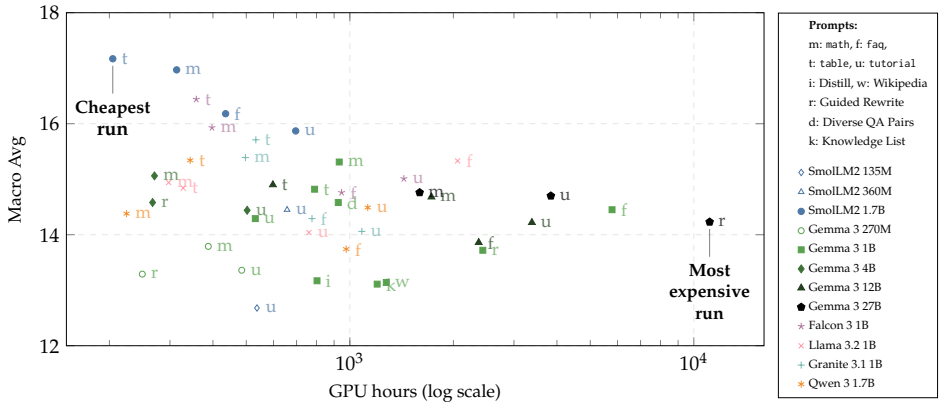

\subsection{Cost-Benefit Analysis of Model Scale}
\label{subsec:cost}

The most efficient iteration (utilizing \texttt{table} with SmolLM2 1.7B) necessitates 8 GPU-days, whereas the most resource-intensive (utilizing \textit{Guided Rewrite} with Gemma 3 27B) consumes over 15 GPU-months (Figure \ref{fig:cost}).
The Pareto frontier is dominated by small models employing the four structured prompts described in \S\ref{subsec:results_rephrase}. 
Specifically, models in the 1B-parameter class, such as Gemma 3 1B and SmolLM2 1.7B, often achieve the optimal cost-to-performance ratio when paired with these structured formats. Scaling the generator to 12B or 27B parameters increases GPU time by 5 to 10 times while simultaneously decreasing performance in most cases.
These findings suggest that researchers should prioritize the design of the rephrasing prompt over the scale of the generator model.
Investing in prompt architecture provides a higher return on compute than simply increasing the parameter count of the generator.\looseness=-1

\section{FinePhrase}
\label{sec:finephrase}
We apply our empirical findings to build \textsc{FinePhrase}, a large-scale synthetic dataset.
The construction of the dataset follows the optimal configuration identified in our analysis: we utilize SmolLM2 1.7B as the generator model (\S\ref{subsec:results_model}) and employ the four pedagogical structured formats: \texttt{math}, \texttt{faq}, \texttt{table}, and \texttt{tutorial} (\S\ref{subsec:results_rephrase}), given the cost-benefit analysis in \S\ref{subsec:cost}.\looseness=-1

\paragraph{Scale and Efficiency.}

\begin{wraptable}{r}{0.5\textwidth}
    \vspace{-1em}
    \centering
    \tiny
    \caption{Generation cost comparison (all H100 GPU-hours).}
    \label{tab:cost-comparison}
    \renewcommand{\arraystretch}{0.8}
    \setlength{\aboverulesep}{1pt}
    \setlength{\belowrulesep}{1pt}
    \resizebox{\linewidth}{!}{
    \begin{tabular}{@{}l@{\hskip 5pt}l@{\hskip 2pt}r@{\hskip 4pt}r@{\hskip 4pt}r@{}}
        \toprule
        \textbf{Dataset} & \textbf{Generator} & \textbf{Tokens} & \textbf{GPU-hrs} & \textbf{Tok/GPU-hr} \\ [-1pt]
        \midrule
        Cosmopedia & Mixtral 8x7B & 25B & $>$10K & $<$2.5M \\
        SYNTH & fine-tuned & 80B & 4K & 20M \\
        REWIRE & Llama 3.3 70B & 400B & $\sim$352K & $\sim$1.1M \\
        \midrule
        \textsc{FinePhrase} & SmolLM2 1.7B & 486B & $\sim$14.7K & $\sim$33.1M \\ [-1pt]
        \bottomrule
    \end{tabular}
    }
    \vspace{-3em}
\end{wraptable}

\textsc{FinePhrase} rephrases documents from FineWeb into four structured formats, yielding 1.35 billion samples and 486 billion completion tokens.\footnote{We include some actual samples in Appendix \ref{appendix:samples}.}
By using 100 NVIDIA H100 GPUs and SmolLM2 1.7B with suffix-32 speculative decoding, the generation pipeline achieves a throughput of approximately 9,200 tokens per second per GPU.
Consequently, the full generation necessitates only 612 GPU-days (approximately 14,700 GPU-hours).
As shown in Table~\ref{tab:cost-comparison}, this throughput represents a 30$\times$ efficiency gain over the process used for REWIRE (which employs Llama 3.3 70B) and a 13$\times$ increase over the methodology of Cosmopedia (which utilizes Mixtral 8x7B).
\textsc{FinePhrase} generates a higher volume of tokens than REWIRE while consuming 24$\times$ less total compute.
This demonstrates that structured rephrasing with compact models provides a step-change in generation throughput.\looseness=-1

\begin{figure}[t]
    \centering
    \includegraphics[width=\linewidth,trim={0 0.4em 0 0},clip]{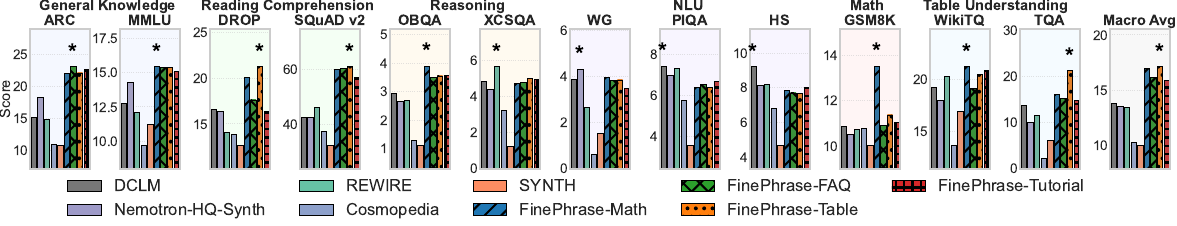}
    \caption{
        \textsc{FinePhrase} prompts vs. baselines. The best result in each task is marked with $\star$.
        All \textsc{FinePhrase} runs use SmolLM2 1.7B on FineWeb, mixed with FineWeb-HQ.
    }
    \label{tab:finephrase}
\end{figure}

\paragraph{Results.}
Figure \ref{tab:finephrase} demonstrates that every \textsc{FinePhrase} variant substantially surpasses the established benchmarks in terms of macro-average scores.
\textsc{FinePhrase}-Table achieves a peak macro-average of 17.18, representing an improvement of 3.41 points over DCLM and 3.63 points over Nemotron-HQ-Synth.
The breakdown results validate the trade-off discussed in \S\ref{subsec:tradeoff}: logically dense \textsc{FinePhrase} formats dominate factual knowledge and reading comprehension (ARC, DROP, SQuAD v2), whereas the DCLM retain a nominal advantage in NLU (PIQA, HellaSwag).
This performance profile validates the design of \textsc{FinePhrase} as a synergistic component intended for integration with original web data.

\section{Conclusions}
\label{sec:conclusions}
We evaluated synthetic pretraining across rephrasing strategies, generator models, and source data.
Accordingly, we found that:
the design of rephrasing prompts primarily determines performance;
structured pedagogical formats exceed the utility of simple paraphrasing;
generator capacities beyond 1B parameters yield negligible gains for most rephrasing prompts; source quality has minimal impact when paired with robust mix-in data; and generative diversity outweighs formatting consistency.
While synthetic tokens provide logical depth, original web data remains essential to preserve commonsense reasoning and linguistic variety.
These principles culminate in \textsc{FinePhrase}, a 486-billion-token dataset that achieves up to a 30$\times$ cost reduction while outperforming established baselines.

\paragraph{Discussion and Limitations.}
While our primary experiments evaluate a proxy 1.2B student model trained on 21B tokens following FineWeb~\citep{fineweb}, our student-scaling experiments (\S\ref{subsec:results_model}) demonstrate that key findings hold up to 6.2B parameters.
A 72-run multi-seed grid varying initialization and data ordering seeds across four student model sizes (Appendix \ref{appendix:multiseed_mixing_ratio}) confirms that run-to-run seed variance ($\pm 0.16$ to $\pm 0.49$) is negligible compared to the synthetic pretraining gain (+4.71 to +5.23), with synthetic mixing actively reducing variance relative to baselines.
Furthermore, sweeping the synthetic token fraction from 10\% to 90\% shows consistent performance gains over baselines (13.8) across the entire range, peaking between 60\% and 80\% synthetic mix (16.8) to confirm our default 50/50 split sits comfortably on a stable performance plateau.
To verify that evaluation improvements are not artifacts of data leakage, our $n$-gram contamination audit (Appendix \ref{appendix:contamination_audit}) confirms that rephrased corpora maintain lower contamination rates than the underlying web source (735 vs. 841 docs/B tokens).
Nonetheless, running full multi-seed confidence intervals across every ablation combination remains computationally prohibitive, evaluation is restricted to standard English benchmarks, and cost comparisons rely on reported figures that may fluctuate across hardware environments.

\section*{Reproducibility Statement}
All training hyperparameters, prompt templates, and evaluation details are provided in \S\ref{sec:setup} and Appendix \ref{appendix:prompt}.
The \textsc{FinePhrase} dataset (486B tokens), the complete ablation data, the underlying generation framework, the code for generating the \textsc{FinePhrase} dataset, and the full suite of benchmark results are available at \huggingfacedown~\url{https://hf.co/datasets/HuggingFaceFW/finephrase}, \huggingfacedown~\url{https://hf.co/buckets/HuggingFaceFW/finephrase-rephrased}, \githubdown~\url{https://github.com/huggingface/finephrase}, \githubdown~\url{https://github.com/huggingface/datatrove/blob/main/examples/inference/finephrase.py}, and \huggingfacedown~\url{https://huggingface.co/spaces/HuggingFaceFW/finephrase/blob/main/app/src/content/assets/data/benchmark-results.csv}, respectively.

\section*{Ethics Statement}
Synthetic data generation raises concerns regarding the practice of training on model outputs.
We address the model collapse concern directly through a pipeline that integrates synthetic data with original web data, which ensures that training does not occur exclusively on the outputs of a model.
We do not filter for or against any specific topics, and the content distribution of the source data remains consistent with the distribution of FineWeb.
Furthermore, the synthetic data is always obtained by rephrasing seed data, and not by simply sampling many times from a model with a fixed prompt.
The dataset and the code are released openly to facilitate scrutiny and replication by the research community.

\section*{Acknowledgment}
The authors acknowledge the use of generative AI tools in the preparation of this manuscript. Specifically, Gemini 3 Thinking was employed to generate Figure \ref{fig:cost} from raw samples, followed by manual verification by the authors.

\bibliography{colm2026_conference,anthology-1,anthology-2}

\begin{thebibliography}{59}
\providecommand{\natexlab}[1]{#1}
\providecommand{\url}[1]{\texttt{#1}}
\expandafter\ifx\csname urlstyle\endcsname\relax
  \providecommand{\doi}[1]{doi: #1}\else
  \providecommand{\doi}{doi: \begingroup \urlstyle{rm}\Url}\fi

\bibitem[Abdin et~al.(2024)Abdin, Aneja, Behl, Bubeck, Eldan, Gunasekar, Harrison, Hewett, Javaheripi, Kauffmann, Lee, Lee, Li, Liu, Mendes, Nguyen, Price, de~Rosa, Saarikivi, Salim, Shah, Wang, Ward, Wu, Yu, Zhang, and Zhang]{phi4}
Marah Abdin, Jyoti Aneja, Harkirat Behl, Sébastien Bubeck, Ronen Eldan, Suriya Gunasekar, Michael Harrison, Russell~J. Hewett, Mojan Javaheripi, Piero Kauffmann, James~R. Lee, Yin~Tat Lee, Yuanzhi Li, Weishung Liu, Caio C.~T. Mendes, Anh Nguyen, Eric Price, Gustavo de~Rosa, Olli Saarikivi, Adil Salim, Shital Shah, Xin Wang, Rachel Ward, Yue Wu, Dingli Yu, Cyril Zhang, and Yi~Zhang.
\newblock Phi-4 technical report.
\newblock \emph{arXiv preprint}, arXiv:2412.08905, 2024.
\newblock URL \url{https://arxiv.org/abs/2412.08905}.

\bibitem[Ainslie et~al.(2023)Ainslie, Lee-Thorp, de~Jong, Zemlyanskiy, Lebron, and Sanghai]{gqa}
Joshua Ainslie, James Lee-Thorp, Michiel de~Jong, Yury Zemlyanskiy, Federico Lebron, and Sumit Sanghai.
\newblock {GQA}: Training generalized multi-query transformer models from multi-head checkpoints.
\newblock In Houda Bouamor, Juan Pino, and Kalika Bali (eds.), \emph{Proceedings of the 2023 Conference on Empirical Methods in Natural Language Processing}, pp.\  4895--4901, Singapore, December 2023. Association for Computational Linguistics.
\newblock \doi{10.18653/v1/2023.emnlp-main.298}.
\newblock URL \url{https://aclanthology.org/2023.emnlp-main.298/}.

\bibitem[Allal et~al.(2025)Allal, Lozhkov, Bakouch, Blazquez, Penedo, Tunstall, Marafioti, Lajar{\'\i}n, Kydl{\'\i}{\v{c}}ek, Srivastav, Lochner, Fahlgren, NGUYEN, Burtenshaw, Fourrier, Zhao, Larcher, Morlon, Zakka, Raffel, Werra, and Wolf]{smollm2}
Loubna~Ben Allal, Anton Lozhkov, Elie Bakouch, Gabriel~Martin Blazquez, Guilherme Penedo, Lewis Tunstall, Andr{\'e}s Marafioti, Agust{\'\i}n~Piqueres Lajar{\'\i}n, Hynek Kydl{\'\i}{\v{c}}ek, Vaibhav Srivastav, Joshua Lochner, Caleb Fahlgren, Xuan~Son NGUYEN, Ben Burtenshaw, Cl{\'e}mentine Fourrier, Haojun Zhao, Hugo Larcher, Mathieu Morlon, Cyril Zakka, Colin Raffel, Leandro~Von Werra, and Thomas Wolf.
\newblock Smol{LM}2: When smol goes big {\textemdash} data-centric training of a fully open small language model.
\newblock In \emph{Proceedings of the Second Conference on Language Modeling}, 2025.
\newblock URL \url{https://openreview.net/forum?id=3JiCl2A14H}.

\bibitem[Ben~Allal et~al.(2024)Ben~Allal, Lozhkov, Penedo, Wolf, and von Werra]{cosmopedia}
Loubna Ben~Allal, Anton Lozhkov, Guilherme Penedo, Thomas Wolf, and Leandro von Werra.
\newblock Cosmopedia, 2024.
\newblock URL \url{https://huggingface.co/datasets/HuggingFaceTB/cosmopedia}.
\newblock Data repository.

\bibitem[Bisk et~al.(2020)Bisk, Zellers, Le~bras, Gao, and Choi]{piqa}
Yonatan Bisk, Rowan Zellers, Ronan Le~bras, Jianfeng Gao, and Yejin Choi.
\newblock {PIQA}: Reasoning about physical commonsense in natural language.
\newblock \emph{Proceedings of the AAAI Conference on Artificial Intelligence}, 34\penalty0 (05):\penalty0 7432--7439, Apr. 2020.
\newblock \doi{10.1609/aaai.v34i05.6239}.
\newblock URL \url{https://ojs.aaai.org/index.php/AAAI/article/view/6239}.

\bibitem[Cheng et~al.(2024)Cheng, Gu, Huang, Bi, Huang, and Wei]{cheng-etal-2024-instruction}
Daixuan Cheng, Yuxian Gu, Shaohan Huang, Junyu Bi, Minlie Huang, and Furu Wei.
\newblock Instruction pre-training: Language models are supervised multitask learners.
\newblock In Yaser Al-Onaizan, Mohit Bansal, and Yun-Nung Chen (eds.), \emph{Proceedings of the 2024 Conference on Empirical Methods in Natural Language Processing}, pp.\  2529--2550, Miami, Florida, USA, November 2024. Association for Computational Linguistics.
\newblock \doi{10.18653/v1/2024.emnlp-main.148}.
\newblock URL \url{https://aclanthology.org/2024.emnlp-main.148/}.

\bibitem[Clark et~al.(2018)Clark, Cowhey, Etzioni, Khot, Sabharwal, Schoenick, and Tafjord]{arc}
Peter Clark, Isaac Cowhey, Oren Etzioni, Tushar Khot, Ashish Sabharwal, Carissa Schoenick, and Oyvind Tafjord.
\newblock Think you have solved question answering? try {ARC}, the {AI2} reasoning challenge.
\newblock \emph{arXiv preprint}, arXiv:1803.05457, 2018.
\newblock URL \url{https://arxiv.org/abs/1803.05457}.

\bibitem[Cobbe et~al.(2021)Cobbe, Kosaraju, Bavarian, Chen, Jun, Kaiser, Plappert, Tworek, Hilton, Nakano, Hesse, and Schulman]{gsm8k}
Karl Cobbe, Vineet Kosaraju, Mohammad Bavarian, Mark Chen, Heewoo Jun, Lukasz Kaiser, Matthias Plappert, Jerry Tworek, Jacob Hilton, Reiichiro Nakano, Christopher Hesse, and John Schulman.
\newblock Training verifiers to solve math word problems.
\newblock \emph{arXiv preprint}, arXiv:2110.14168, 2021.
\newblock URL \url{https://arxiv.org/abs/2110.14168}.

\bibitem[Dao(2024)]{flashattention2}
Tri Dao.
\newblock {FlashAttention-2}: Faster attention with better parallelism and work partitioning.
\newblock In \emph{Proceedings of the Twelfth International Conference on Learning Representations}, 2024.
\newblock URL \url{https://openreview.net/forum?id=mZn2Xyh9Ec}.

\bibitem[DatologyAI et~al.(2025)DatologyAI, Maini, Dorna, Doshi, Carranza, Pan, Urbanek, Burstein, Fang, Deng, Abbas, Larsen, Blakeney, Bannur, Baek, Teh, Schwab, Mongstad, Yin, Wills, Mentzer, Merrick, Monti, Adiga, Joshi, Das, Wang, Gaza, Morcos, and Leavitt]{beyondweb}
DatologyAI, Pratyush Maini, Vineeth Dorna, Parth Doshi, Aldo Carranza, Fan Pan, Jack Urbanek, Paul Burstein, Alex Fang, Alvin Deng, Amro Abbas, Brett Larsen, Cody Blakeney, Charvi Bannur, Christina Baek, Darren Teh, David Schwab, Haakon Mongstad, Haoli Yin, Josh Wills, Kaleigh Mentzer, Luke Merrick, Ricardo Monti, Rishabh Adiga, Siddharth Joshi, Spandan Das, Zhengping Wang, Bogdan Gaza, Ari Morcos, and Matthew Leavitt.
\newblock {BeyondWeb}: Lessons from scaling synthetic data for trillion-scale pretraining.
\newblock \emph{arXiv preprint}, arXiv:2508.10975, 2025.
\newblock URL \url{https://arxiv.org/abs/2508.10975}.

\bibitem[Devlin et~al.(2019)Devlin, Chang, Lee, and Toutanova]{devlin-etal-2019-bert}
Jacob Devlin, Ming-Wei Chang, Kenton Lee, and Kristina Toutanova.
\newblock {BERT}: Pre-training of deep bidirectional transformers for language understanding.
\newblock In Jill Burstein, Christy Doran, and Thamar Solorio (eds.), \emph{Proceedings of the 2019 Conference of the North {A}merican Chapter of the Association for Computational Linguistics: Human Language Technologies, Volume 1 (Long and Short Papers)}, pp.\  4171--4186, Minneapolis, Minnesota, June 2019. Association for Computational Linguistics.
\newblock \doi{10.18653/v1/N19-1423}.
\newblock URL \url{https://aclanthology.org/N19-1423/}.

\bibitem[Dua et~al.(2019)Dua, Wang, Dasigi, Stanovsky, Singh, and Gardner]{drop}
Dheeru Dua, Yizhong Wang, Pradeep Dasigi, Gabriel Stanovsky, Sameer Singh, and Matt Gardner.
\newblock {DROP}: A reading comprehension benchmark requiring discrete reasoning over paragraphs.
\newblock In Jill Burstein, Christy Doran, and Thamar Solorio (eds.), \emph{Proceedings of the 2019 Conference of the North {A}merican Chapter of the Association for Computational Linguistics: Human Language Technologies, Volume 1 (Long and Short Papers)}, pp.\  2368--2378, Minneapolis, Minnesota, June 2019. Association for Computational Linguistics.
\newblock \doi{10.18653/v1/N19-1246}.
\newblock URL \url{https://aclanthology.org/N19-1246/}.

\bibitem[{Falcon-LLM Team}(2024)]{falcon3}
{Falcon-LLM Team}.
\newblock The {Falcon} 3 family of open models, 2024.
\newblock URL \url{https://huggingface.co/blog/falcon3}.
\newblock Blog post.

\bibitem[Friedman \& Dieng(2023)Friedman and Dieng]{friedman2023the}
Dan Friedman and Adji~Bousso Dieng.
\newblock The vendi score: A diversity evaluation metric for machine learning.
\newblock \emph{Transactions on Machine Learning Research}, 2023.
\newblock ISSN 2835-8856.
\newblock URL \url{https://openreview.net/forum?id=g97OHbQyk1}.

\bibitem[Galke et~al.(2024)Galke, Ram, and Raviv]{GalkeLukas2024Dnna}
Lukas Galke, Yoav Ram, and Limor Raviv.
\newblock Deep neural networks and humans both benefit from compositional language structure.
\newblock \emph{Nature Communications}, 15\penalty0 (10816), 2024.
\newblock ISSN 2041-1723.

\bibitem[Gao et~al.(2020)Gao, Biderman, Black, Golding, Hoppe, Foster, Phang, He, Thite, Nabeshima, Presser, and Leahy]{thepile}
Leo Gao, Stella Biderman, Sid Black, Laurence Golding, Travis Hoppe, Charles Foster, Jason Phang, Horace He, Anish Thite, Noa Nabeshima, Shawn Presser, and Connor Leahy.
\newblock The {Pile}: An 800{GB} dataset of diverse text for language modeling.
\newblock \emph{arXiv preprint}, arXiv:2101.00027, 2020.
\newblock URL \url{https://arxiv.org/abs/2101.00027}.

\bibitem[Gema et~al.(2025)Gema, Leang, Hong, Devoto, Mancino, Saxena, He, Zhao, Du, Ghasemi~Madani, Barale, McHardy, Harris, Kaddour, Van~Krieken, and Minervini]{mmluredux}
Aryo~Pradipta Gema, Joshua Ong~Jun Leang, Giwon Hong, Alessio Devoto, Alberto Carlo~Maria Mancino, Rohit Saxena, Xuanli He, Yu~Zhao, Xiaotang Du, Mohammad~Reza Ghasemi~Madani, Claire Barale, Robert McHardy, Joshua Harris, Jean Kaddour, Emile Van~Krieken, and Pasquale Minervini.
\newblock Are we done with {MMLU}?
\newblock In Luis Chiruzzo, Alan Ritter, and Lu~Wang (eds.), \emph{Proceedings of the 2025 Conference of the Nations of the Americas Chapter of the Association for Computational Linguistics: Human Language Technologies (Volume 1: Long Papers)}, pp.\  5069--5096, Albuquerque, New Mexico, April 2025. Association for Computational Linguistics.
\newblock ISBN 979-8-89176-189-6.
\newblock \doi{10.18653/v1/2025.naacl-long.262}.
\newblock URL \url{https://aclanthology.org/2025.naacl-long.262/}.

\bibitem[{Gemma Team} et~al.(2025){Gemma Team}, Kamath, Ferret, Pathak, Vieillard, Merhej, Perrin, Matejovicova, Ramé, Rivière, Rouillard, Mesnard, Cideron, bastien Grill, Ramos, Yvinec, Casbon, Pot, Penchev, Liu, Visin, Kenealy, Beyer, Zhai, Tsitsulin, Busa-Fekete, Feng, Sachdeva, Coleman, Gao, Mustafa, Barr, Parisotto, Tian, Eyal, Cherry, Peter, Sinopalnikov, Bhupatiraju, Agarwal, Kazemi, Malkin, Kumar, Vilar, Brusilovsky, Luo, Steiner, Friesen, Sharma, Sharma, Gilady, Goedeckemeyer, Saade, Feng, Kolesnikov, Bendebury, Abdagic, Vadi, György, Pinto, Das, Bapna, Miech, Yang, Paterson, Shenoy, Chakrabarti, Piot, Wu, Shahriari, Petrini, Chen, Lan, Choquette-Choo, Carey, Brick, Deutsch, Eisenbud, Cattle, Cheng, Paparas, Sreepathihalli, Reid, Tran, Zelle, Noland, Huizenga, Kharitonov, Liu, Amirkhanyan, Cameron, Hashemi, Klimczak-Plucińska, Singh, Mehta, Lehri, Hazimeh, Ballantyne, Szpektor, Nardini, Pouget-Abadie, Chan, Stanton, Wieting, Lai, Orbay, Fernandez, Newlan, yeong Ji, Singh, Black, Yu, Hui,
  Vodrahalli, Greff, Qiu, Valentine, Coelho, Ritter, Hoffman, Watson, Chaturvedi, Moynihan, Ma, Babar, Noy, Byrd, Roy, Momchev, Chauhan, Sachdeva, Bunyan, Botarda, Caron, Rubenstein, Culliton, Schmid, Sessa, Xu, Stanczyk, Tafti, Shivanna, Wu, Pan, Rokni, Willoughby, Vallu, Mullins, Jerome, Smoot, Girgin, Iqbal, Reddy, Sheth, Põder, Bhatnagar, Panyam, Eiger, Zhang, Liu, Yacovone, Liechty, Kalra, Evci, Misra, Roseberry, Feinberg, Kolesnikov, Han, Kwon, Chen, Chow, Zhu, Wei, Egyed, Cotruta, Giang, Kirk, Rao, Black, Babar, Lo, Moreira, Martins, Sanseviero, Gonzalez, Gleicher, Warkentin, Mirrokni, Senter, Collins, Barral, Ghahramani, Hadsell, Matias, Sculley, Petrov, Fiedel, Shazeer, Vinyals, Dean, Hassabis, Kavukcuoglu, Farabet, Buchatskaya, Alayrac, Anil, Dmitry, Lepikhin, Borgeaud, Bachem, Joulin, Andreev, Hardin, Dadashi, and Hussenot]{gemma3}
{Gemma Team}, Aishwarya Kamath, Johan Ferret, Shreya Pathak, Nino Vieillard, Ramona Merhej, Sarah Perrin, Tatiana Matejovicova, Alexandre Ramé, Morgane Rivière, Louis Rouillard, Thomas Mesnard, Geoffrey Cideron, Jean bastien Grill, Sabela Ramos, Edouard Yvinec, Michelle Casbon, Etienne Pot, Ivo Penchev, Gaël Liu, Francesco Visin, Kathleen Kenealy, Lucas Beyer, Xiaohai Zhai, Anton Tsitsulin, Robert Busa-Fekete, Alex Feng, Noveen Sachdeva, Benjamin Coleman, Yi~Gao, Basil Mustafa, Iain Barr, Emilio Parisotto, David Tian, Matan Eyal, Colin Cherry, Jan-Thorsten Peter, Danila Sinopalnikov, Surya Bhupatiraju, Rishabh Agarwal, Mehran Kazemi, Dan Malkin, Ravin Kumar, David Vilar, Idan Brusilovsky, Jiaming Luo, Andreas Steiner, Abe Friesen, Abhanshu Sharma, Abheesht Sharma, Adi~Mayrav Gilady, Adrian Goedeckemeyer, Alaa Saade, Alex Feng, Alexander Kolesnikov, Alexei Bendebury, Alvin Abdagic, Amit Vadi, András György, André~Susano Pinto, Anil Das, Ankur Bapna, Antoine Miech, Antoine Yang, Antonia Paterson, Ashish
  Shenoy, Ayan Chakrabarti, Bilal Piot, Bo~Wu, Bobak Shahriari, Bryce Petrini, Charlie Chen, Charline~Le Lan, Christopher~A. Choquette-Choo, CJ~Carey, Cormac Brick, Daniel Deutsch, Danielle Eisenbud, Dee Cattle, Derek Cheng, Dimitris Paparas, Divyashree~Shivakumar Sreepathihalli, Doug Reid, Dustin Tran, Dustin Zelle, Eric Noland, Erwin Huizenga, Eugene Kharitonov, Frederick Liu, Gagik Amirkhanyan, Glenn Cameron, Hadi Hashemi, Hanna Klimczak-Plucińska, Harman Singh, Harsh Mehta, Harshal~Tushar Lehri, Hussein Hazimeh, Ian Ballantyne, Idan Szpektor, Ivan Nardini, Jean Pouget-Abadie, Jetha Chan, Joe Stanton, John Wieting, Jonathan Lai, Jordi Orbay, Joseph Fernandez, Josh Newlan, Ju~yeong Ji, Jyotinder Singh, Kat Black, Kathy Yu, Kevin Hui, Kiran Vodrahalli, Klaus Greff, Linhai Qiu, Marcella Valentine, Marina Coelho, Marvin Ritter, Matt Hoffman, Matthew Watson, Mayank Chaturvedi, Michael Moynihan, Min Ma, Nabila Babar, Natasha Noy, Nathan Byrd, Nick Roy, Nikola Momchev, Nilay Chauhan, Noveen Sachdeva, Oskar
  Bunyan, Pankil Botarda, Paul Caron, Paul~Kishan Rubenstein, Phil Culliton, Philipp Schmid, Pier~Giuseppe Sessa, Pingmei Xu, Piotr Stanczyk, Pouya Tafti, Rakesh Shivanna, Renjie Wu, Renke Pan, Reza Rokni, Rob Willoughby, Rohith Vallu, Ryan Mullins, Sammy Jerome, Sara Smoot, Sertan Girgin, Shariq Iqbal, Shashir Reddy, Shruti Sheth, Siim Põder, Sijal Bhatnagar, Sindhu~Raghuram Panyam, Sivan Eiger, Susan Zhang, Tianqi Liu, Trevor Yacovone, Tyler Liechty, Uday Kalra, Utku Evci, Vedant Misra, Vincent Roseberry, Vlad Feinberg, Vlad Kolesnikov, Woohyun Han, Woosuk Kwon, Xi~Chen, Yinlam Chow, Yuvein Zhu, Zichuan Wei, Zoltan Egyed, Victor Cotruta, Minh Giang, Phoebe Kirk, Anand Rao, Kat Black, Nabila Babar, Jessica Lo, Erica Moreira, Luiz~Gustavo Martins, Omar Sanseviero, Lucas Gonzalez, Zach Gleicher, Tris Warkentin, Vahab Mirrokni, Evan Senter, Eli Collins, Joelle Barral, Zoubin Ghahramani, Raia Hadsell, Yossi Matias, D.~Sculley, Slav Petrov, Noah Fiedel, Noam Shazeer, Oriol Vinyals, Jeff Dean, Demis Hassabis,
  Koray Kavukcuoglu, Clement Farabet, Elena Buchatskaya, Jean-Baptiste Alayrac, Rohan Anil, Dmitry, Lepikhin, Sebastian Borgeaud, Olivier Bachem, Armand Joulin, Alek Andreev, Cassidy Hardin, Robert Dadashi, and Léonard Hussenot.
\newblock Gemma 3 technical report.
\newblock \emph{arXiv preprint}, arXiv:2503.19786, 2025.
\newblock URL \url{https://arxiv.org/abs/2503.19786}.

\bibitem[Grattafiori et~al.(2024)Grattafiori, Dubey, Jauhri, Pandey, Kadian, Al-Dahle, Letman, Mathur, Schelten, Vaughan, Yang, Fan, Goyal, Hartshorn, Yang, Mitra, Sravankumar, Korenev, Hinsvark, Rao, Zhang, Rodriguez, Gregerson, Spataru, Roziere, Biron, Tang, Chern, Caucheteux, Nayak, Bi, Marra, McConnell, Keller, Touret, Wu, Wong, Ferrer, Nikolaidis, Allonsius, Song, Pintz, Livshits, Wyatt, Esiobu, Choudhary, Mahajan, Garcia-Olano, Perino, Hupkes, Lakomkin, AlBadawy, Lobanova, Dinan, Smith, Radenovic, Guzmán, Zhang, Synnaeve, Lee, Anderson, Thattai, Nail, Mialon, Pang, Cucurell, Nguyen, Korevaar, Xu, Touvron, Zarov, Ibarra, Kloumann, Misra, Evtimov, Zhang, Copet, Lee, Geffert, Vranes, Park, Mahadeokar, Shah, van~der Linde, Billock, Hong, Lee, Fu, Chi, Huang, Liu, Wang, Yu, Bitton, Spisak, Park, Rocca, Johnstun, Saxe, Jia, Alwala, Prasad, Upasani, Plawiak, Li, Heafield, Stone, El-Arini, Iyer, Malik, Chiu, Bhalla, Lakhotia, Rantala-Yeary, van~der Maaten, Chen, Tan, Jenkins, Martin, Madaan, Malo, Blecher,
  Landzaat, de~Oliveira, Muzzi, Pasupuleti, Singh, Paluri, Kardas, Tsimpoukelli, Oldham, Rita, Pavlova, Kambadur, Lewis, Si, Singh, Hassan, Goyal, Torabi, Bashlykov, Bogoychev, Chatterji, Zhang, Duchenne, Çelebi, Alrassy, Zhang, Li, Vasic, Weng, Bhargava, Dubal, Krishnan, Koura, Xu, He, Dong, Srinivasan, Ganapathy, Calderer, Cabral, Stojnic, Raileanu, Maheswari, Girdhar, Patel, Sauvestre, Polidoro, Sumbaly, Taylor, Silva, Hou, Wang, Hosseini, Chennabasappa, Singh, Bell, Kim, Edunov, Nie, Narang, Raparthy, Shen, Wan, Bhosale, Zhang, Vandenhende, Batra, Whitman, Sootla, Collot, Gururangan, Borodinsky, Herman, Fowler, Sheasha, Georgiou, Scialom, Speckbacher, Mihaylov, Xiao, Karn, Goswami, Gupta, Ramanathan, Kerkez, Gonguet, Do, Vogeti, Albiero, Petrovic, Chu, Xiong, Fu, Meers, Martinet, Wang, Wang, Tan, Xia, Xie, Jia, Wang, Goldschlag, Gaur, Babaei, Wen, Song, Zhang, Li, Mao, Coudert, Yan, Chen, Papakipos, Singh, Srivastava, Jain, Kelsey, Shajnfeld, Gangidi, Victoria, Goldstand, Menon, Sharma, Boesenberg,
  Baevski, Feinstein, Kallet, Sangani, Teo, Yunus, Lupu, Alvarado, Caples, Gu, Ho, Poulton, Ryan, Ramchandani, Dong, Franco, Goyal, Saraf, Chowdhury, Gabriel, Bharambe, Eisenman, Yazdan, James, Maurer, Leonhardi, Huang, Loyd, Paola, Paranjape, Liu, Wu, Ni, Hancock, Wasti, Spence, Stojkovic, Gamido, Montalvo, Parker, Burton, Mejia, Liu, Wang, Kim, Zhou, Hu, Chu, Cai, Tindal, Feichtenhofer, Gao, Civin, Beaty, Kreymer, Li, Adkins, Xu, Testuggine, David, Parikh, Liskovich, Foss, Wang, Le, Holland, Dowling, Jamil, Montgomery, Presani, Hahn, Wood, Le, Brinkman, Arcaute, Dunbar, Smothers, Sun, Kreuk, Tian, Kokkinos, Ozgenel, Caggioni, Kanayet, Seide, Florez, Schwarz, Badeer, Swee, Halpern, Herman, Sizov, Guangyi, Zhang, Lakshminarayanan, Inan, Shojanazeri, Zou, Wang, Zha, Habeeb, Rudolph, Suk, Aspegren, Goldman, Zhan, Damlaj, Molybog, Tufanov, Leontiadis, Veliche, Gat, Weissman, Geboski, Kohli, Lam, Asher, Gaya, Marcus, Tang, Chan, Zhen, Reizenstein, Teboul, Zhong, Jin, Yang, Cummings, Carvill, Shepard, McPhie,
  Torres, Ginsburg, Wang, Wu, U, Saxena, Khandelwal, Zand, Matosich, Veeraraghavan, Michelena, Li, Jagadeesh, Huang, Chawla, Huang, Chen, Garg, A, Silva, Bell, Zhang, Guo, Yu, Moshkovich, Wehrstedt, Khabsa, Avalani, Bhatt, Mankus, Hasson, Lennie, Reso, Groshev, Naumov, Lathi, Keneally, Liu, Seltzer, Valko, Restrepo, Patel, Vyatskov, Samvelyan, Clark, Macey, Wang, Hermoso, Metanat, Rastegari, Bansal, Santhanam, Parks, White, Bawa, Singhal, Egebo, Usunier, Mehta, Laptev, Dong, Cheng, Chernoguz, Hart, Salpekar, Kalinli, Kent, Parekh, Saab, Balaji, Rittner, Bontrager, Roux, Dollar, Zvyagina, Ratanchandani, Yuvraj, Liang, Alao, Rodriguez, Ayub, Murthy, Nayani, Mitra, Parthasarathy, Li, Hogan, Battey, Wang, Howes, Rinott, Mehta, Siby, Bondu, Datta, Chugh, Hunt, Dhillon, Sidorov, Pan, Mahajan, Verma, Yamamoto, Ramaswamy, Lindsay, Lindsay, Feng, Lin, Zha, Patil, Shankar, Zhang, Zhang, Wang, Agarwal, Sajuyigbe, Chintala, Max, Chen, Kehoe, Satterfield, Govindaprasad, Gupta, Deng, Cho, Virk, Subramanian, Choudhury,
  Goldman, Remez, Glaser, Best, Koehler, Robinson, Li, Zhang, Matthews, Chou, Shaked, Vontimitta, Ajayi, Montanez, Mohan, Kumar, Mangla, Ionescu, Poenaru, Mihailescu, Ivanov, Li, Wang, Jiang, Bouaziz, Constable, Tang, Wu, Wang, Wu, Gao, Kleinman, Chen, Hu, Jia, Qi, Li, Zhang, Zhang, Adi, Nam, Yu, Wang, Zhao, Hao, Qian, Li, He, Rait, DeVito, Rosnbrick, Wen, Yang, Zhao, and Ma]{llama3}
Aaron Grattafiori, Abhimanyu Dubey, Abhinav Jauhri, Abhinav Pandey, Abhishek Kadian, Ahmad Al-Dahle, Aiesha Letman, Akhil Mathur, Alan Schelten, Alex Vaughan, Amy Yang, Angela Fan, Anirudh Goyal, Anthony Hartshorn, Aobo Yang, Archi Mitra, Archie Sravankumar, Artem Korenev, Arthur Hinsvark, Arun Rao, Aston Zhang, Aurelien Rodriguez, Austen Gregerson, Ava Spataru, Baptiste Roziere, Bethany Biron, Binh Tang, Bobbie Chern, Charlotte Caucheteux, Chaya Nayak, Chloe Bi, Chris Marra, Chris McConnell, Christian Keller, Christophe Touret, Chunyang Wu, Corinne Wong, Cristian~Canton Ferrer, Cyrus Nikolaidis, Damien Allonsius, Daniel Song, Danielle Pintz, Danny Livshits, Danny Wyatt, David Esiobu, Dhruv Choudhary, Dhruv Mahajan, Diego Garcia-Olano, Diego Perino, Dieuwke Hupkes, Egor Lakomkin, Ehab AlBadawy, Elina Lobanova, Emily Dinan, Eric~Michael Smith, Filip Radenovic, Francisco Guzmán, Frank Zhang, Gabriel Synnaeve, Gabrielle Lee, Georgia~Lewis Anderson, Govind Thattai, Graeme Nail, Gregoire Mialon, Guan Pang,
  Guillem Cucurell, Hailey Nguyen, Hannah Korevaar, Hu~Xu, Hugo Touvron, Iliyan Zarov, Imanol~Arrieta Ibarra, Isabel Kloumann, Ishan Misra, Ivan Evtimov, Jack Zhang, Jade Copet, Jaewon Lee, Jan Geffert, Jana Vranes, Jason Park, Jay Mahadeokar, Jeet Shah, Jelmer van~der Linde, Jennifer Billock, Jenny Hong, Jenya Lee, Jeremy Fu, Jianfeng Chi, Jianyu Huang, Jiawen Liu, Jie Wang, Jiecao Yu, Joanna Bitton, Joe Spisak, Jongsoo Park, Joseph Rocca, Joshua Johnstun, Joshua Saxe, Junteng Jia, Kalyan~Vasuden Alwala, Karthik Prasad, Kartikeya Upasani, Kate Plawiak, Ke~Li, Kenneth Heafield, Kevin Stone, Khalid El-Arini, Krithika Iyer, Kshitiz Malik, Kuenley Chiu, Kunal Bhalla, Kushal Lakhotia, Lauren Rantala-Yeary, Laurens van~der Maaten, Lawrence Chen, Liang Tan, Liz Jenkins, Louis Martin, Lovish Madaan, Lubo Malo, Lukas Blecher, Lukas Landzaat, Luke de~Oliveira, Madeline Muzzi, Mahesh Pasupuleti, Mannat Singh, Manohar Paluri, Marcin Kardas, Maria Tsimpoukelli, Mathew Oldham, Mathieu Rita, Maya Pavlova, Melanie Kambadur,
  Mike Lewis, Min Si, Mitesh~Kumar Singh, Mona Hassan, Naman Goyal, Narjes Torabi, Nikolay Bashlykov, Nikolay Bogoychev, Niladri Chatterji, Ning Zhang, Olivier Duchenne, Onur Çelebi, Patrick Alrassy, Pengchuan Zhang, Pengwei Li, Petar Vasic, Peter Weng, Prajjwal Bhargava, Pratik Dubal, Praveen Krishnan, Punit~Singh Koura, Puxin Xu, Qing He, Qingxiao Dong, Ragavan Srinivasan, Raj Ganapathy, Ramon Calderer, Ricardo~Silveira Cabral, Robert Stojnic, Roberta Raileanu, Rohan Maheswari, Rohit Girdhar, Rohit Patel, Romain Sauvestre, Ronnie Polidoro, Roshan Sumbaly, Ross Taylor, Ruan Silva, Rui Hou, Rui Wang, Saghar Hosseini, Sahana Chennabasappa, Sanjay Singh, Sean Bell, Seohyun~Sonia Kim, Sergey Edunov, Shaoliang Nie, Sharan Narang, Sharath Raparthy, Sheng Shen, Shengye Wan, Shruti Bhosale, Shun Zhang, Simon Vandenhende, Soumya Batra, Spencer Whitman, Sten Sootla, Stephane Collot, Suchin Gururangan, Sydney Borodinsky, Tamar Herman, Tara Fowler, Tarek Sheasha, Thomas Georgiou, Thomas Scialom, Tobias Speckbacher,
  Todor Mihaylov, Tong Xiao, Ujjwal Karn, Vedanuj Goswami, Vibhor Gupta, Vignesh Ramanathan, Viktor Kerkez, Vincent Gonguet, Virginie Do, Vish Vogeti, Vítor Albiero, Vladan Petrovic, Weiwei Chu, Wenhan Xiong, Wenyin Fu, Whitney Meers, Xavier Martinet, Xiaodong Wang, Xiaofang Wang, Xiaoqing~Ellen Tan, Xide Xia, Xinfeng Xie, Xuchao Jia, Xuewei Wang, Yaelle Goldschlag, Yashesh Gaur, Yasmine Babaei, Yi~Wen, Yiwen Song, Yuchen Zhang, Yue Li, Yuning Mao, Zacharie~Delpierre Coudert, Zheng Yan, Zhengxing Chen, Zoe Papakipos, Aaditya Singh, Aayushi Srivastava, Abha Jain, Adam Kelsey, Adam Shajnfeld, Adithya Gangidi, Adolfo Victoria, Ahuva Goldstand, Ajay Menon, Ajay Sharma, Alex Boesenberg, Alexei Baevski, Allie Feinstein, Amanda Kallet, Amit Sangani, Amos Teo, Anam Yunus, Andrei Lupu, Andres Alvarado, Andrew Caples, Andrew Gu, Andrew Ho, Andrew Poulton, Andrew Ryan, Ankit Ramchandani, Annie Dong, Annie Franco, Anuj Goyal, Aparajita Saraf, Arkabandhu Chowdhury, Ashley Gabriel, Ashwin Bharambe, Assaf Eisenman, Azadeh
  Yazdan, Beau James, Ben Maurer, Benjamin Leonhardi, Bernie Huang, Beth Loyd, Beto~De Paola, Bhargavi Paranjape, Bing Liu, Bo~Wu, Boyu Ni, Braden Hancock, Bram Wasti, Brandon Spence, Brani Stojkovic, Brian Gamido, Britt Montalvo, Carl Parker, Carly Burton, Catalina Mejia, Ce~Liu, Changhan Wang, Changkyu Kim, Chao Zhou, Chester Hu, Ching-Hsiang Chu, Chris Cai, Chris Tindal, Christoph Feichtenhofer, Cynthia Gao, Damon Civin, Dana Beaty, Daniel Kreymer, Daniel Li, David Adkins, David Xu, Davide Testuggine, Delia David, Devi Parikh, Diana Liskovich, Didem Foss, Dingkang Wang, Duc Le, Dustin Holland, Edward Dowling, Eissa Jamil, Elaine Montgomery, Eleonora Presani, Emily Hahn, Emily Wood, Eric-Tuan Le, Erik Brinkman, Esteban Arcaute, Evan Dunbar, Evan Smothers, Fei Sun, Felix Kreuk, Feng Tian, Filippos Kokkinos, Firat Ozgenel, Francesco Caggioni, Frank Kanayet, Frank Seide, Gabriela~Medina Florez, Gabriella Schwarz, Gada Badeer, Georgia Swee, Gil Halpern, Grant Herman, Grigory Sizov, Guangyi, Zhang, Guna
  Lakshminarayanan, Hakan Inan, Hamid Shojanazeri, Han Zou, Hannah Wang, Hanwen Zha, Haroun Habeeb, Harrison Rudolph, Helen Suk, Henry Aspegren, Hunter Goldman, Hongyuan Zhan, Ibrahim Damlaj, Igor Molybog, Igor Tufanov, Ilias Leontiadis, Irina-Elena Veliche, Itai Gat, Jake Weissman, James Geboski, James Kohli, Janice Lam, Japhet Asher, Jean-Baptiste Gaya, Jeff Marcus, Jeff Tang, Jennifer Chan, Jenny Zhen, Jeremy Reizenstein, Jeremy Teboul, Jessica Zhong, Jian Jin, Jingyi Yang, Joe Cummings, Jon Carvill, Jon Shepard, Jonathan McPhie, Jonathan Torres, Josh Ginsburg, Junjie Wang, Kai Wu, Kam~Hou U, Karan Saxena, Kartikay Khandelwal, Katayoun Zand, Kathy Matosich, Kaushik Veeraraghavan, Kelly Michelena, Keqian Li, Kiran Jagadeesh, Kun Huang, Kunal Chawla, Kyle Huang, Lailin Chen, Lakshya Garg, Lavender A, Leandro Silva, Lee Bell, Lei Zhang, Liangpeng Guo, Licheng Yu, Liron Moshkovich, Luca Wehrstedt, Madian Khabsa, Manav Avalani, Manish Bhatt, Martynas Mankus, Matan Hasson, Matthew Lennie, Matthias Reso, Maxim
  Groshev, Maxim Naumov, Maya Lathi, Meghan Keneally, Miao Liu, Michael~L. Seltzer, Michal Valko, Michelle Restrepo, Mihir Patel, Mik Vyatskov, Mikayel Samvelyan, Mike Clark, Mike Macey, Mike Wang, Miquel~Jubert Hermoso, Mo~Metanat, Mohammad Rastegari, Munish Bansal, Nandhini Santhanam, Natascha Parks, Natasha White, Navyata Bawa, Nayan Singhal, Nick Egebo, Nicolas Usunier, Nikhil Mehta, Nikolay~Pavlovich Laptev, Ning Dong, Norman Cheng, Oleg Chernoguz, Olivia Hart, Omkar Salpekar, Ozlem Kalinli, Parkin Kent, Parth Parekh, Paul Saab, Pavan Balaji, Pedro Rittner, Philip Bontrager, Pierre Roux, Piotr Dollar, Polina Zvyagina, Prashant Ratanchandani, Pritish Yuvraj, Qian Liang, Rachad Alao, Rachel Rodriguez, Rafi Ayub, Raghotham Murthy, Raghu Nayani, Rahul Mitra, Rangaprabhu Parthasarathy, Raymond Li, Rebekkah Hogan, Robin Battey, Rocky Wang, Russ Howes, Ruty Rinott, Sachin Mehta, Sachin Siby, Sai~Jayesh Bondu, Samyak Datta, Sara Chugh, Sara Hunt, Sargun Dhillon, Sasha Sidorov, Satadru Pan, Saurabh Mahajan,
  Saurabh Verma, Seiji Yamamoto, Sharadh Ramaswamy, Shaun Lindsay, Shaun Lindsay, Sheng Feng, Shenghao Lin, Shengxin~Cindy Zha, Shishir Patil, Shiva Shankar, Shuqiang Zhang, Shuqiang Zhang, Sinong Wang, Sneha Agarwal, Soji Sajuyigbe, Soumith Chintala, Stephanie Max, Stephen Chen, Steve Kehoe, Steve Satterfield, Sudarshan Govindaprasad, Sumit Gupta, Summer Deng, Sungmin Cho, Sunny Virk, Suraj Subramanian, Sy~Choudhury, Sydney Goldman, Tal Remez, Tamar Glaser, Tamara Best, Thilo Koehler, Thomas Robinson, Tianhe Li, Tianjun Zhang, Tim Matthews, Timothy Chou, Tzook Shaked, Varun Vontimitta, Victoria Ajayi, Victoria Montanez, Vijai Mohan, Vinay~Satish Kumar, Vishal Mangla, Vlad Ionescu, Vlad Poenaru, Vlad~Tiberiu Mihailescu, Vladimir Ivanov, Wei Li, Wenchen Wang, Wenwen Jiang, Wes Bouaziz, Will Constable, Xiaocheng Tang, Xiaojian Wu, Xiaolan Wang, Xilun Wu, Xinbo Gao, Yaniv Kleinman, Yanjun Chen, Ye~Hu, Ye~Jia, Ye~Qi, Yenda Li, Yilin Zhang, Ying Zhang, Yossi Adi, Youngjin Nam, Yu, Wang, Yu~Zhao, Yuchen Hao, Yundi
  Qian, Yunlu Li, Yuzi He, Zach Rait, Zachary DeVito, Zef Rosnbrick, Zhaoduo Wen, Zhenyu Yang, Zhiwei Zhao, and Zhiyu Ma.
\newblock The {Llama} 3 herd of models.
\newblock \emph{arXiv preprint}, arXiv:2407.21783, 2024.
\newblock URL \url{https://arxiv.org/abs/2407.21783}.

\bibitem[Gunasekar et~al.(2023)Gunasekar, Zhang, Aneja, Mendes, Giorno, Gopi, Javaheripi, Kauffmann, de~Rosa, Saarikivi, Salim, Shah, Behl, Wang, Bubeck, Eldan, Kalai, Lee, and Li]{gunasekar2023textbooksneed}
Suriya Gunasekar, Yi~Zhang, Jyoti Aneja, Caio César~Teodoro Mendes, Allie~Del Giorno, Sivakanth Gopi, Mojan Javaheripi, Piero Kauffmann, Gustavo de~Rosa, Olli Saarikivi, Adil Salim, Shital Shah, Harkirat~Singh Behl, Xin Wang, Sébastien Bubeck, Ronen Eldan, Adam~Tauman Kalai, Yin~Tat Lee, and Yuanzhi Li.
\newblock Textbooks are all you need.
\newblock \emph{arXiv preprint}, arXiv:2306.11644, 2023.
\newblock URL \url{https://arxiv.org/abs/2306.11644}.

\bibitem[Habib et~al.(2023)Habib, Fourrier, Kydlíček, Wolf, and Tunstall]{lighteval}
Nathan Habib, Clémentine Fourrier, Hynek Kydlíček, Thomas Wolf, and Lewis Tunstall.
\newblock {LightEval}: A lightweight framework for {LLM} evaluation, 2023.
\newblock URL \url{https://github.com/huggingface/lighteval}.
\newblock GitHub repository.

\bibitem[Hoffmann et~al.(2022)Hoffmann, Borgeaud, Mensch, Buchatskaya, Cai, Rutherford, de~Las~Casas, Hendricks, Welbl, Clark, Hennigan, Noland, Millican, van~den Driessche, Damoc, Guy, Osindero, Simonyan, Elsen, Vinyals, Rae, and Sifre]{NEURIPS2022_c1e2faff}
Jordan Hoffmann, Sebastian Borgeaud, Arthur Mensch, Elena Buchatskaya, Trevor Cai, Eliza Rutherford, Diego de~Las~Casas, Lisa~Anne Hendricks, Johannes Welbl, Aidan Clark, Thomas Hennigan, Eric Noland, Katherine Millican, George van~den Driessche, Bogdan Damoc, Aurelia Guy, Simon Osindero, Kar\'{e}n Simonyan, Erich Elsen, Oriol Vinyals, Jack Rae, and Laurent Sifre.
\newblock An empirical analysis of compute-optimal large language model training.
\newblock In S.~Koyejo, S.~Mohamed, A.~Agarwal, D.~Belgrave, K.~Cho, and A.~Oh (eds.), \emph{Advances in Neural Information Processing Systems}, volume~35, pp.\  30016--30030. Curran Associates, Inc., 2022.
\newblock URL \url{https://proceedings.neurips.cc/paper_files/paper/2022/file/c1e2faff6f588870935f114ebe04a3e5-Paper-Conference.pdf}.

\bibitem[{IBM Granite Team}(2024)]{granite3}
{IBM Granite Team}.
\newblock Granite 3.0 language models, 2024.
\newblock URL \url{https://github.com/ibm-granite/granite-3.0-language-models}.
\newblock Technical Report.

\bibitem[Joshi et~al.(2017)Joshi, Choi, Weld, and Zettlemoyer]{triviaqa}
Mandar Joshi, Eunsol Choi, Daniel Weld, and Luke Zettlemoyer.
\newblock {T}rivia{QA}: A large scale distantly supervised challenge dataset for reading comprehension.
\newblock In Regina Barzilay and Min-Yen Kan (eds.), \emph{Proceedings of the 55th Annual Meeting of the Association for Computational Linguistics (Volume 1: Long Papers)}, pp.\  1601--1611, Vancouver, Canada, July 2017. Association for Computational Linguistics.
\newblock \doi{10.18653/v1/P17-1147}.
\newblock URL \url{https://aclanthology.org/P17-1147/}.

\bibitem[Kang et~al.(2025)Kang, Ardalani, Kuchnik, Emad, Elhoushi, Sengupta, Li, Raghavendra, Jia, and Wu]{demystifyingsynth}
Feiyang Kang, Newsha Ardalani, Michael Kuchnik, Youssef Emad, Mostafa Elhoushi, Shubhabrata Sengupta, Shang-Wen Li, Ramya Raghavendra, Ruoxi Jia, and Carole-Jean Wu.
\newblock Demystifying synthetic data in {LLM} pre-training: A systematic study of scaling laws, benefits, and pitfalls.
\newblock In Christos Christodoulopoulos, Tanmoy Chakraborty, Carolyn Rose, and Violet Peng (eds.), \emph{Proceedings of the 2025 Conference on Empirical Methods in Natural Language Processing}, pp.\  10739--10758, Suzhou, China, November 2025. Association for Computational Linguistics.
\newblock ISBN 979-8-89176-332-6.
\newblock \doi{10.18653/v1/2025.emnlp-main.544}.
\newblock URL \url{https://aclanthology.org/2025.emnlp-main.544/}.

\bibitem[Kaplan et~al.(2020)Kaplan, McCandlish, Henighan, Brown, Chess, Child, Gray, Radford, Wu, and Amodei]{kaplan2020scalinglawsneurallanguage}
Jared Kaplan, Sam McCandlish, Tom Henighan, Tom~B. Brown, Benjamin Chess, Rewon Child, Scott Gray, Alec Radford, Jeffrey Wu, and Dario Amodei.
\newblock Scaling laws for neural language models.
\newblock \emph{arXiv preprint}, arXiv:2001.08361, 2020.
\newblock URL \url{https://arxiv.org/abs/2001.08361}.

\bibitem[Kiyomaru et~al.(2026)Kiyomaru, Oda, Kodama, Liu, and Kawahara]{kiyomaru-etal-2026-scaling}
Hirokazu Kiyomaru, Yusuke Oda, Takashi Kodama, Chaoran Liu, and Daisuke Kawahara.
\newblock Scaling data-constrained language models with synthetic data.
\newblock In Vera Demberg, Kentaro Inui, and Llu{\'i}s Marquez (eds.), \emph{Findings of the {A}ssociation for {C}omputational {L}inguistics: {EACL} 2026}, pp.\  1002--1016, Rabat, Morocco, March 2026. Association for Computational Linguistics.
\newblock ISBN 979-8-89176-386-9.
\newblock \doi{10.18653/v1/2026.findings-eacl.52}.
\newblock URL \url{https://aclanthology.org/2026.findings-eacl.52/}.

\bibitem[Kwon et~al.(2023)Kwon, Li, Zhuang, Sheng, Zheng, Yu, Gonzalez, Zhang, and Stoica]{vllm}
Woosuk Kwon, Zhuohan Li, Siyuan Zhuang, Ying Sheng, Lianmin Zheng, Cody~Hao Yu, Joseph Gonzalez, Hao Zhang, and Ion Stoica.
\newblock Efficient memory management for large language model serving with {PagedAttention}.
\newblock In \emph{Proceedings of the 29th Symposium on Operating Systems Principles}, SOSP '23, pp.\  611–626, New York, NY, USA, 2023. Association for Computing Machinery.
\newblock ISBN 9798400702297.
\newblock \doi{10.1145/3600006.3613165}.
\newblock URL \url{https://doi.org/10.1145/3600006.3613165}.

\bibitem[Kydl{\'\i}{\v{c}}ek et~al.(2025)Kydl{\'\i}{\v{c}}ek, Penedo, and von Werra]{kydlicek2025finepdfs}
Hynek Kydl{\'\i}{\v{c}}ek, Guilherme Penedo, and Leandro von Werra.
\newblock {FinePDFs}: Liberating 3{T} of the finest tokens from {PDFs}, 2025.
\newblock URL \url{https://huggingface.co/spaces/HuggingFaceFW/FinePDFsBlog}.
\newblock Blog post.

\bibitem[Langlais(2025)]{synthpleias}
Pierre-Carl Langlais.
\newblock {SYNTH}: the new data frontier, 2025.
\newblock URL \url{https://huggingface.co/blog/Pclanglais/synth-data-frontier}.
\newblock Blog post.

\bibitem[Li et~al.(2024)Li, Fang, Smyrnis, Ivgi, Jordan, Gadre, Bansal, Guha, Keh, Arora, Garg, Xin, Muennighoff, Heckel, Mercat, Chen, Gururangan, Wortsman, Albalak, Bitton, Nezhurina, Abbas, Hsieh, Ghosh, Gardner, Kilian, Zhang, Shao, Pratt, Sanyal, Ilharco, Daras, Marathe, Gokaslan, Zhang, Chandu, Nguyen, Vasiljevic, Kakade, Song, Sanghavi, Faghri, Oh, Zettlemoyer, Lo, El-Nouby, Pouransari, Toshev, Wang, Groeneveld, Soldaini, Koh, Jitsev, Kollar, Dimakis, Carmon, Dave, Schmidt, and Shankar]{datacomp}
Jeffrey Li, Alex Fang, Georgios Smyrnis, Maor Ivgi, Matt Jordan, Samir Gadre, Hritik Bansal, Etash Guha, Sedrick Keh, Kushal Arora, Saurabh Garg, Rui Xin, Niklas Muennighoff, Reinhard Heckel, Jean Mercat, Mayee Chen, Suchin Gururangan, Mitchell Wortsman, Alon Albalak, Yonatan Bitton, Marianna Nezhurina, Amro Abbas, Cheng-Yu Hsieh, Dhruba Ghosh, Josh Gardner, Maciej Kilian, Hanlin Zhang, Rulin Shao, Sarah Pratt, Sunny Sanyal, Gabriel Ilharco, Giannis Daras, Kalyani Marathe, Aaron Gokaslan, Jieyu Zhang, Khyathi Chandu, Thao Nguyen, Igor Vasiljevic, Sham Kakade, Shuran Song, Sujay Sanghavi, Fartash Faghri, Sewoong Oh, Luke Zettlemoyer, Kyle Lo, Alaaeldin El-Nouby, Hadi Pouransari, Alexander Toshev, Stephanie Wang, Dirk Groeneveld, Luca Soldaini, Pang~Wei Koh, Jenia Jitsev, Thomas Kollar, Alexandros~G. Dimakis, Yair Carmon, Achal Dave, Ludwig Schmidt, and Vaishaal Shankar.
\newblock {DataComp-LM}: In search of the next generation of training sets for language models.
\newblock In A.~Globerson, L.~Mackey, D.~Belgrave, A.~Fan, U.~Paquet, J.~Tomczak, and C.~Zhang (eds.), \emph{Advances in Neural Information Processing Systems}, volume~37, pp.\  14200--14282. Curran Associates, Inc., 2024.
\newblock \doi{10.52202/079017-0455}.
\newblock URL \url{https://proceedings.neurips.cc/paper_files/paper/2024/file/19e4ea30dded58259665db375885e412-Paper-Datasets_and_Benchmarks_Track.pdf}.

\bibitem[Lin et~al.(2021)Lin, Lee, Qiao, and Ren]{xcsqa}
Bill~Yuchen Lin, Seyeon Lee, Xiaoyang Qiao, and Xiang Ren.
\newblock Common sense beyond {E}nglish: Evaluating and improving multilingual language models for commonsense reasoning.
\newblock In Chengqing Zong, Fei Xia, Wenjie Li, and Roberto Navigli (eds.), \emph{Proceedings of the 59th Annual Meeting of the Association for Computational Linguistics and the 11th International Joint Conference on Natural Language Processing (Volume 1: Long Papers)}, pp.\  1274--1287, Online, August 2021. Association for Computational Linguistics.
\newblock \doi{10.18653/v1/2021.acl-long.102}.
\newblock URL \url{https://aclanthology.org/2021.acl-long.102/}.

\bibitem[Loshchilov \& Hutter(2019)Loshchilov and Hutter]{adamw}
Ilya Loshchilov and Frank Hutter.
\newblock Decoupled weight decay regularization.
\newblock In \emph{Proceedings of the Seventh International Conference on Learning Representations}, 2019.
\newblock URL \url{https://openreview.net/forum?id=Bkg6RiCqY7}.

\bibitem[Maini et~al.(2024)Maini, Seto, Bai, Grangier, Zhang, and Jaitly]{wrap}
Pratyush Maini, Skyler Seto, Richard Bai, David Grangier, Yizhe Zhang, and Navdeep Jaitly.
\newblock Rephrasing the web: A recipe for compute and data-efficient language modeling.
\newblock In Lun-Wei Ku, Andre Martins, and Vivek Srikumar (eds.), \emph{Proceedings of the 62nd Annual Meeting of the Association for Computational Linguistics (Volume 1: Long Papers)}, pp.\  14044--14072, Bangkok, Thailand, August 2024. Association for Computational Linguistics.
\newblock \doi{10.18653/v1/2024.acl-long.757}.
\newblock URL \url{https://aclanthology.org/2024.acl-long.757/}.

\bibitem[Mihaylov et~al.(2018)Mihaylov, Clark, Khot, and Sabharwal]{openbookqa}
Todor Mihaylov, Peter Clark, Tushar Khot, and Ashish Sabharwal.
\newblock Can a suit of armor conduct electricity? a new dataset for open book question answering.
\newblock In Ellen Riloff, David Chiang, Julia Hockenmaier, and Jun{'}ichi Tsujii (eds.), \emph{Proceedings of the 2018 Conference on Empirical Methods in Natural Language Processing}, pp.\  2381--2391, Brussels, Belgium, October-November 2018. Association for Computational Linguistics.
\newblock \doi{10.18653/v1/D18-1260}.
\newblock URL \url{https://aclanthology.org/D18-1260/}.

\bibitem[Muennighoff et~al.(2023)Muennighoff, Rush, Barak, Le~Scao, Tazi, Piktus, Pyysalo, Wolf, and Raffel]{NEURIPS2023_9d89448b}
Niklas Muennighoff, Alexander Rush, Boaz Barak, Teven Le~Scao, Nouamane Tazi, Aleksandra Piktus, Sampo Pyysalo, Thomas Wolf, and Colin~A Raffel.
\newblock Scaling data-constrained language models.
\newblock In A.~Oh, T.~Naumann, A.~Globerson, K.~Saenko, M.~Hardt, and S.~Levine (eds.), \emph{Advances in Neural Information Processing Systems}, volume~36, pp.\  50358--50376. Curran Associates, Inc., 2023.
\newblock URL \url{https://proceedings.neurips.cc/paper_files/paper/2023/file/9d89448b63ce1e2e8dc7af72c984c196-Paper-Conference.pdf}.

\bibitem[Nguyen et~al.(2025)Nguyen, Li, Golovneva, Zettlemoyer, Oh, Schmidt, and Li]{rewire}
Thao Nguyen, Yang Li, Olga Golovneva, Luke Zettlemoyer, Sewoong Oh, Ludwig Schmidt, and Xian Li.
\newblock Recycling the web: A method to enhance pre-training data quality and quantity for language models.
\newblock In \emph{Proceedings of the Second Conference on Language Modeling}, 2025.
\newblock URL \url{https://openreview.net/forum?id=lkjhBdz3rn}.

\bibitem[{NVIDIA} et~al.(2025){NVIDIA}, Blakeman, Grattafiori, Basant, Gupta, Khattar, Renduchintala, Vavre, Shukla, Bercovich, Ficek, Shaposhnikov, Kondratenko, Bukharin, Milesi, Taghibakhshi, Liu, Barton, Mahabaleshwarkar, Klein, Zuker, Geifman, Shen, Bhiwandiwalla, Tao, Agrusa, Verma, Guan, Mandarwal, Mehta, Aithal, Poojary, Ahamed, Mishra, Thekkumpate, Dattagupta, Zhu, Sadeghi, Simkin, Lanir, Schifferer, Nushi, Kartal, Rouhani, Ginsburg, Norick, Soubasis, Kisacanin, Yu, Catanzaro, del Mundo, Hwang, Wang, Hsieh, Zhang, Yu, Mungekar, Patel, Alexiuk, Parisien, Neale, Meurillon, Mosk-Aoyama, Su, Corneil, Afrimi, Lo, Rohrer, Serebrenik, Gitman, Levy, Stosic, Mosallanezhad, Narayanan, Nathawani, Rekesh, Yared, Kakwani, Ahn, Riach, Stosic, Minasyan, Lin, Long, Long, Segal, Lantz, Evans, Ning, Chung, Harper, Tramel, Galinkin, Pounds, Briones, Bakhturina, Tsykunov, Ladhak, Wang, Jia, Soares, Chen, Galko, Sun, Siino, Agam, Ajjanagadde, Bhatt, Prasad, Armstrong, Shen, Batmaz, Nalbandyan, Qian, Sharma, Ross, Ngo,
  Hum, Sahota, Wang, Soni, Upadhyay, Mao, Nguyen, Nguyen, Cunningham, Galil, Shahaf, Gitman, Loshchilov, Schen, Levy, Moshkov, Golan, Putterman, Kautz, Scowcroft, Casper, Mitra, Glick, Chen, Oliver, Zhang, Zeng, Lou, Zhang, Choi, Huang, Conway, Guman, Kamalu, Greco, Cohen, Jennings, Daw, Vialard, Yi, Parmar, Xu, Zhu, Briski, Cheung, Luna, Wyss, Santhanam, Shih, Kong, Bhardwaj, Shankar, Puvvada, Pawelec, Anik, McAfee, Sleiman, Derczynski, Ding, Wei, Liebenwein, Vega, Grover, Segbroeck, de~Melo, Nazemi, Sreedhar, Kilaru, Ashkenazi, Romeijn, Chochowski, Cai, Kliegl, Moosaei, Kulka, Novikov, Samadi, Corpuz, Wang, Price, Andersch, Boone, Evans, Martinez, Khona, Chrzanowski, Lee, Dabbah, Shoeybi, Patwary, Mulepati, Nabwani, Hereth, Assaf, Habibi, Zmora, Haber, Sessions, Bhatia, Jukar, Pope, Ludwig, Tajbakhsh, Ailon, Juluru, Sharma, Hrinchuk, Kuchaiev, Delalleau, Olabiyi, Argov, Puny, Tropp, Xie, Chadha, Shamis, Gibbons, Molchanov, Morkisz, Dykas, Jin, Xu, Januszewski, Thombre, Varshney, Gundecha, Tredak, Miao, Wan,
  Mahabadi, Garg, El-Yaniv, Zilberstein, Shafipour, Harang, Izzo, Shahbazyan, Garg, Borkar, Gala, Islam, Hesse, Waleffe, Watve, Koren, Zhang, Hewett, Hewett, Prenger, Timbrook, Mahdavi, Modi, Kriman, Lim, Kariyappa, Satheesh, Kaji, Pasumarthi, Muralidharan, Narentharen, Narenthiran, Bak, Kashirsky, Poulos, Mor, Ramasamy, Acharya, Ghosh, Sreenivas, Thomas, Fan, Gopal, Prabhumoye, Pachori, Toshniwal, Ding, Singh, Sun, Ithape, Majumdar, Singhal, Sergienko, Alborghetti, Ge, Devare, Barua, Panguluri, Gupta, Priyadarshi, Akter, Bui, Ene, Kong, Do, Blankevoort, Moon, Balough, Asida, Natan, Ronen, Konuk, Vashishth, Karpas, De, Noorozi, Noroozi, Srinivasan, Elango, Cui, Korthikanti, Rao, Kurin, Lavrukhin, Anisimov, Jiang, Ahmad, Du, Ping, Zhou, Jennings, Zhang, Prazuch, Ren, Karnati, Choi, Meyer, Wu, Zhang, Qin, Lin, Geifman, Fu, Subara, Suhara, Gao, Moshe, Dong, Zhu, Liu, Chen, and Yan]{nemotron3}
{NVIDIA}, Aaron Blakeman, Aaron Grattafiori, Aarti Basant, Abhibha Gupta, Abhinav Khattar, Adi Renduchintala, Aditya Vavre, Akanksha Shukla, Akhiad Bercovich, Aleksander Ficek, Aleksandr Shaposhnikov, Alex Kondratenko, Alexander Bukharin, Alexandre Milesi, Ali Taghibakhshi, Alisa Liu, Amelia Barton, Ameya~Sunil Mahabaleshwarkar, Amir Klein, Amit Zuker, Amnon Geifman, Amy Shen, Anahita Bhiwandiwalla, Andrew Tao, Anjulie Agrusa, Ankur Verma, Ann Guan, Anubhav Mandarwal, Arham Mehta, Ashwath Aithal, Ashwin Poojary, Asif Ahamed, Asit Mishra, Asma~Kuriparambil Thekkumpate, Ayush Dattagupta, Banghua Zhu, Bardiya Sadeghi, Barnaby Simkin, Ben Lanir, Benedikt Schifferer, Besmira Nushi, Bilal Kartal, Bita~Darvish Rouhani, Boris Ginsburg, Brandon Norick, Brandon Soubasis, Branislav Kisacanin, Brian Yu, Bryan Catanzaro, Carlo del Mundo, Chantal Hwang, Charles Wang, Cheng-Ping Hsieh, Chenghao Zhang, Chenhan Yu, Chetan Mungekar, Chintan Patel, Chris Alexiuk, Christopher Parisien, Collin Neale, Cyril Meurillon, Damon
  Mosk-Aoyama, Dan Su, Dane Corneil, Daniel Afrimi, Daniel Lo, Daniel Rohrer, Daniel Serebrenik, Daria Gitman, Daria Levy, Darko Stosic, David Mosallanezhad, Deepak Narayanan, Dhruv Nathawani, Dima Rekesh, Dina Yared, Divyanshu Kakwani, Dong Ahn, Duncan Riach, Dusan Stosic, Edgar Minasyan, Edward Lin, Eileen Long, Eileen~Peters Long, Elad Segal, Elena Lantz, Ellie Evans, Elliott Ning, Eric Chung, Eric Harper, Eric Tramel, Erick Galinkin, Erik Pounds, Evan Briones, Evelina Bakhturina, Evgeny Tsykunov, Faisal Ladhak, Fay Wang, Fei Jia, Felipe Soares, Feng Chen, Ferenc Galko, Frank Sun, Frankie Siino, Gal~Hubara Agam, Ganesh Ajjanagadde, Gantavya Bhatt, Gargi Prasad, George Armstrong, Gerald Shen, Gorkem Batmaz, Grigor Nalbandyan, Haifeng Qian, Harsh Sharma, Hayley Ross, Helen Ngo, Herbert Hum, Herman Sahota, Hexin Wang, Himanshu Soni, Hiren Upadhyay, Huizi Mao, Huy~C Nguyen, Huy~Q Nguyen, Iain Cunningham, Ido Galil, Ido Shahaf, Igor Gitman, Ilya Loshchilov, Itamar Schen, Itay Levy, Ivan Moshkov, Izik Golan,
  Izzy Putterman, Jan Kautz, Jane~Polak Scowcroft, Jared Casper, Jatin Mitra, Jeffrey Glick, Jenny Chen, Jesse Oliver, Jian Zhang, Jiaqi Zeng, Jie Lou, Jimmy Zhang, Jinhang Choi, Jining Huang, Joey Conway, Joey Guman, John Kamalu, Johnny Greco, Jonathan Cohen, Joseph Jennings, Joyjit Daw, Julien~Veron Vialard, Junkeun Yi, Jupinder Parmar, Kai Xu, Kan Zhu, Kari Briski, Katherine Cheung, Katherine Luna, Keith Wyss, Keshav Santhanam, Kevin Shih, Kezhi Kong, Khushi Bhardwaj, Kirthi Shankar, Krishna~C. Puvvada, Krzysztof Pawelec, Kumar Anik, Lawrence McAfee, Laya Sleiman, Leon Derczynski, Li~Ding, Lizzie Wei, Lucas Liebenwein, Luis Vega, Maanu Grover, Maarten~Van Segbroeck, Maer~Rodrigues de~Melo, Mahdi Nazemi, Makesh~Narsimhan Sreedhar, Manoj Kilaru, Maor Ashkenazi, Marc Romeijn, Marcin Chochowski, Mark Cai, Markus Kliegl, Maryam Moosaei, Matt Kulka, Matvei Novikov, Mehrzad Samadi, Melissa Corpuz, Mengru Wang, Meredith Price, Michael Andersch, Michael Boone, Michael Evans, Miguel Martinez, Mikail Khona, Mike
  Chrzanowski, Minseok Lee, Mohammad Dabbah, Mohammad Shoeybi, Mostofa Patwary, Nabin Mulepati, Najeeb Nabwani, Natalie Hereth, Nave Assaf, Negar Habibi, Neta Zmora, Netanel Haber, Nicola Sessions, Nidhi Bhatia, Nikhil Jukar, Nikki Pope, Nikolai Ludwig, Nima Tajbakhsh, Nir Ailon, Nirmal Juluru, Nishant Sharma, Oleksii Hrinchuk, Oleksii Kuchaiev, Olivier Delalleau, Oluwatobi Olabiyi, Omer~Ullman Argov, Omri Puny, Oren Tropp, Ouye Xie, Parth Chadha, Pasha Shamis, Paul Gibbons, Pavlo Molchanov, Pawel Morkisz, Peter Dykas, Peter Jin, Pinky Xu, Piotr Januszewski, Pranav~Prashant Thombre, Prasoon Varshney, Pritam Gundecha, Przemek Tredak, Qing Miao, Qiyu Wan, Rabeeh~Karimi Mahabadi, Rachit Garg, Ran El-Yaniv, Ran Zilberstein, Rasoul Shafipour, Rich Harang, Rick Izzo, Rima Shahbazyan, Rishabh Garg, Ritika Borkar, Ritu Gala, Riyad Islam, Robert Hesse, Roger Waleffe, Rohit Watve, Roi Koren, Ruoxi Zhang, Russell Hewett, Russell~J. Hewett, Ryan Prenger, Ryan Timbrook, Sadegh Mahdavi, Sahil Modi, Samuel Kriman, Sangkug
  Lim, Sanjay Kariyappa, Sanjeev Satheesh, Saori Kaji, Satish Pasumarthi, Saurav Muralidharan, Sean Narentharen, Sean Narenthiran, Seonmyeong Bak, Sergey Kashirsky, Seth Poulos, Shahar Mor, Shanmugam Ramasamy, Shantanu Acharya, Shaona Ghosh, Sharath~Turuvekere Sreenivas, Shelby Thomas, Shiqing Fan, Shreya Gopal, Shrimai Prabhumoye, Shubham Pachori, Shubham Toshniwal, Shuoyang Ding, Siddharth Singh, Simeng Sun, Smita Ithape, Somshubra Majumdar, Soumye Singhal, Stas Sergienko, Stefania Alborghetti, Stephen Ge, Sugam~Dipak Devare, Sumeet~Kumar Barua, Suseella Panguluri, Suyog Gupta, Sweta Priyadarshi, Syeda~Nahida Akter, Tan Bui, Teodor-Dumitru Ene, Terry Kong, Thanh Do, Tijmen Blankevoort, Tim Moon, Tom Balough, Tomer Asida, Tomer~Bar Natan, Tomer Ronen, Tugrul Konuk, Twinkle Vashishth, Udi Karpas, Ushnish De, Vahid Noorozi, Vahid Noroozi, Venkat Srinivasan, Venmugil Elango, Victor Cui, Vijay Korthikanti, Vinay Rao, Vitaly Kurin, Vitaly Lavrukhin, Vladimir Anisimov, Wanli Jiang, Wasi~Uddin Ahmad, Wei Du, Wei
  Ping, Wenfei Zhou, Will Jennings, William Zhang, Wojciech Prazuch, Xiaowei Ren, Yashaswi Karnati, Yejin Choi, Yev Meyer, Yi-Fu Wu, Yian Zhang, Yigong Qin, Ying Lin, Yonatan Geifman, Yonggan Fu, Yoshi Subara, Yoshi Suhara, Yubo Gao, Zach Moshe, Zhen Dong, Zhongbo Zhu, Zihan Liu, Zijia Chen, and Zijie Yan.
\newblock {NVIDIA} {Nemotron} 3: Efficient and open intelligence.
\newblock \emph{arXiv preprint}, arXiv:2512.20856, 2025.
\newblock URL \url{https://arxiv.org/abs/2512.20856}.

\bibitem[Pasupat \& Liang(2015)Pasupat and Liang]{wikitablequestions}
Panupong Pasupat and Percy Liang.
\newblock Compositional semantic parsing on semi-structured tables.
\newblock In Chengqing Zong and Michael Strube (eds.), \emph{Proceedings of the 53rd Annual Meeting of the Association for Computational Linguistics and the 7th International Joint Conference on Natural Language Processing (Volume 1: Long Papers)}, pp.\  1470--1480, Beijing, China, July 2015. Association for Computational Linguistics.
\newblock \doi{10.3115/v1/P15-1142}.
\newblock URL \url{https://aclanthology.org/P15-1142/}.

\bibitem[Penedo et~al.(2023)Penedo, Malartic, Hesslow, Cojocaru, Alobeidli, Cappelli, Pannier, Almazrouei, and Launay]{NEURIPS2023_fa3ed726}
Guilherme Penedo, Quentin Malartic, Daniel Hesslow, Ruxandra Cojocaru, Hamza Alobeidli, Alessandro Cappelli, Baptiste Pannier, Ebtesam Almazrouei, and Julien Launay.
\newblock The {RefinedWeb} dataset for {F}alcon {LLM}: Outperforming curated corpora with web data only.
\newblock In A.~Oh, T.~Naumann, A.~Globerson, K.~Saenko, M.~Hardt, and S.~Levine (eds.), \emph{Advances in Neural Information Processing Systems}, volume~36, pp.\  79155--79172. Curran Associates, Inc., 2023.
\newblock URL \url{https://proceedings.neurips.cc/paper_files/paper/2023/file/fa3ed726cc5073b9c31e3e49a807789c-Paper-Datasets_and_Benchmarks.pdf}.

\bibitem[Penedo et~al.(2024{\natexlab{a}})Penedo, Kydl\'{\i}\v{c}ek, allal, Lozhkov, Mitchell, Raffel, Von~Werra, and Wolf]{fineweb}
Guilherme Penedo, Hynek Kydl\'{\i}\v{c}ek, Loubna~Ben allal, Anton Lozhkov, Margaret Mitchell, Colin Raffel, Leandro Von~Werra, and Thomas Wolf.
\newblock The {FineWeb} datasets: Decanting the web for the finest text data at scale.
\newblock In A.~Globerson, L.~Mackey, D.~Belgrave, A.~Fan, U.~Paquet, J.~Tomczak, and C.~Zhang (eds.), \emph{Advances in Neural Information Processing Systems}, volume~37, pp.\  30811--30849. Curran Associates, Inc., 2024{\natexlab{a}}.
\newblock \doi{10.52202/079017-0970}.
\newblock URL \url{https://proceedings.neurips.cc/paper_files/paper/2024/file/370df50ccfdf8bde18f8f9c2d9151bda-Paper-Datasets_and_Benchmarks_Track.pdf}.

\bibitem[Penedo et~al.(2024{\natexlab{b}})Penedo, Kydlíček, Cappelli, Sasko, and Wolf]{datatrove}
Guilherme Penedo, Hynek Kydlíček, Alessandro Cappelli, Mario Sasko, and Thomas Wolf.
\newblock {DataTrove}: Large scale data processing, 2024{\natexlab{b}}.
\newblock URL \url{https://github.com/huggingface/datatrove}.
\newblock GitHub repository.

\bibitem[Qin et~al.(2025)Qin, Dong, Zhang, Dong, Huang, Yang, KHADEMI, Zhang, Awadalla, Fung, Chen, Cheng, and Wei]{synthlawscaling}
Zeyu Qin, Qingxiu Dong, Xingxing Zhang, Li~Dong, Xiaolong Huang, Ziyi Yang, MAHMOUD KHADEMI, Dongdong Zhang, Hany~Hassan Awadalla, Yi~R. Fung, Weizhu Chen, Minhao Cheng, and Furu Wei.
\newblock Scaling laws of synthetic data for language model.
\newblock In \emph{Proceedings of the Second Conference on Language Modeling}, 2025.
\newblock URL \url{https://openreview.net/forum?id=UmUXPXHtdl}.

\bibitem[Radford et~al.(2018)Radford, Narasimhan, Salimans, and Sutskever]{radford2018improving}
Alec Radford, Karthik Narasimhan, Tim Salimans, and Ilya Sutskever.
\newblock Improving language understanding by generative pre-training, 2018.
\newblock URL \url{https://cdn.openai.com/research-covers/language-unsupervised/language_understanding_paper.pdf}.
\newblock Technical report.

\bibitem[Raffel et~al.(2020)Raffel, Shazeer, Roberts, Lee, Narang, Matena, Zhou, Li, and Liu]{c4}
Colin Raffel, Noam Shazeer, Adam Roberts, Katherine Lee, Sharan Narang, Michael Matena, Yanqi Zhou, Wei Li, and Peter~J. Liu.
\newblock Exploring the limits of transfer learning with a unified text-to-text transformer.
\newblock \emph{Journal of Machine Learning Research}, 21\penalty0 (140):\penalty0 1--67, 2020.
\newblock URL \url{http://jmlr.org/papers/v21/20-074.html}.

\bibitem[Rajpurkar et~al.(2018)Rajpurkar, Jia, and Liang]{squad2}
Pranav Rajpurkar, Robin Jia, and Percy Liang.
\newblock Know what you don{'}t know: Unanswerable questions for {SQ}u{AD}.
\newblock In Iryna Gurevych and Yusuke Miyao (eds.), \emph{Proceedings of the 56th Annual Meeting of the Association for Computational Linguistics (Volume 2: Short Papers)}, pp.\  784--789, Melbourne, Australia, July 2018. Association for Computational Linguistics.
\newblock \doi{10.18653/v1/P18-2124}.
\newblock URL \url{https://aclanthology.org/P18-2124/}.

\bibitem[Sakaguchi et~al.(2021)Sakaguchi, Bras, Bhagavatula, and Choi]{winogrande}
Keisuke Sakaguchi, Ronan~Le Bras, Chandra Bhagavatula, and Yejin Choi.
\newblock {WinoGrande}: an adversarial winograd schema challenge at scale.
\newblock \emph{Commun. ACM}, 64\penalty0 (9):\penalty0 99–106, August 2021.
\newblock ISSN 0001-0782.
\newblock \doi{10.1145/3474381}.
\newblock URL \url{https://doi.org/10.1145/3474381}.

\bibitem[Shumailov et~al.(2024)Shumailov, Shumaylov, Zhao, Papernot, Anderson, and Gal]{modelcollapse}
Ilia Shumailov, Zakhar Shumaylov, Yiren Zhao, Nicolas Papernot, Ross Anderson, and Yarin Gal.
\newblock {AI} models collapse when trained on recursively generated data.
\newblock \emph{Nature}, 631:\penalty0 755--759, 2024.
\newblock \doi{10.1038/s41586-024-07566-y}.
\newblock URL \url{https://doi.org/10.1038/s41586-024-07566-y}.

\bibitem[Singh et~al.(2026)Singh, Krauss, Jaghouar, Sirovatka, Goddard, Obied, Ong, Straube, Fern, Harley, Stewart, Kealty, Panahi, Kirsten, Deshpande, Vij, Bresnu, Veldurthi, Ravishankar, Bishnoi, Team, Team, Team, McQuade, Hagemann, and Atkins]{singh2026arceetrinitylargetechnical}
Varun Singh, Lucas Krauss, Sami Jaghouar, Matej Sirovatka, Charles Goddard, Fares Obied, Jack~Min Ong, Jannik Straube, Fern, Aria Harley, Conner Stewart, Colin Kealty, Maziyar Panahi, Simon Kirsten, Anushka Deshpande, Anneketh Vij, Arthur Bresnu, Pranav Veldurthi, Raghav Ravishankar, Hardik Bishnoi, DatologyAI Team, Arcee~AI Team, Prime~Intellect Team, Mark McQuade, Johannes Hagemann, and Lucas Atkins.
\newblock Arcee trinity large technical report.
\newblock \emph{arXiv preprint}, arXiv:2602.17004, 2026.
\newblock URL \url{https://arxiv.org/abs/2602.17004}.

\bibitem[Su et~al.(2025)Su, Kong, Lin, Jennings, Norick, Kliegl, Patwary, Shoeybi, and Catanzaro]{nemotroncc}
Dan Su, Kezhi Kong, Ying Lin, Joseph Jennings, Brandon Norick, Markus Kliegl, Mostofa Patwary, Mohammad Shoeybi, and Bryan Catanzaro.
\newblock Nemotron-{CC}: Transforming {C}ommon {C}rawl into a refined long-horizon pretraining dataset.
\newblock In Wanxiang Che, Joyce Nabende, Ekaterina Shutova, and Mohammad~Taher Pilehvar (eds.), \emph{Proceedings of the 63rd Annual Meeting of the Association for Computational Linguistics (Volume 1: Long Papers)}, pp.\  2459--2475, Vienna, Austria, July 2025. Association for Computational Linguistics.
\newblock ISBN 979-8-89176-251-0.
\newblock \doi{10.18653/v1/2025.acl-long.123}.
\newblock URL \url{https://aclanthology.org/2025.acl-long.123/}.

\bibitem[Su et~al.(2024)Su, Ahmed, Lu, Pan, Bo, and Liu]{rope}
Jianlin Su, Murtadha Ahmed, Yu~Lu, Shengfeng Pan, Wen Bo, and Yunfeng Liu.
\newblock {RoFormer}: Enhanced transformer with rotary position embedding.
\newblock \emph{Neurocomputing}, 568:\penalty0 127063, 2024.
\newblock ISSN 0925-2312.
\newblock \doi{https://doi.org/10.1016/j.neucom.2023.127063}.
\newblock URL \url{https://www.sciencedirect.com/science/article/pii/S0925231223011864}.

\bibitem[Tazi et~al.(2025)Tazi, Mom, Zhao, Nguyen, Mekkouri, Werra, and Wolf]{ultrascale_playbook}
Nouamane Tazi, Ferdinand Mom, Haojun Zhao, Phuc Nguyen, Mohamed Mekkouri, Leandro Werra, and Thomas Wolf.
\newblock The ultra-scale playbook: Training {LLMs} on {GPU} clusters, 2025.
\newblock URL \url{https://huggingface.co/spaces/nanotron/ultrascale-playbook}.
\newblock Blog post.

\bibitem[Wang et~al.(2025)Wang, Fu, Cai, Tang, Lyu, Fang, Zheng, Zhou, Zeng, Xiao, Han, and Liu]{ultrafineweb}
Yudong Wang, Zixuan Fu, Jie Cai, Peijun Tang, Hongya Lyu, Yewei Fang, Zhi Zheng, Jie Zhou, Guoyang Zeng, Chaojun Xiao, Xu~Han, and Zhiyuan Liu.
\newblock {Ultra-FineWeb}: Efficient data filtering and verification for high-quality {LLM} training data.
\newblock \emph{arXiv preprint}, arXiv:2505.05427, 2025.
\newblock URL \url{https://arxiv.org/abs/2505.05427}.

\bibitem[Yamaguchi et~al.(2026)Yamaguchi, Mi, and Aletras]{yamaguchi-etal-2026-enhancing}
Atsuki Yamaguchi, Maggie Mi, and Nikolaos Aletras.
\newblock Enhancing linguistic competence of language models through pre-training with language learning tasks.
\newblock In Maria Liakata, Viviane~P. Moreira, Jiajun Zhang, and David Jurgens (eds.), \emph{Proceedings of the 64th Annual Meeting of the {A}ssociation for {C}omputational {L}inguistics (Volume 2: Short Papers)}, pp.\  316--336, San Diego, California, United States, July 2026. Association for Computational Linguistics.
\newblock ISBN 979-8-89176-391-3.
\newblock \doi{10.18653/v1/2026.acl-short.27}.
\newblock URL \url{https://aclanthology.org/2026.acl-short.27/}.

\bibitem[Yang et~al.(2024)Yang, Yang, Hui, Zheng, Yu, Zhou, Li, Li, Liu, Huang, Dong, Wei, Lin, Tang, Wang, Yang, Tu, Zhang, Ma, Yang, Xu, Zhou, Bai, He, Lin, Dang, Lu, Chen, Yang, Li, Xue, Ni, Zhang, Wang, Peng, Men, Gao, Lin, Wang, Bai, Tan, Zhu, Li, Liu, Ge, Deng, Zhou, Ren, Zhang, Wei, Ren, Liu, Fan, Yao, Zhang, Wan, Chu, Liu, Cui, Zhang, Guo, and Fan]{qwen2}
An~Yang, Baosong Yang, Binyuan Hui, Bo~Zheng, Bowen Yu, Chang Zhou, Chengpeng Li, Chengyuan Li, Dayiheng Liu, Fei Huang, Guanting Dong, Haoran Wei, Huan Lin, Jialong Tang, Jialin Wang, Jian Yang, Jianhong Tu, Jianwei Zhang, Jianxin Ma, Jianxin Yang, Jin Xu, Jingren Zhou, Jinze Bai, Jinzheng He, Junyang Lin, Kai Dang, Keming Lu, Keqin Chen, Kexin Yang, Mei Li, Mingfeng Xue, Na~Ni, Pei Zhang, Peng Wang, Ru~Peng, Rui Men, Ruize Gao, Runji Lin, Shijie Wang, Shuai Bai, Sinan Tan, Tianhang Zhu, Tianhao Li, Tianyu Liu, Wenbin Ge, Xiaodong Deng, Xiaohuan Zhou, Xingzhang Ren, Xinyu Zhang, Xipin Wei, Xuancheng Ren, Xuejing Liu, Yang Fan, Yang Yao, Yichang Zhang, Yu~Wan, Yunfei Chu, Yuqiong Liu, Zeyu Cui, Zhenru Zhang, Zhifang Guo, and Zhihao Fan.
\newblock Qwen2 technical report.
\newblock \emph{arXiv preprint}, arXiv:2407.10671, 2024.
\newblock URL \url{https://arxiv.org/abs/2407.10671}.

\bibitem[Yang et~al.(2025{\natexlab{a}})Yang, Li, Yang, Zhang, Hui, Zheng, Yu, Gao, Huang, Lv, Zheng, Liu, Zhou, Huang, Hu, Ge, Wei, Lin, Tang, Yang, Tu, Zhang, Yang, Yang, Zhou, Zhou, Lin, Dang, Bao, Yang, Yu, Deng, Li, Xue, Li, Zhang, Wang, Zhu, Men, Gao, Liu, Luo, Li, Tang, Yin, Ren, Wang, Zhang, Ren, Fan, Su, Zhang, Zhang, Wan, Liu, Wang, Cui, Zhang, Zhou, and Qiu]{qwen3}
An~Yang, Anfeng Li, Baosong Yang, Beichen Zhang, Binyuan Hui, Bo~Zheng, Bowen Yu, Chang Gao, Chengen Huang, Chenxu Lv, Chujie Zheng, Dayiheng Liu, Fan Zhou, Fei Huang, Feng Hu, Hao Ge, Haoran Wei, Huan Lin, Jialong Tang, Jian Yang, Jianhong Tu, Jianwei Zhang, Jianxin Yang, Jiaxi Yang, Jing Zhou, Jingren Zhou, Junyang Lin, Kai Dang, Keqin Bao, Kexin Yang, Le~Yu, Lianghao Deng, Mei Li, Mingfeng Xue, Mingze Li, Pei Zhang, Peng Wang, Qin Zhu, Rui Men, Ruize Gao, Shixuan Liu, Shuang Luo, Tianhao Li, Tianyi Tang, Wenbiao Yin, Xingzhang Ren, Xinyu Wang, Xinyu Zhang, Xuancheng Ren, Yang Fan, Yang Su, Yichang Zhang, Yinger Zhang, Yu~Wan, Yuqiong Liu, Zekun Wang, Zeyu Cui, Zhenru Zhang, Zhipeng Zhou, and Zihan Qiu.
\newblock Qwen3 technical report.
\newblock \emph{arXiv preprint}, arXiv:2505.09388, 2025{\natexlab{a}}.
\newblock URL \url{https://arxiv.org/abs/2505.09388}.

\bibitem[Yang et~al.(2025{\natexlab{b}})Yang, Band, Li, Candes, and Hashimoto]{syntheticcpt}
Zitong Yang, Neil Band, Shuangping Li, Emmanuel Candes, and Tatsunori Hashimoto.
\newblock Synthetic continued pretraining.
\newblock In \emph{Proceedings of the Thirteenth International Conference on Learning Representations}, 2025{\natexlab{b}}.
\newblock URL \url{https://openreview.net/forum?id=07yvxWDSla}.

\bibitem[Zellers et~al.(2019)Zellers, Holtzman, Bisk, Farhadi, and Choi]{hellaswag}
Rowan Zellers, Ari Holtzman, Yonatan Bisk, Ali Farhadi, and Yejin Choi.
\newblock {H}ella{S}wag: Can a machine really finish your sentence?
\newblock In Anna Korhonen, David Traum, and Llu{\'i}s M{\`a}rquez (eds.), \emph{Proceedings of the 57th Annual Meeting of the Association for Computational Linguistics}, pp.\  4791--4800, Florence, Italy, July 2019. Association for Computational Linguistics.
\newblock \doi{10.18653/v1/P19-1472}.
\newblock URL \url{https://aclanthology.org/P19-1472/}.

\bibitem[Zhu et~al.(2015)Zhu, Kiros, Zemel, Salakhutdinov, Urtasun, Torralba, and Fidler]{7410368}
Yukun Zhu, Ryan Kiros, Rich Zemel, Ruslan Salakhutdinov, Raquel Urtasun, Antonio Torralba, and Sanja Fidler.
\newblock Aligning books and movies: Towards story-like visual explanations by watching movies and reading books.
\newblock In \emph{Proceedings of the 2015 IEEE International Conference on Computer Vision (ICCV)}, pp.\  19--27, 2015.
\newblock \doi{10.1109/ICCV.2015.11}.

\end{thebibliography}
\bibliographystyle{colm2026_conference}

\appendix

\section{Related work}
\label{sec:related}

\paragraph{Synthetic Data for Pretraining.}
Foundational research such as WRAP~\citep{wrap} demonstrated that rephrasing web text into diverse styles improves pretraining at a 1.3B scale.
Nemotron-CC~\citep{nemotroncc} scaled this approach to 6.3T tokens using multiple prompts; we isolate the individual prompts of that study and reveals that only Diverse QA Pairs matches the performance of the DCLM baseline.
REWIRE~\citep{rewire} employs Llama 3.3 70B for guided rewriting and suggests that reasoning capabilities of large models are necessary to process data of low quality.
The comparison in \S\ref{subsec:results_model} reveals no consistent advantage for larger models. Indeed, \textsc{FinePhrase} achieves superior results with a 1.7B model at a cost up to 30$\times$ lower than the guided rewrite strategy of REWIRE.
BeyondWeb~\citep{beyondweb} reports strong performance with continuation and summarization rephrasing but did not provide a public data release.
We test those prompts and show that neither matches the performance of DCLM.
Cosmopedia~\citep{cosmopedia} generates content from scratch rather than rephrasing. This approach underperforms every rephrasing approach in our analysis.
EntiGraph~\citep{syntheticcpt} targets continued pretraining with entity-centric augmentation; their diversity scaling complements our finding that prompt-level diversity saturates at approximately 20B tokens.

\paragraph{Understanding Synthetic Data.}
\citet{demystifyingsynth} trained over 1,000 LLMs and determined that a mixture of $\sim$30\% synthetic data substantially accelerates convergence, while generator models exceeding 8B parameters do not yield superior pretraining data.
This study extends this analysis to prompt design, the choice of mix-in datasets, and the phenomenon of template collapse.
While \citet{synthlawscaling} investigate the scaling laws of synthetic data, we focus on design choices at fixed scale.
The literature on model collapse~\citep{modelcollapse} documents the degradation resulting from training exclusively on model outputs. Our pipeline circumvents this issue by consistently mixing synthetic content with the original data.\looseness=-1

\section{Prompt Templates}
\label{appendix:prompt}

\subsection{Nemotron-CC Prompts}
\label{appendix:prompt_nemotron}

\paragraph{Diverse QA Pairs.}

\begin{Verbatim}[breaklines]
Task: Read the text, ask questions and answer them.
Follow these instructions:
1. Ask diverse questions that require different cognitive skills or cover different aspects of the text.
1. Ask questions in various forms such as:
    - Yes/No questions that require determining whether a statement is true or false.
    - Open-ended questions that begin with words like what, how, when, where, why and who.
    - Multi-choice questions that offers two or more options to choose from. Include the options in the question.
    - Comparison questions that compare two quantities or objects and determine the relationship between them.
    - Reading comprehension questions that test the ability to understand and analyze the text.
    - Problem-solving questions that test the ability to solve mathematical, physical, or logical problems.

1. Focus on asking questions about factual information, important knowledge, or concrete details in the text.
1. Write questions and answers using clear and concise language.
1. Use plain text. Do not use Markdown.
1. Each question and answer pair should be on a separate line. Tag the question with "Question:" and the answer with "Answer:".

Text:
[TEXT]
Task:
After reading the above text, ask up to 8 questions and provide the correct answers following the instructions. Give your response in this format:
Here are the questions and answers based on the provided text:
- Question: [first question] Answer: [first answer]
- Question: [second question] Answer: [second answer]

....
\end{Verbatim}

\paragraph{Extract Knowledge.}
\begin{Verbatim}[breaklines]
Your task is to rewrite knowledge from the provided text following these instructions:
- Rewrite the text as a passage or passages using easy-to-understand and high-quality English like sentences in textbooks and Wikipedia.
- Focus on content in disciplines such as humanities, social sciences, natural sciences, technology, engineering, math, law and legal, business, management, art, education, agricultural sciences, politics, and history.
- Disregard content that does not contain useful facts or knowledge.
- Retain examples, explanations of reasoning processes, and supporting evidence to maintain the text's depth and context.
- Do not add or alter details. Only restate what is already in the text.
- Write in plain text.
- Do not add titles, subtitles, note, or comment.

Text:
[TEXT]
Task:
Rewrite facts and knowledge from the above text as a passage or passages following the instructions.
\end{Verbatim}

\paragraph{Distill.}

\begin{Verbatim}[breaklines]
Your task is to read and paraphrase the provided text following these instructions:
- Aim to create a condensed but accurate and informative version of the original text, not a simplistic summary.
- Capture and preserve the crucial information, key concepts, important values, and factual details in the original text, while making it more readable and accessible.
- Retain technical terms, specialized vocabulary, and complex concepts.
- Retain examples, explanations of reasoning processes, and supporting evidence to maintain the text's depth and context.
- Only include information that is present in the original text. Do not adding new or unsubstantiated claims.
- Write in plain text.

Here is the text:
[TEXT]
Task:
After thoroughly reading the above text, paraphrase it in high-quality and clear English following the instructions.
\end{Verbatim}

\paragraph{Wikipedia.}
\begin{Verbatim}[breaklines]
For the following paragraph give me a diverse paraphrase of the same in high quality English language as in sentences on Wikipedia. Begin your answer on a separate line with "Here is a paraphrased version:".
Text:
[TEXT]
\end{Verbatim}

\paragraph{Knowledge List.}
\begin{Verbatim}[breaklines]
Review the text and extract the key information. Follow these instructions:
- Carefully read the above text and provide a concise and organized list of factual information, concrete details, key concepts, and important numbers and statistics extracted from the text.
- Ensure each point is clear, specific, and supported by the original text.
- Ensure the extract text is information-dense and easier to learn from.
- Do not add titles or headings.

Text:
[TEXT]
Task:
Extract the factual information, concrete details, and key concepts from the above text following the instructions.
\end{Verbatim}

\subsection{REWIRE}
\label{appendix:prompt_rewire}

\begin{Verbatim}[breaklines]
Below is a draft from an AI Assistant when trying to accomplish task or solving a problem. Analyze and understand the task and problem(s) to be solved. Then pretend to be the expert who is most skillful to acomplish this task, write down the detailed thinking process and internal monologue that went into identifying a strategy and lay out a plan about how to solve this problem. Experts usually apply meta-reasoning and planning to reason about how to best accomplish the task before jumping to solution.
Deliberate meta-reasoning also involves reflection which can help identify issues and take a step back to explore other paths. Below are some generic examples of starting questions experts could ask themselves during meta-reasoning process. The expert will come up with the most relevant questions that can help with their thinking process, which are also very specific to the task.
Let's first try to understand the task and exactly what problem(s) to be solved. What is the core issue or problem that needs to be addressed? What are the key assumptions underlying this problem?
How can I break down this problem into smaller, more manageable parts? How can I simplify the problem so that it is easier to solve?
What kinds of solution typically are produced for this kind of problem specification? Given the problem specification and the current best solution, have a guess about other possible solutions. Let's imagine the current best solution is totally wrong, what other ways are there to think about the problem specific
What is the best way to modify this current best solution, given what you know about these kinds of problem specification?
Am I on the right track? Let's check our progress so far.
Let's make a step by step plan and implement it with good notion and explanation.
Finally, write an improved response after thinking about how to accomplish the task. Take information and details from the original draft whenever they are useful. Therefore, the improved response should not be shorter than the original response. The improved response should have better formatting and readability, with more coherent and in-depth reasoning, while removing any noise or digression. Note that the best experts chosen to answer each prompt may be different, so please make sure the you do not sound like the same expert for all tasks.
IMPORTANT: Start your analysis and thinking right away. DO NOT add any filler text, explanations or notes about your response. Put the thinking and planning between {'<'}thinking starts{'>'} and {'<'}thinking ends{'>'}, and the improved response between {'<'}improved response starts{'>'} and {'<'}improved response ends{'>'}.
Original Draft: [TEXT]
\end{Verbatim}

\subsection{BeyondWeb}
\label{appendix:prompt_beyondweb}

\paragraph{Continue.}
\begin{Verbatim}[breaklines]
Continue the following text in the same style as the original. Start with the continuation directly.
Text:
[TEXT]
\end{Verbatim}

\paragraph{Summarize.}
\begin{Verbatim}[breaklines]
Summarize the following text. Write a standalone summary without referencing the text. Directly start with the summary. Do not say anything else.
Text:
[TEXT]
Summary:
\end{Verbatim}

\subsection{Structured Prompts}
\label{appendix:structured_prompts}
\paragraph{\texttt{math}.}
\begin{Verbatim}[breaklines]
Rewrite the document to create a mathematical word problem based on the numerical data or relationships in the text. Provide a step-by-step solution that shows the calculation process clearly. Create a problem that requires multi-step reasoning and basic arithmetic operations. It should include the question followed by a detailed solution showing each calculation step. Output only the problem and solution, nothing else.
Document:
[TEXT]
\end{Verbatim}

\paragraph{\texttt{faq}.}
\begin{Verbatim}[breaklines]
Rewrite the document as a comprehensive FAQ (Frequently Asked Questions). Extract or infer the key questions a reader would have about this topic, then provide clear, direct answers. Order questions logically—from foundational to advanced, or by topic area. Each answer should be self-contained and understandable without reference to other answers. Ensure the FAQ works as a standalone document. Output only the FAQ, nothing else.
Document:
[TEXT]
\end{Verbatim}

\paragraph{\texttt{table}.}
\begin{Verbatim}[breaklines]
Rewrite the document as a structured table that organizes the key information, then generate one question-answer pair based on the table. First extract the main data points and organize them into a clear table format with appropriate headers using markdown table syntax with proper alignment. After the table, generate one insightful question that can be answered using the table data. Provide a clear, concise answer to the question based on the information in the table. Output only the table followed by the question-answer pair, nothing else.
Document:
[TEXT]
\end{Verbatim}

\paragraph{\texttt{tutorial}.}
\begin{Verbatim}[breaklines]
Rewrite the document as a clear, step-by-step tutorial or instructional guide. Use numbered steps or bullet points where appropriate to enhance clarity. Preserve all essential information while ensuring the style feels didactic and easy to follow. Output only the tutorial, nothing else.
Document:
[TEXT]
\end{Verbatim}

\section{Implementation Details}
\label{appendix:impl}
We built our pipeline using DataTrove~\citep[v0.6.0]{datatrove}, with distributed inference for synthetic data generation managed through vLLM~\citep[v0.8.4]{vllm}, training handled by Nanotron~\citep[v0.4]{ultrascale_playbook}\footnote{\url{https://github.com/huggingface/nanotron/}}, and evaluation executed via LightEval~\citep[v0.9.1dev0]{lighteval}.

\section{Supplementary Results}
\label{appendix:results}

\begin{itemize}[nosep,leftmargin=*]
    \item Table \ref{tab:existing_data_appendix}: Compares existing web-crawled and synthetic datasets (e.g., DCLM, FineWeb, Cosmopedia).
    \item Table \ref{tab:baselines_appendix}: Evaluates existing rephrasing methods such as \textit{Diverse QA Pairs}, \textit{Distill}, \textit{Knowledge List}, and \textit{Guided Rewriting}.
    \item Table \ref{tab:new-prompts_appenendix}: Provides a detailed breakdown of performance across the four new pedagogical structured prompts (\texttt{math}, \texttt{faq}, \texttt{table}, \texttt{tutorial}).
    \item Table \ref{tab:model_scale_appendix}: Examines the impact of model scaling on performance across 270M to 27B parameter regimes using Gemma 3 models.
    \item Table \ref{tab:model-family_appendix}: Reports mean scores and standard deviations across major model families.
    \item Table \ref{tab:synth-vs-mixed_appendix}: Compares purely synthetic training against mixed training ($\mathcal{D}_\text{mix}$) for each specialized prompt.
    \item Table \ref{tab:mix-in_appendix}: Analyzes the performance sensitivity relative to the quality of mix-in data ($\mathcal{D}_\text{mix}$).
    \item Table \ref{tab:source_data_appendix}: Investigates the performance sensitivity relative to the choice of the source dataset ($\mathcal{D}_\text{source}$).
    \item Table \ref{tab:guided_rewrite}: Summarizes macro-averaged scores for model scales using the guided rewrite baseline.
    \item Table \ref{tab:varying_quality_scale_prompt}: Provides a multi-dimensional macro-average comparison across data quality, model scale, and prompt type.
\end{itemize}

\begin{table}[th]
    \centering
    \caption{Breakdown scores ($\times 100$) for different existing datasets at 10K steps.
    $\star$ indicates a synthetic dataset.}
    \label{tab:existing_data_appendix}
\resizebox{\textwidth}{!}{
\begin{tabular}{lcccccccccccc}
\toprule
& \multicolumn{2}{c}{\textbf{General Knowledge}} & \multicolumn{2}{c}{\textbf{Reading Comp.}} & \multicolumn{2}{c}{\textbf{Reasoning}} & \multicolumn{3}{c}{\textbf{NLU}} & \textbf{Math} & \multicolumn{2}{c}{\textbf{Table Understanding}} \\
\cmidrule(lr){2-3}\cmidrule(lr){4-5}\cmidrule(lr){6-7}\cmidrule(lr){8-10}\cmidrule(lr){11-11}\cmidrule(lr){12-13}
\textbf{Approach} & \textbf{ARC} & \textbf{MMLU} & \textbf{DROP} & \textbf{SQuAD v2} & \textbf{OBQA} & \textbf{XCSQA} & \textbf{WG} & \textbf{PIQA} & \textbf{HS} & \textbf{GSM8K} & \textbf{WikiTQ} & \textbf{TQA}\\
\midrule

DCLM & 15.26 & 12.78 & 16.55 & 42.56 & 2.94 & 4.84 & 3.90 & 7.42 & 9.25 & 10.87 & 19.31 & 13.76\\
Ultra-FineWeb & 15.04 & 12.09 & 14.67 & 40.44 & 2.09 & 2.83 & 3.63 & 6.96 & 9.03 & 10.32 & 17.91 & 14.62\\
FineWeb-HQ & 17.31 & 12.15 & 12.33 & 34.54 & 2.37 & 2.41 & 2.63 & 5.73 & 7.51 & 10.85 & 15.01 & 11.62\\
FineWeb-LQ & 3.87 & 6.97 & 12.17 & 23.86 & 0.78 & 0.69 & 2.45 & 6.30 & 8.95 & 9.48 & 12.75 & 10.83\\
\midrule
Nemotron-HQ-Synth$\star$ & 18.33 & 14.30 & 16.39 & 42.56 & 2.67 & 4.36 & 4.34 & 6.98 & 8.13 & 10.53 & 17.99 & 10.04\\
REWIRE$\star$ & 14.87 & 12.12 & 14.04 & 46.08 & 2.71 & 5.65 & 2.69 & 7.32 & 8.20 & 10.75 & 20.33 & 11.60\\
Cosmopedia$\star$ & 10.99 & 9.72 & 13.85 & 37.40 & 1.30 & 3.23 & 0.64 & 5.76 & 6.85 & 10.78 & 13.69 & 2.34\\
SYNTH$\star$ & 10.87 & 11.23 & 12.68 & 32.27 & 1.13 & 1.26 & 1.54 & 3.56 & 4.71 & 10.05 & 16.93 & 6.07\\

\bottomrule
\end{tabular}
}
\end{table}

\begin{table}[th]
    \centering
    \caption{Breakdown scores for different existing rephrasing approaches.}
    \label{tab:baselines_appendix}
\resizebox{\textwidth}{!}{
\begin{tabular}{lcccccccccccc}
\toprule
& \multicolumn{2}{c}{\textbf{General Knowledge}} & \multicolumn{2}{c}{\textbf{Reading Comp.}} & \multicolumn{2}{c}{\textbf{Reasoning}} & \multicolumn{3}{c}{\textbf{NLU}} & \textbf{Math} & \multicolumn{2}{c}{\textbf{Table Understanding}} \\
\cmidrule(lr){2-3}\cmidrule(lr){4-5}\cmidrule(lr){6-7}\cmidrule(lr){8-10}\cmidrule(lr){11-11}\cmidrule(lr){12-13}
\textbf{Approach} & \textbf{ARC} & \textbf{MMLU} & \textbf{DROP} & \textbf{SQuAD v2} & \textbf{OBQA} & \textbf{XCSQA} & \textbf{WG} & \textbf{PIQA} & \textbf{HS} & \textbf{GSM8K} & \textbf{WikiTQ} & \textbf{TQA}\\
\midrule

\rowcolor{gray!20}DCLM & 15.26 & 12.78 & 16.55 & 42.56 & 2.94 & 4.84 & 3.90 & 7.42 & 9.25 & 10.87 & 19.31 & 13.76\\
\midrule
Diverse QA Pairs & 21.56 & 13.58 & 17.62 & 51.65 & 3.03 & 3.06 & 3.38 & 5.86 & 7.51 & 10.91 & 17.93 & 12.03\\
Distill & 20.28 & 13.58 & 14.88 & 39.65 & 2.64 & 3.19 & 3.27 & 6.20 & 7.53 & 10.94 & 16.38 & 12.61\\
Wikipedia & 19.46 & 13.52 & 14.24 & 42.90 & 2.61 & 2.91 & 3.60 & 5.90 & 7.40 & 10.41 & 16.38 & 11.77\\
Knowledge List & 19.96 & 14.06 & 17.83 & 39.91 & 3.06 & 2.94 & 2.46 & 6.03 & 7.32 & 11.16 & 13.96 & 11.52\\
Extract Knowledge & 17.52 & 12.69 & 13.56 & 34.45 & 2.33 & 2.68 & 3.57 & 5.60 & 6.97 & 10.49 & 14.50 & 10.45\\
\midrule
Guided Rewrite & 19.63 & 13.39 & 16.83 & 44.95 & 2.80 & 3.34 & 3.16 & 5.69 & 7.44 & 10.97 & 17.60 & 11.93\\
\midrule
Summarize & 19.51 & 13.45 & 16.43 & 39.55 & 2.78 & 2.92 & 2.88 & 5.57 & 7.18 & 11.20 & 16.43 & 10.83\\
Continue & 20.49 & 13.82 & 15.18 & 43.27 & 3.09 & 3.88 & 4.07 & 6.08 & 7.52 & 10.88 & 17.94 & 12.03\\

\bottomrule
\end{tabular}
}
\end{table}

\begin{table}[th]
    \centering
    \caption{Breakdown scores for new pedagogical structured prompts vs. DCLM.}
    \label{tab:new-prompts_appenendix}
\resizebox{\textwidth}{!}{
\begin{tabular}{lcccccccccccc}
\toprule
& \multicolumn{2}{c}{\textbf{General Knowledge}} & \multicolumn{2}{c}{\textbf{Reading Comp.}} & \multicolumn{2}{c}{\textbf{Reasoning}} & \multicolumn{3}{c}{\textbf{NLU}} & \textbf{Math} & \multicolumn{2}{c}{\textbf{Table Understanding}} \\
\cmidrule(lr){2-3}\cmidrule(lr){4-5}\cmidrule(lr){6-7}\cmidrule(lr){8-10}\cmidrule(lr){11-11}\cmidrule(lr){12-13}
\textbf{Approach} & \textbf{ARC} & \textbf{MMLU} & \textbf{DROP} & \textbf{SQuAD v2} & \textbf{OBQA} & \textbf{XCSQA} & \textbf{WG} & \textbf{PIQA} & \textbf{HS} & \textbf{GSM8K} & \textbf{WikiTQ} & \textbf{TQA}\\
\midrule

\rowcolor{gray!20}DCLM & 15.26 & 12.78 & 16.55 & 42.56 & 2.94 & 4.84 & 3.90 & 7.42 & 9.25 & 10.87 & 19.31 & 13.76\\
\midrule
\texttt{math} & 20.73 & 14.37 & 16.04 & 53.81 & 2.89 & 3.76 & 3.12 & 6.11 & 7.68 & 12.36 & 20.30 & 14.50\\
\texttt{faq} & 21.57 & 14.22 & 16.50 & 50.20 & 3.47 & 3.27 & 2.74 & 6.09 & 7.70 & 10.85 & 16.80 & 13.46\\
\texttt{table} & 22.76 & 14.66 & 14.64 & 47.37 & 3.23 & 3.91 & 2.88 & 5.94 & 7.38 & 11.24 & 21.61 & 15.45\\
\texttt{tutorial} & 21.34 & 14.11 & 17.98 & 45.49 & 3.06 & 3.51 & 3.71 & 6.07 & 7.71 & 10.54 & 17.29 & 14.42\\

\bottomrule
\end{tabular}
}
\end{table}

\begin{table}[th]
    \centering
    \caption{Breakdown scores for different Gemma 3 model scales.}
    \label{tab:model_scale_appendix}
\resizebox{\textwidth}{!}{
\begin{tabular}{lcccccccccccc}
\toprule
& \multicolumn{2}{c}{\textbf{General Knowledge}} & \multicolumn{2}{c}{\textbf{Reading Comp.}} & \multicolumn{2}{c}{\textbf{Reasoning}} & \multicolumn{3}{c}{\textbf{NLU}} & \textbf{Math} & \multicolumn{2}{c}{\textbf{Table Understanding}} \\
\cmidrule(lr){2-3}\cmidrule(lr){4-5}\cmidrule(lr){6-7}\cmidrule(lr){8-10}\cmidrule(lr){11-11}\cmidrule(lr){12-13}
\textbf{Approach} & \textbf{ARC} & \textbf{MMLU} & \textbf{DROP} & \textbf{SQuAD v2} & \textbf{OBQA} & \textbf{XCSQA} & \textbf{WG} & \textbf{PIQA} & \textbf{HS} & \textbf{GSM8K} & \textbf{WikiTQ} & \textbf{TQA}\\
\midrule

\rowcolor{gray!20} DCLM & 15.26 & 12.78 & 16.55 & 42.56 & 2.94 & 4.84 & 3.90 & 7.42 & 9.25 & 10.87 & 19.31 & 13.76\\
\midrule
270M & 19.85 & 13.61 & 18.57 & 44.42 & 2.87 & 2.70 & 2.90 & 5.90 & 7.40 & 10.60 & 17.31 & 12.70\\
1B & 20.73 & 14.37 & 16.04 & 53.81 & 2.89 & 3.76 & 3.12 & 6.11 & 7.68 & 12.36 & 20.30 & 14.50\\
4B & 19.32 & 14.35 & 19.24 & 46.03 & 3.15 & 3.43 & 3.32 & 5.60 & 7.44 & 13.51 & 20.88 & 15.03\\
12B & 19.77 & 13.97 & 17.30 & 43.39 & 3.15 & 3.01 & 3.83 & 5.67 & 7.44 & 13.96 & 19.86 & 14.84\\
27B & 20.81 & 14.82 & 17.11 & 42.08 & 3.36 & 2.84 & 4.25 & 5.80 & 7.19 & 14.30 & 20.51 & 13.79\\

\bottomrule
\end{tabular}
}
\end{table}

\begin{table}[th]
    \centering
    \caption{Breakdown scores for different model families across the four new prompts (all $\sim$1B scale) with standard deviations in parentheses.}
    \label{tab:model-family_appendix}
\resizebox{\textwidth}{!}{
\begin{tabular}{lcccccccccccc}
\toprule
& \multicolumn{2}{c}{\textbf{General Knowledge}} & \multicolumn{2}{c}{\textbf{Reading Comp.}} & \multicolumn{2}{c}{\textbf{Reasoning}} & \multicolumn{3}{c}{\textbf{NLU}} & \textbf{Math} & \multicolumn{2}{c}{\textbf{Table Understanding}} \\
\cmidrule(lr){2-3}\cmidrule(lr){4-5}\cmidrule(lr){6-7}\cmidrule(lr){8-10}\cmidrule(lr){11-11}\cmidrule(lr){12-13}
\textbf{Approach} & \textbf{ARC} & \textbf{MMLU} & \textbf{DROP} & \textbf{SQuAD v2} & \textbf{OBQA} & \textbf{XCSQA} & \textbf{WG} & \textbf{PIQA} & \textbf{HS} & \textbf{GSM8K} & \textbf{WikiTQ} & \textbf{TQA}\\
\midrule

\rowcolor{gray!20}DCLM & 15.26 (0.00) & 12.78 (0.00) & 16.55 (0.00) & 42.56 (0.00) & 2.94 (0.00) & 4.84 (0.00) & 3.90 (0.00) & 7.42 (0.00) & 9.25 (0.00) & 10.87 (0.00) & 19.31 (0.00) & 13.76 (0.00)\\
\midrule
SmolLM2 & 22.47 (0.46) & 15.34 (0.14) & 18.86 (1.96) & 59.67 (1.58) & 3.63 (0.16) & 4.84 (0.11) & 3.79 (0.18) & 6.48 (0.13) & 7.82 (0.15) & 11.71 (1.06) & 20.48 (0.81) & 16.80 (2.64)\\
Falcon 3 & 22.20 (0.44) & 15.37 (0.23) & 17.29 (2.45) & 51.55 (2.39) & 3.30 (0.13) & 4.59 (0.13) & 3.73 (0.23) & 6.33 (0.24) & 7.75 (0.12) & 11.33 (0.54) & 20.45 (1.49) & 16.00 (1.72)\\
Qwen 3 & 22.29 (2.07) & 15.02 (0.79) & 15.47 (1.07) & 46.24 (3.00) & 3.28 (0.15) & 3.84 (0.56) & 3.51 (0.21) & 6.11 (0.22) & 7.54 (0.16) & 11.47 (0.68) & 18.76 (0.72) & 13.30 (1.56)\\
Gemma 3 1B & 21.60 (0.74) & 14.34 (0.21) & 16.29 (1.19) & 49.22 (3.14) & 3.16 (0.21) & 3.61 (0.24) & 3.11 (0.37) & 6.05 (0.06) & 7.62 (0.14) & 11.25 (0.69) & 19.00 (2.02) & 14.46 (0.70)\\
Granite 3 & 20.49 (0.39) & 14.66 (0.19) & 17.20 (2.04) & 50.90 (2.99) & 3.27 (0.17) & 4.01 (0.25) & 3.02 (0.42) & 6.28 (0.14) & 7.71 (0.09) & 11.20 (0.79) & 18.43 (1.54) & 14.56 (2.23)\\
LLaMA 3.2 & 21.32 (0.85) & 14.40 (0.30) & 16.79 (1.29) & 49.26 (3.37) & 3.23 (0.18) & 4.03 (0.15) & 3.29 (0.26) & 5.92 (0.20) & 7.49 (0.19) & 11.35 (0.79) & 19.52 (1.33) & 13.93 (1.57)\\

\bottomrule
\end{tabular}
}
\end{table}

\begin{table}[t]
    \centering
    \caption{Breakdown scores between synthetic-only and mixed training.}
    \label{tab:synth-vs-mixed_appendix}
\resizebox{\textwidth}{!}{
\begin{tabular}{lcccccccccccc}
\toprule
& \multicolumn{2}{c}{\textbf{General Knowledge}} & \multicolumn{2}{c}{\textbf{Reading Comp.}} & \multicolumn{2}{c}{\textbf{Reasoning}} & \multicolumn{3}{c}{\textbf{NLU}} & \textbf{Math} & \multicolumn{2}{c}{\textbf{Table Understanding}} \\
\cmidrule(lr){2-3}\cmidrule(lr){4-5}\cmidrule(lr){6-7}\cmidrule(lr){8-10}\cmidrule(lr){11-11}\cmidrule(lr){12-13}
\textbf{Approach} & \textbf{ARC} & \textbf{MMLU} & \textbf{DROP} & \textbf{SQuAD v2} & \textbf{OBQA} & \textbf{XCSQA} & \textbf{WG} & \textbf{PIQA} & \textbf{HS} & \textbf{GSM8K} & \textbf{WikiTQ} & \textbf{TQA}\\
\midrule

\rowcolor{gray!20}DCLM & 15.26 & 12.78 & 16.55 & 42.56 & 2.94 & 4.84 & 3.90 & 7.42 & 9.25 & 10.87 & 19.31 & 13.76\\
\midrule
\texttt{math} w/o $\mathcal{D}_\text{mix}$ & 20.24 & 14.02 & 17.23 & 54.07 & 2.64 & 2.96 & 3.02 & 5.76 & 6.99 & 12.36 & 20.34 & 14.26\\
\texttt{math} & 20.73 & 14.37 & 16.04 & 53.81 & 2.89 & 3.76 & 3.12 & 6.11 & 7.68 & 12.36 & 20.30 & 14.50\\
\midrule
\texttt{faq} w/o $\mathcal{D}_\text{mix}$ & 21.34 & 13.02 & 16.28 & 40.95 & 2.47 & 3.24 & 2.70 & 5.55 & 6.63 & 10.83 & 16.24 & 11.11\\
\texttt{faq} & 21.57 & 14.22 & 16.50 & 50.20 & 3.47 & 3.27 & 2.74 & 6.09 & 7.70 & 10.85 & 16.80 & 13.46\\
\midrule
\texttt{table} w/o $\mathcal{D}_\text{mix}$ & 21.73 & 13.68 & 15.54 & 46.61 & 2.63 & 3.62 & 2.91 & 5.02 & 6.05 & 10.91 & 17.03 & 12.00\\
\texttt{table} & 22.76 & 14.66 & 14.64 & 47.37 & 3.23 & 3.91 & 2.88 & 5.94 & 7.38 & 11.24 & 21.61 & 15.45\\
\midrule
\texttt{tutorial} w/o $\mathcal{D}_\text{mix}$ & 20.62 & 12.65 & 15.67 & 33.88 & 2.38 & 2.93 & 2.64 & 5.66 & 6.39 & 10.82 & 14.98 & 12.03\\
\texttt{tutorial} & 21.34 & 14.11 & 17.98 & 45.49 & 3.06 & 3.51 & 3.71 & 6.07 & 7.71 & 10.54 & 17.29 & 14.42\\

\bottomrule
\end{tabular}
}
\end{table}

\begin{table}[t]
    \centering
    \caption{Breakdown scores for different mix-in dataset $\mathcal{D}_\text{mix}$ choices with FWHQ (HQ) or FWLQ (LQ) as $\mathcal{D}_\text{source}$.
    Parentheses indicate the mix-in dataset quality (HQ vs. LQ).}
    \label{tab:mix-in_appendix}
\resizebox{\textwidth}{!}{
\begin{tabular}{lcccccccccccc}
\toprule
& \multicolumn{2}{c}{\textbf{General Knowledge}} & \multicolumn{2}{c}{\textbf{Reading Comp.}} & \multicolumn{2}{c}{\textbf{Reasoning}} & \multicolumn{3}{c}{\textbf{NLU}} & \textbf{Math} & \multicolumn{2}{c}{\textbf{Table Understanding}} \\
\cmidrule(lr){2-3}\cmidrule(lr){4-5}\cmidrule(lr){6-7}\cmidrule(lr){8-10}\cmidrule(lr){11-11}\cmidrule(lr){12-13}
\textbf{Approach} & \textbf{ARC} & \textbf{MMLU} & \textbf{DROP} & \textbf{SQuAD v2} & \textbf{OBQA} & \textbf{XCSQA} & \textbf{WG} & \textbf{PIQA} & \textbf{HS} & \textbf{GSM8K} & \textbf{WikiTQ} & \textbf{TQA}\\
\midrule

\rowcolor{gray!20}FineWeb-LQ & 3.87 & 6.97 & 12.17 & 23.86 & 0.78 & 0.69 & 2.45 & 6.30 & 8.95 & 9.48 & 12.75 & 10.83\\
\rowcolor{gray!20}FineWeb-HQ & 17.31 & 12.15 & 12.33 & 34.54 & 2.37 & 2.41 & 2.63 & 5.73 & 7.51 & 10.85 & 15.01 & 11.62\\
\midrule
Cosmopedia (LQ) & 11.20 & 9.28 & 14.26 & 38.40 & 1.31 & 1.66 & 1.13 & 5.99 & 6.93 & 10.73 & 14.48 & 6.85\\
Cosmopedia (HQ) & 19.53 & 12.74 & 13.24 & 43.39 & 2.38 & 2.85 & 2.20 & 6.33 & 7.37 & 11.26 & 16.26 & 10.97\\
\midrule
DCLM (LQ) & 12.37 & 11.20 & 16.54 & 37.59 & 2.70 & 3.76 & 3.99 & 7.28 & 8.97 & 10.36 & 17.12 & 13.83\\
DCLM (HQ) & 20.34 & 13.91 & 15.67 & 43.10 & 3.08 & 4.95 & 4.00 & 7.28 & 8.87 & 10.59 & 19.97 & 14.56\\
\midrule
FineWeb-HQ (LQ) & 18.34 & 12.84 & 14.60 & 38.16 & 2.32 & 2.92 & 3.69 & 6.78 & 8.14 & 10.31 & 18.69 & 12.81\\
FineWeb-HQ (HQ) & 21.34 & 14.11 & 17.98 & 45.49 & 3.06 & 3.51 & 3.71 & 6.07 & 7.71 & 10.54 & 17.29 & 14.42\\
\midrule
FineWeb-LQ (LQ) & 5.73 & 7.56 & 10.98 & 29.96 & 1.35 & 1.84 & 2.75 & 6.94 & 8.39 & 9.36 & 14.56 & 10.19\\
FineWeb-LQ (HQ) & 18.30 & 12.34 & 16.15 & 44.62 & 2.35 & 3.59 & 3.27 & 7.22 & 8.75 & 10.18 & 18.17 & 12.54\\

\bottomrule
\end{tabular}
}
\end{table}

\begin{table}[t]
    \centering
    \caption{Breakdown scores for different source dataset $\mathcal{D}_\text{source}$ choices.
    Parentheses indicate the mix-in dataset used (Cosmopedia, DCLM, or FWHQ/FWLQ).
    }
    \label{tab:source_data_appendix}
\resizebox{\textwidth}{!}{
\begin{tabular}{lcccccccccccc}
\toprule
& \multicolumn{2}{c}{\textbf{General Knowledge}} & \multicolumn{2}{c}{\textbf{Reading Comp.}} & \multicolumn{2}{c}{\textbf{Reasoning}} & \multicolumn{3}{c}{\textbf{NLU}} & \textbf{Math} & \multicolumn{2}{c}{\textbf{Table Understanding}} \\
\cmidrule(lr){2-3}\cmidrule(lr){4-5}\cmidrule(lr){6-7}\cmidrule(lr){8-10}\cmidrule(lr){11-11}\cmidrule(lr){12-13}
\textbf{Approach} & \textbf{ARC} & \textbf{MMLU} & \textbf{DROP} & \textbf{SQuAD v2} & \textbf{OBQA} & \textbf{XCSQA} & \textbf{WG} & \textbf{PIQA} & \textbf{HS} & \textbf{GSM8K} & \textbf{WikiTQ} & \textbf{TQA}\\
\midrule

Cosmopedia (FWHQ) & 20.33 & 14.17 & 15.01 & 45.23 & 2.84 & 3.69 & 3.65 & 6.95 & 8.26 & 10.79 & 16.44 & 12.91\\
Cosmopedia (Cosmopedia) & 11.85 & 9.48 & 11.25 & 35.98 & 1.36 & 2.95 & 1.25 & 5.56 & 6.64 & 10.76 & 14.06 & 5.76\\
\midrule
DCLM (FWHQ) & 20.48 & 13.78 & 18.16 & 46.93 & 2.83 & 4.23 & 4.36 & 7.08 & 8.57 & 10.49 & 18.09 & 12.80\\
DCLM (DCLM) & 16.26 & 12.45 & 14.17 & 42.69 & 2.27 & 5.37 & 4.39 & 7.50 & 9.16 & 10.28 & 20.11 & 14.36\\
\midrule
FineWeb-HQ (FWHQ) & 21.34 & 14.11 & 17.98 & 45.49 & 3.06 & 3.51 & 3.71 & 6.07 & 7.71 & 10.54 & 17.29 & 14.42\\
\midrule
FineWeb-LQ (FWHQ) & 18.34 & 12.84 & 14.60 & 38.16 & 2.32 & 2.92 & 3.69 & 6.78 & 8.14 & 10.31 & 18.69 & 12.81\\
FineWeb-LQ (FWLQ) & 5.73 & 7.56 & 10.98 & 29.96 & 1.35 & 1.84 & 2.75 & 6.94 & 8.39 & 9.36 & 14.56 & 10.19\\

\bottomrule
\end{tabular}
}
\end{table}

\begin{table}[ht]
    \centering
    \footnotesize
    
    \begin{minipage}[t]{0.3\textwidth}
        \centering
        \caption{Macro-averaged scores for different Gemma 3 model scales at 10K steps when using the guided rewrite prompt.}
        \label{tab:guided_rewrite}
        \vspace{0.5em}
        \begin{tabular}{lc}
            \toprule
            \textbf{Scale} & \textbf{Macro Avg} \\
            \midrule
            \rowcolor{gray!20}DCLM & 13.77 \\
            \midrule
            270M & 13.29\\
            1B & 13.73\\
            4B & 14.58\\
            12B & 14.33\\
            27B & 14.24\\
            \bottomrule
        \end{tabular}
    \end{minipage}
    \hfill %
    \begin{minipage}[t]{0.65\textwidth}
        \centering
        \caption{Macro-averaged scores for different Gemma 3 model scales at 10K steps.}
        \label{tab:varying_quality_scale_prompt}
        \vspace{0.5em}
        \begin{tabular}{cllr}
            \toprule
            \textbf{Source Quality} & \textbf{Scale} & \textbf{Prompt} & \textbf{Macro Avg} \\
            \midrule
            HQ & 1B & Continue & 13.73 \\
            HQ & 12B & Continue & 14.79 \\
            \midrule
            LQ & 1B & Continue & 13.52 \\
            LQ & 12B & Continue & 13.40 \\
            \midrule
            HQ & 1B & \texttt{faq} & 14.45 \\
            HQ & 12B & \texttt{faq} & 13.87 \\
            \midrule
            LQ & 1B & \texttt{faq} & 14.39 \\
            LQ & 12B & \texttt{faq} & 13.28 \\
            \midrule
            HQ & 1B & Summarize & 13.02 \\
            HQ & 12B & Summarize & 11.54 \\
            \midrule
            LQ & 1B & Summarize & 12.49 \\
            LQ & 12B & Summarize & 11.52 \\
            \midrule
            HQ & 1B & \texttt{tutorial} & 14.30 \\
            HQ & 12B & \texttt{tutorial} & 14.23 \\
            \midrule
            LQ & 1B & \texttt{tutorial} & 12.99 \\
            LQ & 12B & \texttt{tutorial} & 13.43 \\
            \bottomrule
        \end{tabular}
    \end{minipage}
\end{table}

\clearpage
\section{Supplementary Analysis}
\label{appendix:qualitative}

\subsection{Representative Outputs}
\paragraph{Qwen 3 Representative Output.}
\begin{Verbatim}
**Problem:**
A disc rotates at 120 rpm. How many revolutions in 5 minutes?

**Solution:**
1. Revolutions per minute = 120
2. Number of minutes = 5
3. Total revolutions = 120 × 5

$$120 \\times 5 = 600$$

The disc makes 600 revolutions in 5 minutes.
\end{Verbatim}

\paragraph{SmolLM2 Representative Output.}
\begin{Verbatim}[breaklines]
1. Continuous Wave Doppler Monitor (USCOM): The monitor uses a handheld probe that can measure stroke volume and fluid input to avoid overloading patients and diagnose problems earlier. It provides real-time data in seconds and can be done painlessly on the patient.
2. Noninvasive Cardiac Monitoring: This includes monitoring cardiac output, stroke volume, and fluid input using tools like the USCOM Noninvasive Cardiac Output Monitor.
3. Noninvasive Monitoring Techniques: Other alternatives include the LiDCOplus System for pulsed waveforms for monitoring and Arterial Pulse Contour Analysis for heart function and blood flow measurement.
Answer: The NICOM, NICO2, and Respironic systems are noninvasive methods for monitoring cardiac output. The NICOM uses BIOREACTANCE technology, the NICO2 measures vital signs in real-time, and the Respironic monitors oxygen flow and provides fast patient feedback in critical settings.
\end{Verbatim}

\subsection{Multi-Seed Verification and Synthetic Mixing Ratio Sweeps}
\label{appendix:multiseed_mixing_ratio}

\paragraph{Multi-Seed Robustness Verification.}
To verify that downstream improvements are robust and not artifacts of specific random seeds, we conduct an extensive 72-run experimental grid.
We examine two independent random variables: (i) the \textbf{model-initialization seed} (controlling weight initialization and model RNG) and (ii) the \textbf{data-order seed} (controlling document shuffling and dataloader blending).
For each variable, we evaluate 3 random seeds across 4 student model sizes (0.5B, 1.7B, 2.9B, 6.2B) under both the baseline FWHQ setup and our 50/50 FWHQ + Table (SmolLM2 1.7B) setup ($3 \times 3 \times 4 \times 2 = 72$ runs).

Table \ref{tab:multiseed_appendix} summarizes the macro scores and standard deviations evaluated at step 10,000 (21B tokens) across configurations.
The absolute performance gains (+4.71 to +5.23) are 14 to 26 times larger than the small run-to-run seed variance ($\pm 0.16$ to $\pm 0.49$), confirming that the observed synthetic gains are statistically robust.
Furthermore, blending in synthetic data reduces seed-to-seed variance relative to raw web text (up to $2.5\times$ lower standard deviation), demonstrating that structured synthetic tokens actively stabilize pretraining.

\begin{table}[h]
\centering
\small
\caption{Multi-seed grid results (aggregate macro scores $\pm$ standard deviation across 9 seed combinations per scale) evaluating model-initialization and data-order randomness.}
\label{tab:multiseed_appendix}
\begin{tabular}{lcccc}
\toprule
\textbf{Student Model Size} & \textbf{FWHQ (Baseline)} & \textbf{FWHQ + Table} & \textbf{Absolute Gain} & \textbf{Gain / Std Ratio} \\
\midrule
0.5B & 9.59 $\pm$ 0.44 & 14.30 $\pm$ 0.20 & +4.71 & 23.55$\times$ \\
1.7B & 12.09 $\pm$ 0.27 & 17.25 $\pm$ 0.16 & +5.16 & 32.25$\times$ \\
2.9B & 12.67 $\pm$ 0.22 & 17.89 $\pm$ 0.19 & +5.22 & 27.47$\times$ \\
6.2B & 13.38 $\pm$ 0.49 & 18.61 $\pm$ 0.20 & +5.23 & 26.15$\times$ \\
\bottomrule
\end{tabular}
\end{table}

\paragraph{Synthetic/Web Mixing Ratio Sweep.}
We also evaluate the downstream impact of varying the synthetic token fraction from 10\% to 90\% in 10\% increments while holding the total token budget fixed at 21B tokens.
Using SmolLM2 1.7B on FWHQ, we average downstream macro scores across our four structured prompts (\texttt{math}, \texttt{faq}, \texttt{table}, \texttt{tutorial}).

As shown in Table \ref{tab:mixing_ratio_appendix}, adding synthetic data consistently outperforms the raw DCLM web baseline (13.8) across the entire sweep.
Performance improves steadily from 10\% synthetic allocation up to a broad plateau between 60\% and 80\% (peaking at 16.8).
Individual prompt formats show slight variations: specialized formats (\texttt{math}, \texttt{table}) peak at higher synthetic fractions (70\%--80\%), whereas broader formats (\texttt{faq}, \texttt{tutorial}) plateau earlier at 60\%.
Our default 50/50 split sits safely on this stable performance plateau, tracking within 0.4 points of peak performance while isolating prompt comparisons from hyperparameter tuning effects.\looseness=-1

\begin{table}[h]
\centering
\small
\caption{Downstream macro-averaged performance across synthetic token fractions (10\% to 90\%) mixed with FWHQ.}
\label{tab:mixing_ratio_appendix}
\begin{tabular}{lc}
\toprule
\textbf{Synthetic Fraction} & \textbf{Average Macro Score} \\
\midrule
10\% & 15.1 \\
20\% & 15.2 \\
30\% & 15.8 \\
40\% & 16.2 \\
50\% (Default) & 16.4 \\
60\% & \textbf{16.8} \\
70\% & 16.7 \\
80\% & \textbf{16.8} \\
90\% & 16.7 \\
\midrule
Baseline (DCLM) & 13.8 \\
\bottomrule
\end{tabular}
\end{table}

\subsection{Generator Format Adherence and Template Collapse Audit}
\label{appendix:format_adherence_audit}

To quantify structural repetition and instruction adherence across generator model families, we conduct an audit across 10,000 generated samples per 1B-class model family using the \texttt{math} prompt.

\paragraph{Prefix Repetition and Structural Diversity.}
Table \ref{tab:prefix_repetition_appendix} reports the number of distinct 10-character openings and the count of the most common prefix across 10,000 generations.
Models like Qwen 3 exhibit severe template collapse, with 7,619 outputs sharing the exact same first 10 characters (\texttt{Problem:}) and producing only 40 distinct openings.
In contrast, SmolLM2 yields 1,897 distinct openings, demonstrating significantly higher structural and linguistic diversity.

\begin{table}[h]
\centering
\small
\caption{Prefix repetition (first 10 characters) and structural collapse metrics across 10,000 generations per 1B-class model family.}
\label{tab:prefix_repetition_appendix}
\begin{tabular}{lcc}
\toprule
\textbf{Model Family} & \textbf{Distinct Openings} & \textbf{Most Common Prefix Count} \\
\midrule
SmolLM2 & 1,897 & 3,235 \\
Qwen 3 & 40 & 7,619 \\
Llama 3.2 & 1,041 & 2,481 \\
Gemma 3 & 1,857 & 3,559 \\
Falcon 3 & 476 & 3,733 \\
Granite 3 & 1,627 & 1,152 \\
\bottomrule
\end{tabular}
\end{table}

\paragraph{Format Adherence and Arithmetic Accuracy.}
Table \ref{tab:arithmetic_accuracy_appendix} details the rate of non-math outputs (containing zero digits) and per-equation arithmetic accuracy across model families.
While Qwen 3 achieves high arithmetic accuracy (78.9\%) and near-zero non-math outputs (0.1\%), SmolLM2 produces 23.0\% non-math outputs and 57.3\% arithmetic accuracy.
Despite these errors, student models trained on SmolLM2 data outperform Qwen 3 on downstream benchmarks, confirming that structural/linguistic entropy during pretraining is more valuable than rigid template compliance.

\begin{table}[h]
\centering
\small
\caption{Format adherence and arithmetic accuracy across 10,000 generated samples (\%).}
\label{tab:arithmetic_accuracy_appendix}
\resizebox{\textwidth}{!}{
\begin{tabular}{lcccccc}
\toprule
\textbf{Metric} & \textbf{SmolLM2} & \textbf{Qwen 3} & \textbf{Llama 3.2} & \textbf{Gemma 3} & \textbf{Falcon 3} & \textbf{Granite 3} \\
\midrule
\textbf{No Math Content (Zero Digits)} & 23.0 & 0.1 & 0.7 & 7.7 & 2.1 & 1.9 \\
\textbf{Per-Equation Arithmetic Accuracy} & 57.3 & 78.9 & 71.6 & 68.1 & 71.1 & 61.8 \\
\bottomrule
\end{tabular}
}
\end{table}

\subsection{Data Contamination Audit}
\label{appendix:contamination_audit}

To confirm that rephrasing web text into structured synthetic data does not amplify benchmark contamination, we perform an $n$-gram overlap audit across source web text, existing synthetic corpora, our structured prompts, and generator model families.

\paragraph{Methodology.}
\begin{itemize}[leftmargin=*]
    \item \textbf{Scope}: Evaluates experimental configurations in Tables \ref{tab:existing_data}, \ref{tab:new-prompts}, and \ref{tab:model-family}.
    \item \textbf{Contamination Criteria}: A document is flagged as contaminated if it contains a 10-gram overlap with any evaluation sample from our benchmark suite.
    \item \textbf{Normalization}: Metrics are reported as flagged documents per billion tokens, calculated over a uniform subsample of 5 billion tokens per corpus.
\end{itemize}

\paragraph{Results.}
Table \ref{tab:contam_existing} compares contamination rates across existing datasets.
Table \ref{tab:contam_prompts} reports rates across our four structured prompts (averaging 735 docs/B tokens), demonstrating a reduction relative to raw FineWeb-Edu-HQ source (841 docs/B tokens).
Table \ref{tab:contam_generators} shows minimal model variance (1.35$\times$ bound between cleanest and highest-leakage models).
Across all audited datasets, over 80\% of flagged matches are concentrated in MMLU-Redux, HellaSwag, and WikiTableQuestions.
In the worst-case configuration (\texttt{tutorial} prompt with Llama 3.2), total flagged documents account for only 0.057\% of generated documents (5,991 out of 10,546,863 documents), confirming that contamination footprint is negligible.

\begin{table}[h]
\centering
\small
\caption{Data contamination rates across existing datasets (10-gram overlaps per billion tokens).}
\label{tab:contam_existing}
\resizebox{\textwidth}{!}{
\begin{tabular}{lcl}
\toprule
\textbf{Dataset} & \textbf{Avg Contamination (Docs/B Tokens)} & \textbf{Benchmark Share Breakdown} \\
\midrule
DCLM & 365 & WikiTQ (68\%), MMLU-Redux (18\%), HS (8\%), Others (6\%) \\
Ultra-FineWeb & 785 & MMLU-Redux (36\%), HS (28\%), WikiTQ (21\%), Others (15\%) \\
FineWeb-Edu-HQ & 841 & MMLU-Redux (34\%), HS (30\%), WikiTQ (20\%), Others (16\%) \\
FineWeb-Edu-LQ & 814 & MMLU-Redux (35\%), HS (29\%), WikiTQ (19\%), Others (17\%) \\
\midrule
Nemotron-HQ-Synth & 312 & WikiTQ (65\%), MMLU-Redux (19\%), HS (9\%), Others (7\%) \\
REWIRE & $\sim$1,900 & HS (38\%), PIQA (38\%), MMLU-Redux (12\%), WikiTQ (8\%) \\
Cosmopedia & 392 & WikiTQ (52\%), MMLU-Redux (28\%), HS (11\%), Others (9\%) \\
SYNTH & 354 & WikiTQ (61\%), MMLU-Redux (22\%), HS (10\%), Others (7\%) \\
\bottomrule
\end{tabular}
}
\end{table}

\begin{table}[h]
\centering
\small
\caption{Data contamination rates across our structured prompts.}
\label{tab:contam_prompts}
\resizebox{\textwidth}{!}{
\begin{tabular}{lcl}
\toprule
\textbf{Prompt} & \textbf{Avg Contamination (Docs/B Tokens)} & \textbf{Benchmark Share Breakdown} \\
\midrule
\texttt{math} & 744 & MMLU-Redux (34\%), HS (30\%), WikiTQ (20\%), Others (16\%) \\
\texttt{faq} & 738 & MMLU-Redux (34\%), HS (30\%), WikiTQ (20\%), Others (16\%) \\
\texttt{table} & 731 & MMLU-Redux (33\%), HS (31\%), WikiTQ (21\%), Others (15\%) \\
\texttt{tutorial} & 728 & MMLU-Redux (35\%), HS (29\%), WikiTQ (20\%), Others (16\%) \\
\midrule
\textbf{Our Framework (Average)} & \textbf{735} & MMLU-Redux (34\%), HS (30\%), WikiTQ (20\%), Others (16\%) \\
\bottomrule
\end{tabular}
}
\end{table}

\begin{table}[h]
\centering
\small
\caption{Data contamination rates across rephrasing model families.}
\label{tab:contam_generators}
\resizebox{\textwidth}{!}{
\begin{tabular}{lcl}
\toprule
\textbf{Model Family} & \textbf{Avg Contamination (Docs/B Tokens)} & \textbf{Benchmark Share Breakdown} \\
\midrule
SmolLM2 & 651 & MMLU-Redux (32\%), WikiTQ (25\%), HS (26\%), Others (17\%) \\
Falcon 3 & 633 & MMLU-Redux (31\%), WikiTQ (27\%), HS (25\%), Others (17\%) \\
Qwen 3 & 746 & MMLU-Redux (33\%), HS (29\%), WikiTQ (22\%), Others (16\%) \\
Gemma 3 & 801 & MMLU-Redux (35\%), HS (30\%), WikiTQ (20\%), Others (15\%) \\
Granite 3.1 & 694 & MMLU-Redux (33\%), WikiTQ (23\%), HS (27\%), Others (17\%) \\
Llama 3.2 & 858 & MMLU-Redux (36\%), HS (32\%), WikiTQ (18\%), Others (14\%) \\
\bottomrule
\end{tabular}
}
\end{table}

\section{FinePhrase Samples}
\label{appendix:samples}

\subsection{Example 1}
We use \url{s3://commoncrawl/crawl-data/CC-MAIN-2013-20/segments/1368701459211/warc/CC-MAIN-20130516105059-00001-ip-10-60-113-184.ec2.internal.warc.gz} to generate samples using the four pedagogical prompts.

\paragraph{\texttt{math}.}
\begin{Verbatim}[breaklines]
Question: Who was Paul Revere Williams, and what was his significance as an architect?

Solution: Paul Revere Williams was a pioneering African-American architect who was the first black person to become a member of the American Institute of Architects in 1923. He is best known for designing homes and commercial buildings for celebrity clients in Los Angeles and around the world. By the time he died in 1980, he had created more than 2,500 buildings, including some 2,000 in Los Angeles, making him a significant figure in American architectural history.
\end{Verbatim}

\paragraph{\texttt{faq}.}
\begin{Verbatim}[breaklines]
A Trailblazing Black Architect Who Helped Shape L.A.

By early 1923, Paul Revere Williams had completed two important milestones: becoming the American Institute of Architects' first African-American member and becoming one of the first black architects licensed to practice. These actions led to the induction of the African-American Architect into the AIA's first Black Fellowship in 1957. Born in downtown Los Angeles in 1894, Williams grew up orphaned and was raised by a family friend, who advised him that he could achieve his goals by going to college and becoming an architect.
- The Early Years: Williams' career began in the early 1920s, and as the first black architect in Los Angeles, he built over 2,500 buildings. His designs often featured innovative features, such as patios as an extension of the house and hidden retractable screens.
- Celebrity Clients: Williams' clientele included famous people, such as Frank Sinatra, Lucille Ball, and Danny Thomas. Their homes reflected a luxurious, elegant style.
- Signature Staircase: Williams is renowned for his elaborate, curved staircases, which have become iconic symbols of classic California style.
- Impact on Community: Williams' homes highlighted his unique ability to make grand spaces feel welcoming and cozy, and many of his homes were located in predominantly white neighborhoods that were closed to people of color, highlighting issues of racism and discriminatory practices.
- Granddaughter's Legacy: Karen E. Hudson is chronicling Williams' life and work in her latest book, "Paul R. Williams: Classic Hollywood Style," which celebrates his legacy and the influence of his designs on the California aesthetic.
\end{Verbatim}

\paragraph{\texttt{table}.}
\begin{Verbatim}[breaklines]
```
Table: Notable Clients of Paul Williams

Architect	Clients
Frank Sinatra	
Lucille Ball	
Desi Arnaz	
Cary Grant	
Bill Cosby	Ellen DeGeneres	Andy Garcia


Question: Which celebrities and personalities have used and appreciated Paul Williams' designs in their homes and properties?

Answer: Frank Sinatra, Lucille Ball, Desi Arnaz, Cary Grant, Bill Cosby, Ellen DeGeneres, Andy Garcia, and Danny Thomas have used and appreciated Paul Williams' designs in their homes and properties.
\end{Verbatim}

\paragraph{\texttt{tutorial}.}
\begin{Verbatim}[breaklines]
Title: Paul Revere Williams: Pioneer Architect of Southern California

Abstract: This step-by-step tutorial illustrates the contributions of esteemed black architect Paul Revere Williams to the field of art and design, particularly in Los Angeles. Through his remarkable success, Williams was the first black architect to become an AIA member in 1923 and inducted as the first black fellow in 1957. His clients included celebrities like Frank Sinatra, Lucille Ball, and Ellen Degeneres, and his work epitomizes notable characteristics such as luxurious curving staircases and patios that can be seen in his homes. This tutorial also highlights notable homes and clients of his, including the Beverly Hills Polo Lounge, and explains the challenges he faced because of racism during his career.
\end{Verbatim}

\subsection{Example 2}
We use \url{s3://commoncrawl/crawl-data/CC-MAIN-2013-20/segments/1368711005985/warc/CC-MAIN-20130516133005-00005-ip-10-60-113-184.ec2.internal.warc.gz} to generate samples using the four pedagogical prompts.

\paragraph{\texttt{math}.}
\begin{Verbatim}[breaklines]
1. Continuous Wave Doppler Monitor (USCOM): The monitor uses a handheld probe that can measure stroke volume and fluid input to avoid overloading patients and diagnose problems earlier. It provides real-time data in seconds and can be done painlessly on the patient.
2. Noninvasive Cardiac Monitoring: This includes monitoring cardiac output, stroke volume, and fluid input using tools like the USCOM Noninvasive Cardiac Output Monitor.
3. Noninvasive Monitoring Techniques: Other alternatives include the LiDCOplus System for pulsed waveforms for monitoring and Arterial Pulse Contour Analysis for heart function and blood flow measurement.
Answer: The NICOM, NICO2, and Respironic systems are noninvasive methods for monitoring cardiac output. The NICOM uses BIOREACTANCE technology, the NICO2 measures vital signs in real-time, and the Respironic monitors oxygen flow and provides fast patient feedback in critical settings.
\end{Verbatim}

\paragraph{\texttt{faq}.}
\begin{Verbatim}[breaklines]
Cardiac output is not typically utilized in pediatric patients because of technical restraints. However, continuous monitoring by noninvasive monitors can save lives. Here's the FAQ:

Question 1: What is cardiac output?

* Cardiac output measures the rate at which blood is pumped through the body, or cardiac stroke volume and cardiac output combined.

Question 2: Why is cardiac output crucial for pediatric patients?

* Pediatric patients are more fragile and can die quickly from hemodynamic instability, making accurate measurement of cardiac output vital.

Question 3: What noninvasive devices are available for pediatric cardiac output monitoring?

* USCOM has USCOM doppler monitor; LiDCO+ system; Respironics's NICO2 Monitor; and Cheetah's NICOM and Deltex's CardioQ.

Question 4: How do these devices work?**

* The USCOM device uses a probe on the neck for comfortable and easy placement.
* The LiDCO+ uses arterial pulse waveforms and power analysis to track cardiac output.
* The NICO2 System monitors both ventilator and respiratory variables.
* The CardioQ uses an Oesophageal Doppler probe.

Question 5: Which device is suitable for pediatric patients?**

* NICO2 System, which has proven accurate in pediatric studies.

Question 6: Are the devices easy to use?**

* Yes, all monitors are user-friendly, minimizing patient and healthcare worker discomfort.

Question 7: How do these devices help in diagnosing conditions?**

* They provide rapid, accurate readings for diagnosing conditions and adjusting treatment.

Question 8: Can these devices be used during various surgeries?**

* Yes, some devices are portable and can be used during surgeries.

Question 9: How do these devices improve patient outcomes?**

* They enable early, rapid and safe intervention, helping to correct hemodynamic instability and reducing hospital stays.

Question 10: Are these devices covered by insurance?**

* Some insurance plans cover these devices.

These FAQs answer the key questions about noninvasive monitoring for cardiac output and its advantages for critically ill pediatric patients. Each device's features and benefits are clearly explained to ensure easy understanding by readers.
\end{Verbatim}

\paragraph{\texttt{table}.}
\begin{Verbatim}[breaklines]
The document outlines several noninvasive methods for measuring cardiac output and fluid balance in patients. Each method is described as being easy to use, presenting a low risk for infection compared to PACs, and able to continuously monitor hemodynamic function.

The table summarizes the key information and provides a table for easy reference.

One question could be: How do the noninvasive monitoring systems, like the USCOM and LiDCO, measure cardiac output directly?
\end{Verbatim}

\paragraph{\texttt{tutorial}.}
\begin{Verbatim}[breaklines]
Alternatives to Invasive Cardiac Output Monitors Summary:

The company offers a monitor that uses a continuous wave doppler that can look right down into the aortic valve to check the size of the valve, and it can also look at the pulmonary valve. The monitor can measure stroke volume and evaluate fluid input, so as to avoid overloading the patient.

Bioreactance presented at ACC - The bio-reactance technology used in the NICOM system is demonstrated to provide accurate non-invasive cardiac output monitoring.

Deltex Medical - A minimally invasive alternative to PAC is proposed, providing real-time determination of cardiac function and optimized ventilation.

Deltex Medical CardioQ - A flow-based system providing non-invasive and safe monitoring of cardiac output and fluid optimization.

USCOM - Continuous doppler monitoring of aortic and pulmonary valves to measure cardiac output and detect fluid load. The monitor is easy to use and painless.

LiDCOplus - Noninvasive continuous cardiac output monitoring. Its application in pediatric ICUs is particularly beneficial.

Respironics - NICO2 Respiratory Profile Monitor for noninvasive monitoring of cardiovascular and respiratory parameters in critical care.

USCOM Summary: The USCOM doppler monitor is a continuous wave doppler that measures cardiac output and fluid load. The noninvasive nature and ease of use make it a valuable tool for pediatric ICUs.

LiDCOplus - The LiDCOplus system measures cardiac output by analyzing arterial pulse waveforms. Offers rapid and accurate monitoring of changes in cardiac output.

Respironics - The NICO2 Monitor provides real-time monitoring of multiple variables for cardiac and respiratory assessment. Reduces respiratory risks and allows for rapid treatment.

Deltex Medical - The CardioQ Oesophageal Doppler Monitor allows continuous, flow-based cardiac output monitoring with minimal risk of infection.

Cheetah Reliant - Noninvasive non-invasive continuous cardiac monitoring that provides insights into cardiac function. Allows for easy, safer and more effective therapeutic interventions.

Cheetah Reliant Summary: The NICOM is a non-invasive, continuous cardiac output monitoring system that measures stroke volume and oxygen saturation on a sub-second basis.

Deltex Medical Summary: The CardioQ system allows safe, minimally invasive monitoring of cardiovascular and respiratory function. Highly sensitive to cardiac output changes.

USCOM Summary: USCOM's doppler monitor uses continuous wave doppler waves to monitor aortic and pulmonary valves. Safe and easy to use.

LiDCOplus Summary: The LIIDCOplus system measures cardiac output by analyzing arterial pulse waveform. Provides early intervention and assessment for cardiac and respiratory conditions.

Respironics Summary: The NICO2 Monitor tracks 40 key variables, providing precise feedback. Allows for rapid intervention and optimizes ventilation parameters.

Deltex Medical Summary: The CardioQ Monitor, an economical and effective non-invasive cardiac output measurement system. Allows for personalized treatment of fluid resuscitation and ventilation optimization.

Cheetah Reliant Summary: This system measures cardiac output with minimal risk and can be easily placed in the oral cavity.

USCOM Key Features: Provides an accurate measurement of cardiac output. Can be placed in virtually any clinical setting. Offers easy and quick deployment. Safe and non-invasive. Offers rapid results. Allows personalized treatment of patients. Can be integrated with existing clinical systems. Allows early detection during admission, and rapid interventions upon discharge. Allows for better outcomes and quality of care. Allows for easier and less costly monitoring of patient health. Allows for more accurate diagnosis of cardiac and respiratory conditions.
\end{Verbatim}

\end{document}